\documentclass[a4paper]{cas-sc}
\usepackage[numbers]{natbib}
\usepackage{caption} % 放在导言区
\usepackage{subcaption} 
\usepackage{array}
\usepackage{algorithm}      % 提供 algorithm 环境
\usepackage[noend]{algpseudocode}    % 提供 algorithmicx 环境，支持 \Require 等命令
\usepackage[labelfont=bf, labelsep=colon]{caption}
\usepackage{hyperref}  % 确保导入了 hyperref 包
\usepackage{tabularx} % 确保加载此包
\usepackage[most]{tcolorbox} % 彩色框
\usepackage{multirow}     % 用于多行单元格（本表未使用，但保留以防扩展）
\usepackage{xcolor}  
\usepackage{booktabs} % 用于三线表
\usepackage{threeparttable}
\usepackage{makecell} % 导言区引入
\usepackage{enumitem} % 在导言区添加一次
\usepackage{adjustbox} % For adjustbox environment
\def\tsc#1{\csdef{#1}{\textsc{\lowercase{#1}}\xspace}}
\tsc{WGM}
\tsc{QE}
\tsc{EP}
\tsc{PMS}
\tsc{BEC}
\tsc{DE}
\begin{document}
\hypersetup{citecolor=blue,linkcolor=blue,urlcolor=blue}
\let\WriteBookmarks\relax
\def\floatpagepagefraction{1}
\def\textpagefraction{.001}
\shorttitle{}
\shortauthors{T. Bao et~al.}
%\begin{frontmatter}

\title [mode = title]{SurveyAgent-HKA: A multi-agent framework for scientific survey generation with LLMs and human knowledge augmentation}                      
\author[1,2]{Tong Bao}[style=chinese] 
\ead{tbao@njust.edu.cn}

\author[2]{Mir Tafseer Nayeem}
\ead{mnayeem@ualberta.ca}

\author[3]{Yi Zhao}[style=chinese]
\ead{yizhao93@ahu.edu.cn}

\author[2]{Davood Rafiei}
\ead{drafiei@ualberta.ca}

\author[1]{Chengzhi Zhang}[style=chinese, orcid=0000-0001-9522-2914]
\ead{zhangcz@njust.edu.cn}

\cortext[cor1]{Corresponding author: Chengzhi Zhang.}

\affiliation[1]{organization={Department of Information Management, Nanjing University of Science and Technology},
              city={Nanjing},
              country={China}}
              
\affiliation[2]{organization={Department of Computing Science, University of Alberta},
                city={Edmonton},
                country={Canada}}

\affiliation[3]{organization={School of Management, Anhui University},
city={Hefei},
country={China}}

\begin{abstract}
Automatic scientific survey generation has become an important task in scientific document processing. The common approach of retrieving literature from a single source (e.g., arXiv) and generating surveys through a one-pass large language model (LLM) call often leads to limited reference coverage and, more importantly, fails to replicate the expert-driven revision process that is crucial for writing high-quality surveys. In this paper, we introduce SurveyAgent-HKA, a multi-agent framework that improves end-to-end scientific survey generation by incorporating knowledge derived from published surveys and peer-review comments. The framework decomposes survey generation into well-defined sub-tasks handled by LLM-powered agent. It first retrieves relevant papers from multiple sources and identifies key topics through clustering to construct an initial outline, which is then refined using outlines from related human-written surveys. Based on the refined outline, topic-focused papers are retrieved and re-ranked to select for drafting a well-grounded survey. Then, we identify common issues raised by experts in peer-review comments from published surveys to guide the revisions and finalize the survey. Experiments on two domains show that our approach outperforms mainstream baselines in citation quality, structural consistency, and content quality. Furthermore, our framework is efficient in both time and cost, making it a practical solution for broader AI-assisted scientific writing applications.

\end{abstract}
% \begin{highlights}
% \item Research highlights item 1
% \item Research highlights item 2
% \item Research highlights item 3
% \end{highlights}

\maketitle

\section{Introduction}\label{sec1}
Within the scientific research community, the number of published academic papers has been growing rapidly. Taking preprints as an example, arXiv received more than 20,000 submissions per month on average, with over 280,000 submissions in 2025, nearly three times the number from a decade ago\footnote{\url{https://arxiv.org/stats/monthly_submissions}}. The rapid increase in publications helps accelerate scientific progress but also makes it harder for researchers to keep up with the latest developments \citep{bib1,bib2,bib3}. Scientific surveys play an important role in synthesizing existing research, highlighting current limitations and challenges, and offering valuable insights for future directions \citep{bib4,bib5}. However, the dramatic increase in publications poses significant concerns to traditional survey writing. On the one hand, drafting a high-quality scientific survey often requires weeks or even months of effort,  including literature search, paper screening, and summarization, especially in fast-moving fields \citep{bib6, PAUL2020101717}. On the other hand, survey quality largely depends on the author's domain knowledge and personal judgment, which may result in missed important studies or potential biases \citep{bib7, Wee03032016}. Therefore, automated scientific survey writing has become an important task in scientific document processing.

Recent advances in large language models (LLMs) \citep{brown2020language} have enabled promising progress in automated survey generation  \citep{wu2024automated, lai2024instruct, wang2024AutoSurvey, liang2025surveyx}. However, directly generating scientific surveys with LLMs remains challenging, as the generated content often lacks the logical coherence and structural rigor required for long-form scientific papers. Existing studies usually formulate scientific survey generation as a two-stage process: \textbf{(1) literature retrieval}, and \textbf{(2) survey generation}. In the retrieval stage, many studies use retrieval-augmented generation (RAG) methods \citep{izacard2022few, borgeaud2022improving, gao2023retrieval, jiang-etal-2023-active} to integrate external sources into an LLM for retrieving reliable literature. In the generation stage, evaluate whether LLMs can produce high-quality survey outlines and content, using both automatic metrics and human judgments. Despite the recent progress of LLM-based scientific survey generation, existing studies still treat survey generation mainly as a model-centered generation process. That is, the system retrieves a set of papers, asks LLMs to organize the content, and then uses prompt-based or self-refinement strategies to improve the generated draft. Although this paradigm has improved the efficiency of survey writing, it still differs substantially from how high-quality human-written surveys are produced. 

\textbf{First}, existing systems still provide limited support for building a reliable literature foundation. Many methods retrieve papers from a single platform, such as arXiv \citep{wang2024AutoSurvey, liang2025surveyx} or Semantic Scholar \citep{bao-etal-2025-surveygen}. In addition, most of them rely mainly on titles and abstracts for survey generation \citep{wang2024AutoSurvey, bao-etal-2025-surveygen, yan2025surveyforgeoutlineheuristicsmemorydriven}, without fully using full-text content, metadata, citation relations, and other scholarly signals. Such limited evidence may lead to incomplete or biased literature selection, reducing the coverage and reliability of the generated survey. \textbf{Second}, existing retrieval methods often pay limited attention to the quality of candidate papers. Most of them select papers based on the semantic similarity between the survey topic and textual information, such as titles and abstracts  \citep{yan2025surveyforgeoutlineheuristicsmemorydriven,tang2024llms}. While this approach facilitates efficient retrieval, it does not evaluate whether a paper is influential or makes an important contribution. A high-quality survey should not only synthesize prior work but also highlight studies that have made important contributions to the field and guided subsequent research  \citep{paul2020art,kanellos2019impact}.
Therefore, retrieval only on textual similarity may include papers that are relevant but less important, which weakens the scholarly value of the survey. \textbf{Third}, {\color{black} existing systems mainly rely on LLM-generated outlines, LLM-generated critiques, or prompt-based refinement during survey generation \citep{lai2024instruct, wang2024AutoSurvey, liang2025surveyx, yan2025surveyforgeoutlineheuristicsmemorydriven}. As a result, the planning and revision stages remain largely model-internal. In contrast, human-written surveys are often shaped by structural conventions from published surveys and refined through expert feedback from peer review \citep{bib5, bib7, Wee03032016}. These human-created knowledge artifacts provide valuable guidance on how a survey can be improved in terms of coverage, logic, and academic rigor. However, such human knowledge is time-consuming and has not been systematically incorporated into existing LLM-based survey generation frameworks.}

To address the above issues, we propose SurveyAgent-HKA, a multi-\textbf{\underline{agent}} collaborative approach for scientific \textbf{\underline{survey}} generation with LLMs and \textbf{\underline{h}}uman \textbf{\underline{k}}nowledge \textbf{\underline{a}}ugmentation. In SurveyAgent-HKA, we decompose survey generation into multiple subtasks, including literature retrieval, outline generation, survey writing, and survey refinement.  In the literature retrieval stage, we design a unified framework to integrate diverse literature sources, expanding the pool of candidate papers for subsequent generation and evaluation.  Next, we perform topic clustering on the retrieved literature to generate an initial outline that covers multiple subtopics closely aligned with the survey and further refine it using human-authored outlines from existing published surveys to improve its overall quality. In the generation stage, we first identify literature closely related to the subtopics,  then perform citation network analysis and re-ranking to select influential and representative papers for each topic, which are used to generate the corresponding sections and integrated to form the draft survey. Finally, peer review comments were collected from open published surveys on related topics to distill commonly raised concerns from reviewers into a structured checklist, which is used to refine the draft survey from the perspective of human experts, enhancing its overall quality to maintain academic rigor. In SurveyAgent-HKA, each subtask is assigned an LLM-powered agent, enabling collaboration and improving automated generation efficiency. Experiments conducted in two domains validate the performance of our proposed approach compared to the baseline. The contributions of our work are outlined below:
\begin{itemize}
    \item We propose SurveyAgent-HKA, a multi-agent framework that integrates human structural and review knowledge into scientific survey generation.
    
    \item We designed a unified literature retrieval pipeline, which not only helps retrieve relevant literature from multiple data sources but also selects representative papers through advanced citation network analysis and a re-ranking module, ensuring the coverage and reliability of the references for generating well-grounded scientific surveys.
     
    \item  {\color{black}We introduce two types of human knowledge augmentation for automatic survey generation. Hierarchical outlines extracted from human-written surveys provide structural guidance for outline construction, while real peer-review comments support targeted multi-round revision. To the best of our knowledge, this is the first work to use real review feedback from published survey papers to guide automatic survey revision.}

    \item Experiments on two domains show that the surveys generated by our model outperform mainstream baselines in citation quality, structural consistency, and content quality, with each module in SurveyAgent-HKA contributing to the overall performance. {\color{black}A user study further validates the practical value of the generated surveys}. 
\end{itemize}

The structure of the paper is as follows: In Section \ref{sec:related_work}, we review related work on scientific survey generation. In Section \ref{sec:method}, we present our proposed SurveyAgent-HKA. Section \ref{sec:experiments} describes the experimental setup and implementation details. In Section \ref{sec:results}, we present the experimental results and discuss the implications and limitations in Section \ref{sec:disscussion}. Finally, we summarize the conclusion and future work in Section \ref{sec:conclusion}.

\section{Related work}
\label{sec:related_work}
\subsection{Automated scientific survey generation}
Early approaches to automated scientific survey generation mainly relied on pre-trained models \citep{jha-etal-2015-content,hu-wan-2014-automatic,kasanishi-etal-2023-scireviewgen}, which aimed to compose survey drafts through the selection and aggregation of representative sentences from multiple sources. However, due to their extractive or template-driven nature, the generated surveys often lacked coherence and systematic organization. With the rapid development of LLMs, recent studies have used their capabilities to improve scientific survey generation \citep{wu2024automated, lai2024instruct, liang2025surveyx, tang2024llms}. For example, \citet{wang2024AutoSurvey} proposed AutoSurvey, which uses RAG to retrieve relevant literature, generate structured outlines, and produce survey content. Based on this work, \citet{liang2025surveyx} improved the search process with expanded keywords and used an attribute-based strategy to extract key information for survey writing. \citet{yan2025surveyforgeoutlineheuristicsmemorydriven} introduced SurveyForge, which achieved strong performance in LLM-based survey generation in reference selection, outline organization, and content quality. Recently, several practical systems have also been developed to further support LLM-based scientific writing. For instance, DeepResearch~\citep{deepresearch2025} and ScholarAI~\citep{scholarai2025} provide unified platforms to draft scientific surveys with minimal manual effort. At the evaluation stage, several studies have assessed LLM-generated surveys in terms of reference accuracy, hallucination rate, information coverage, and human preference \citep{wen2025interactivesurveyllmbasedpersonalizedinteractive,10.1145/3706598.3714047,qiu2025completingsystematicreviewhours, nguye2025surveyg}. {\color{black}Beyond full survey generation, recent work on scholarly retrieval and literature synthesis also provides useful insights for this task. For example, OpenScholar \citep{asai2026synthesizing} and Ai2 ScholarQA \citep{singh2025ai2} use retrieval-augmented methods to synthesize citation-grounded answers or reports from large-scale scientific corpora. In addition, systematic-review automation has also explored how LLMs can support search, screening, data extraction, and evidence synthesis \citep{weng2025cycleresearcher}.} 

\textbf{Existing Limitations.}
In summary, these studies provide valuable foundations for automated survey generation. However, several limitations remain. First, existing approaches show fragmented retrieval sources and text-based textural ranking strategy, with most experiments conducted in the computer science domain and using arXiv as the primary data source. This limited setting, together with an overreliance on semantic similarity for paper ranking, may reduce the breadth and reliability of the retrieved literature. Second, while many methods aim to optimize surveys using the generative capabilities of LLMs, they fail to incorporate human-written feedback and revision improvements, which are essential to ensure the survey meets academic standards expected in scientific papers. Third, cross-domain evaluation is still limited. Most studies focus on computer science, while survey writing practices can differ greatly across disciplines. This makes it difficult to assess whether these methods can generalize to other domains. Finally, in the evaluation process, LLMs are mainly used to assess the generated survey, with limited comparison to human-written ground truth. Without such comparison, it's difficult to determine if the generated surveys align with human expectations, which makes it unclear what the upper bound of these models is for this task and how they might impact real-world applicability.

{\color{black}We differ from previous works in several aspects, as summarized in Table \ref{tab1:baselines}}. \textbf{First}, our SurveyAgent-HKA framework supports multi-source data retrieval, which not only provides a more diverse and comprehensive set of references but also incorporates quality signals to evaluate and re-rank the literature, helping to identify more representative and influential publications in the field. \textbf{Second}, SurveyAgent-HKA supports downloading and parsing full-text PDFs of candidate papers, automatically extracting key information to produce a structured summary. By leveraging full texts rather than abstracts alone, this approach provides a more complete view of each paper for survey generation. \textbf{Third}, Third, SurveyAgent-HKA incorporates human-derived knowledge into both outline construction and multi-round revision. This design helps the generated surveys better follow structural conventions from published surveys and address common issues raised by expert reviewers. \textbf{Lastly}, our framework supports cross-domain survey generation based on a multi-source data retrieval system and compares the results with human-written gold standards, offering strong generalization capabilities compared to baseline approaches.

\subsection{Multi-agent systems for text generation}
A Multi-Agent System (MAS) is composed of several autonomous agents that work together to achieve a common goal. These agents are often powered by LLMs and have the ability to understand human instructions or make self-directed decisions based on their environment, task requirements, or the information inputted \citep{10.24963/ijcai.2024/890, Zhang2021, Zhao_Huang_Xu_Lin_Liu_Huang_2024,saadaoui2025coordinated}. MAS has been applied to various content generation tasks such as medical treatment \citep{li2025agenthospitalsimulacrumhospital, wang2023voyageropenendedembodiedagent,han2025plug}, programming assistance \citep{islam-etal-2024-mapcoder, hong2024metagpt}, and social interaction simulation \citep{gao2025s3socialnetworksimulationlarge}. In long-text writing tasks, \citet{huot2025agentsroomnarrativegeneration} proposed AgentsRoom, which employs multiple agents to handle narrative planning and content generation for story writing separately. \citet{wang2024autopatentmultiagentframeworkautomatic} designed AutoPatent, where a planner agent, writer agents, and an examiner agent collaboratively perform the patent drafting process. \citet{shao-etal-2024-assisting} propose STORM, a multi-agent system capable of generating long-form articles comparable in scope and depth to Wikipedia entries. \citet{darcy2024margmultiagentreviewgeneration} propose MARG, an approach for generating feedback that leverages multiple LLM instances to engage in internal discussions for providing more specific and helpful feedback on scientific papers. \citet{xia2025storywriter} presents StoryWriter, a modular, open-source multi-agent framework designed for scalable and controllable long story generation.  {\color{black}Recent studies on agentic RAG and LLM self-critique further demonstrate the value of agent-based workflows in knowledge-intensive scientific tasks. For example, SIM-RAG \citep{yang2025knowing} and RAG-CRITIC \citep{dong2025rag} show that retrieval and self-evaluation can improve the reliability of generated content.}

To sum up, the above studies demonstrate that MAS holds strong potential for long-form document generation. However, its application in scientific writing remains underexplored.  In this work, we view scientific survey writing as a multi-stage workflow that naturally benefits from task decomposition and role-based agent collaboration. Accordingly, we design an LLM-powered multi-agent framework for this task. By using scientific survey generation as a representative scenario, we investigate the potential of MAS for scientific content generation and provide insights for broader AI-assisted scientific writing.

\begin{table}[h!]
\centering
\caption{{\color{black}Comparison with mainstream survey generation baselines. Our framework supports multi-source literature retrieval, full-text utilization,  human knowledge augmentation, and enables cross-domain survey generation. Moreover, we compare against human-written ground truth for evaluation rather than relying on LLM-based evaluation to avoid potential bias.}}

\begin{tabular}{lcccccccc}
\toprule
\textbf{Model} & \textbf{Year} & \textbf{\makecell{Data\\Source}} & \textbf{\makecell{Quality\\Signals}} & \textbf{\makecell{Re-ranking\\Module}} & \textbf{\makecell{Full\\Text}} & \textbf{\makecell{Knowledge\\Augmentation}} & \textbf{\makecell{Cross\\Domains}} & \textbf{\makecell{Ground\\Truth}} \\
\midrule
Naive-RAG \citep{lai2024instruct} &2024&Single & x & x & x & x & x & \checkmark \\
AutoSurvey  \citep{wang2024AutoSurvey} &2024 &Single & x & x & x & x & x & x\\
SurveyX  \citep{liang2025surveyx} &2025 &Multi & x & x & \checkmark & x & x & x\\
SurveyGen  \citep{bao-etal-2025-surveygen} &2025 &Single & \checkmark & \checkmark & x & x & \checkmark & \checkmark \\
SurveyForge  \citep{yan2025surveyforgeoutlineheuristicsmemorydriven} &2025 &Single & x & \checkmark & x & \checkmark & x & x\\
\textbf{SurveyAgent-HKA(Ours)} & - &Multi & \checkmark & \checkmark & \checkmark & \checkmark & \checkmark & \checkmark \\
\bottomrule
\end{tabular}
\label{tab1:baselines}
\end{table}
\vspace{-10pt}

\section{Methodology}
\label{sec:method}
This section first outlines the task formulation for automatic scientific survey generation and then provides the design of our SurveyAgent-HKA and its implementation in the scientific survey generation task.

\subsection{Problem formalization}
Automatic scientific survey generation is formally defined as: Given a survey topic \( T \) and a large academic corpus \( D = \{d_1, d_2, \dots, d_n\} \), the task aims to draft a comprehensive survey \( S \) that summarizes the subset of documents  \( D \) that are relevant to \( T \). Specifically, this task can be decomposed into two sequential stages: \textbf{(1)} \textbf{\textbf{paper retrieval}}: for a survey topic \( T \), a retrieval system identifies a subset of relevant papers \(\mathcal{R}_{\mathbf{T}} \in D\), ensuring sufficient coverage to provide a thorough overview of the topic (formally, $T, D \to \mathcal{R}_{\mathbf{T}}$). \textbf{(2)} \textbf{\textbf{survey generation}}: based on the topic \( T \) and the retrieved document set $\mathcal{R}_{\mathbf{T}}$, a generative system generates the final survey $\mathbf{S}$ (formally, $T, \mathcal{R}_{\mathbf{T}} \to S$). The output  \( S \) is expected to be structured into sections corresponding to subtopics within \( T \) and to incorporate claims from the retrieved literature $\mathcal{R}_{\mathbf{T}}$  into a long-form and well-structured survey. The main symbols used in this paper are summarized in Table~\ref{tab:symbols}.

\begin{table}[ht]
\centering
\caption{List of symbols used in this paper.}
\label{tab:symbols}
\begin{tabular}{ll}
\toprule
\textbf{Symbol} & \textbf{Description} \\ \midrule
\( T \) & Survey topic \\ 
\( D \) & Set of documents in the academic corpus \\ 
\( \mathcal{R}_{\mathbf{T}} \) & Set of relevant papers for the survey topic \( T \) \\ 
\( \mathcal{K} \) & Cluster of papers based on embeddings \\ 
\( \mathcal{P} \) &  Prompt used to guide LLMs for generation \\ 
\( \mathcal{O} \) &  The generated survey outline \\ 
\( \mathcal{E} \) & Template tree for extracting knowledge from candidate literature \\ 
\( \mathcal{F} \) & A specific section in the survey \\ 
\( S \) & The generated survey \\ 
\texttt{[RETRIEVER]} & Agent  for retrieving literature from external databases  \\
\texttt{[RANKER]} & Agent  for assessing and ranking retrieved papers \\
\texttt{[EXTRACTOR]} & Agent  for extracting structured knowledge from retrieved literature \\
\texttt{[OUTLINER]} & Agent  for constructing the hierarchical outline of the survey \\
\texttt{[WRITER]} & Agent  for expanding the outline into detailed textual sections and drafting the survey \\
\texttt{[REVIEWER]} & Agent for reviewing and revising the generated survey \\
\texttt{[REFINER]} & Agent  for polishing and finalizing the generated survey \\
\( C \)  & Central Router (coordinates the order of agent execution) \\
\( M \) & Memory Module (stores intermediate results and context information) \\
\bottomrule
\end{tabular}
\end{table}

\begin{figure}[pos=htbp]
    \centering
    \includegraphics[width=0.70\linewidth]{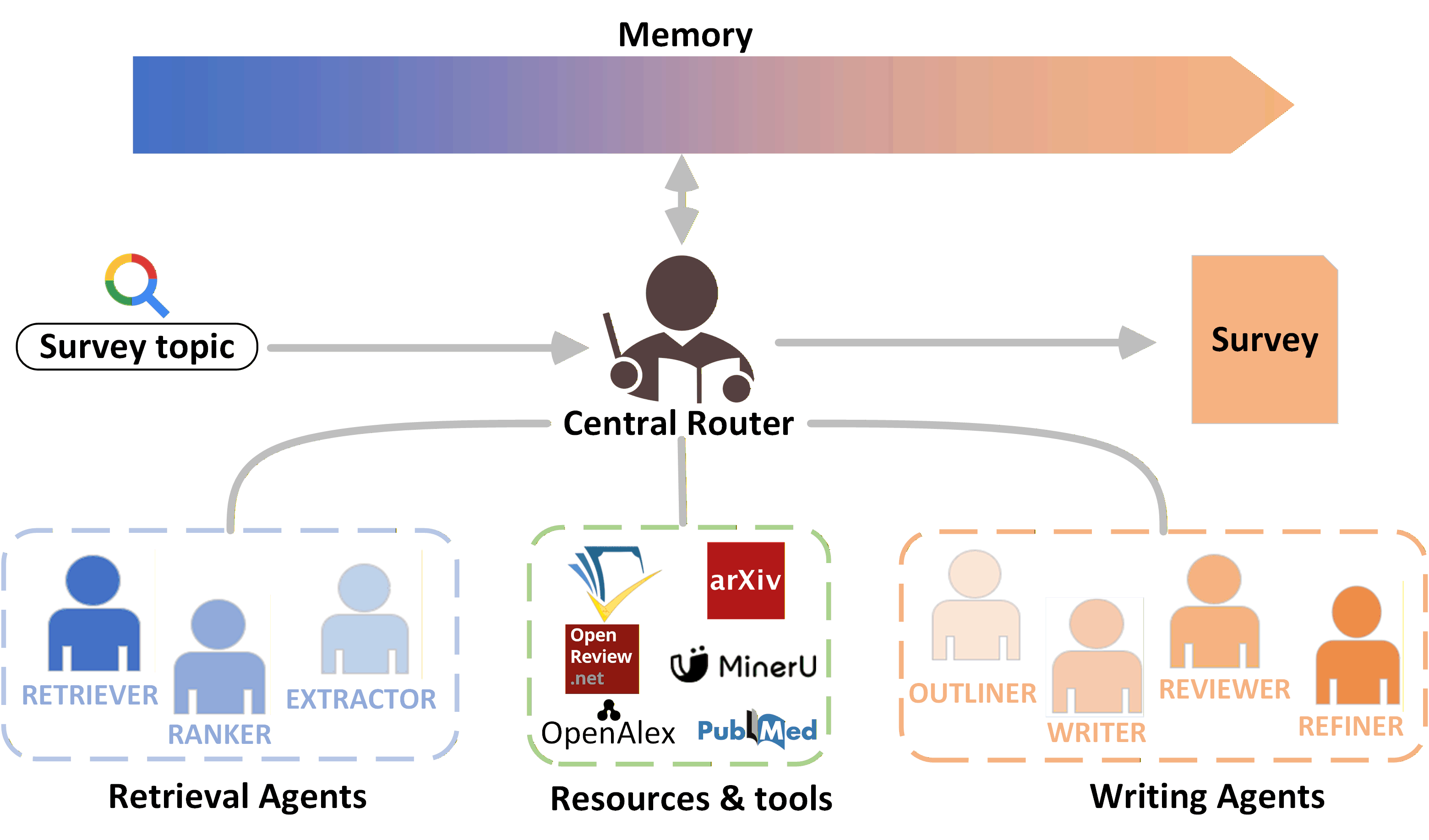} 
\caption{The framework of SurveyAgent-HKA, which involves multiple agents collaborating through a central router and a memory module, with shared resources and tools supporting end-to-end scientific survey generation.}
\label{fig:1}
\end{figure}

\subsection{SurveyAgent-HKA: The overall architecture}
Figure \ref{fig:1} presents the overall architecture of SurveyAgent-HKA. Given a survey topic \( T \), the goal of the framework is to automatically generate a corresponding scientific survey \( S \). To achieve this, we break the survey generation into several sub-tasks and deploy multiple agents, each responsible for a unique stage of the process, that work together under a central router to draft the topic-related survey. Below, we describe the definition of the agents and the agent design in our SurveyAgent-HKA framework.

\begin{itemize}
    \item \textbf{Agents.} An agent \( a \in \mathcal{A} \) is defined as an autonomous entity designed to perform specific subtasks that contribute to achieving the overall goal. In this definition, an agent can be an LLM that was fine-tuned on task-specific data, a zero-shot prompted LLM with a user-defined prompt, or a simple functional module with instructions.  In scientific survey writing, we focus on the zero-shot prompted LLM as an agent, as it has strong capabilities to understand user input and execute the complex generation process. However, we treat this agent as a modular component to support future work, which may integrate other types of agents (e.g., image recognition, speech understanding, and multimodal LLMs) for different purposes. We design two types of agents: \textit{\textbf{retrieval agents}} and \textit{\textbf{writing agents}}, along with \textit{\textbf{central router}} and \textit{\textbf{memory module}}, each of which performs distinct functions to collaborate in the survey generation pipeline. 

    \item \textbf{Retrieval Agents.} Retrieval agents are designed for retrieving, filtering, and analyzing relevant literature  $\mathcal{R}_{\mathbf{T}}$ from external databases based on the survey topic \( T \).  We design three retrieval agents, each responsible for specific phases of the retrieval process. The \textbf{[RETRIEVER]} is responsible for searching external literature databases \( D \) (e.g., Semantic Scholar, arXiv) to collect papers related to the target survey topic \( T \), together with their metadata (e.g., DOI, abstracts, citations, and accessible PDF links). In addition, it is also responsible for retrieving existing survey-related materials, such as human-authored survey outlines and topic-relevant peer-review comments, to support downstream processes.
    The \textbf{[RANKER]} is designed to evaluate the candidate papers and to re-rank them to obtain a more representative and accurate subset $\mathcal{R}_{\mathbf{T}}$. Finally, the \textbf{[EXTRACTOR]} parses PDFs and extracts the information required for subsequent writing stages from the candidate literature. 
    
    \item \textbf{Writing Agents.} Writing agents are designed to summarize and generate survey content based on the retrieved literature $\mathcal{R}_{\mathbf{T}}$. Given that LLMs often struggle to generate sufficiently long text content in a one-step call, we divide the writing process into three steps and design corresponding agents to perform each of them. First, the \textbf{[OUTLINER]} constructs a hierarchical outline \( \mathcal{O} \) that captures the main topics and logical flow of the survey, serving as a guiding framework for follow-up survey generation. The \textbf{[WRITER]} develops the outline \( \mathcal{O} \) into a set of detailed sections \( \{S_{i-1}, S_i, \dots\, S_{n} \} \subseteq S \). Each section is grounded in claims and evidence drawn from the literature retrieved for that specific subtopic by the retrieval agents, and these sections are integrated to form the full survey draft \( S \).  Then, the \textbf{[REVIEWER]} simulates a peer reviewer by assessing the draft survey \(S\) against insights observed in real-world survey peer reviews and produces targeted suggestions for improvement. Finally, the \textbf{[REFINER]} polishes the draft  \( S \), improving its clarity, coherence, and structure to meet academic writing standards. Therefore, the final survey can be regarded as the product of a pipeline of components, each building upon the outputs generated by the individual agents.

    \item \textbf{Central Router.}  As different agents are responsible for executing specific sub-tasks of the survey generation process, we design a \textit{central router } $C$ to manage the collaboration among agents according to the workflow of scientific survey writing. Given an input instruction or an intermediate output $s_t$ along with a set of designed agents $A$, the router $C$ decides which agent $agent_{t+1} \in A$ should be invoked next. To keep track of the workflow, at each time step $t$, we record the agent's label together with its output as a tuple $(\mathit{agent}_t, \mathit{agent}_t^\mathrm{output})$, which serves as input reference for the next agent to be called. Additionally, the central router is responsible for determining the stopping condition (e.g., when all subtasks are completed or when there is no further improvement in results).

    \item \textbf{Memory Module.}  Since the entire survey generation process is based on collaboration between different agents, a \textit{memory module}  $M$ is designed to save and update the status throughout the execution of the framework. Unlike the \textit{central router}, which focuses on task delegation, the \textit{memory module} serves as dynamic storage that shares key information throughout the process. This includes agent outputs, intermediate results, and any context that the next agents need to refer to.  For instance, if \textbf{[OUTLINER]} constructs a survey outline that needs to be referenced later by \textbf{[WRITER]} or \textbf{[REFINER]}, the structure is stored and made accessible in memory. In this module, different agents can share information to avoid generating duplicate or conflicting content. Additionally, the memory module can track all retrieved references and validate them in the final version to maintain consistency, preventing the LLM from including hallucinated references during generation.

    \end{itemize}
    
\begin{figure}[pos=htbp]
    \centering
    \includegraphics[width=\linewidth]{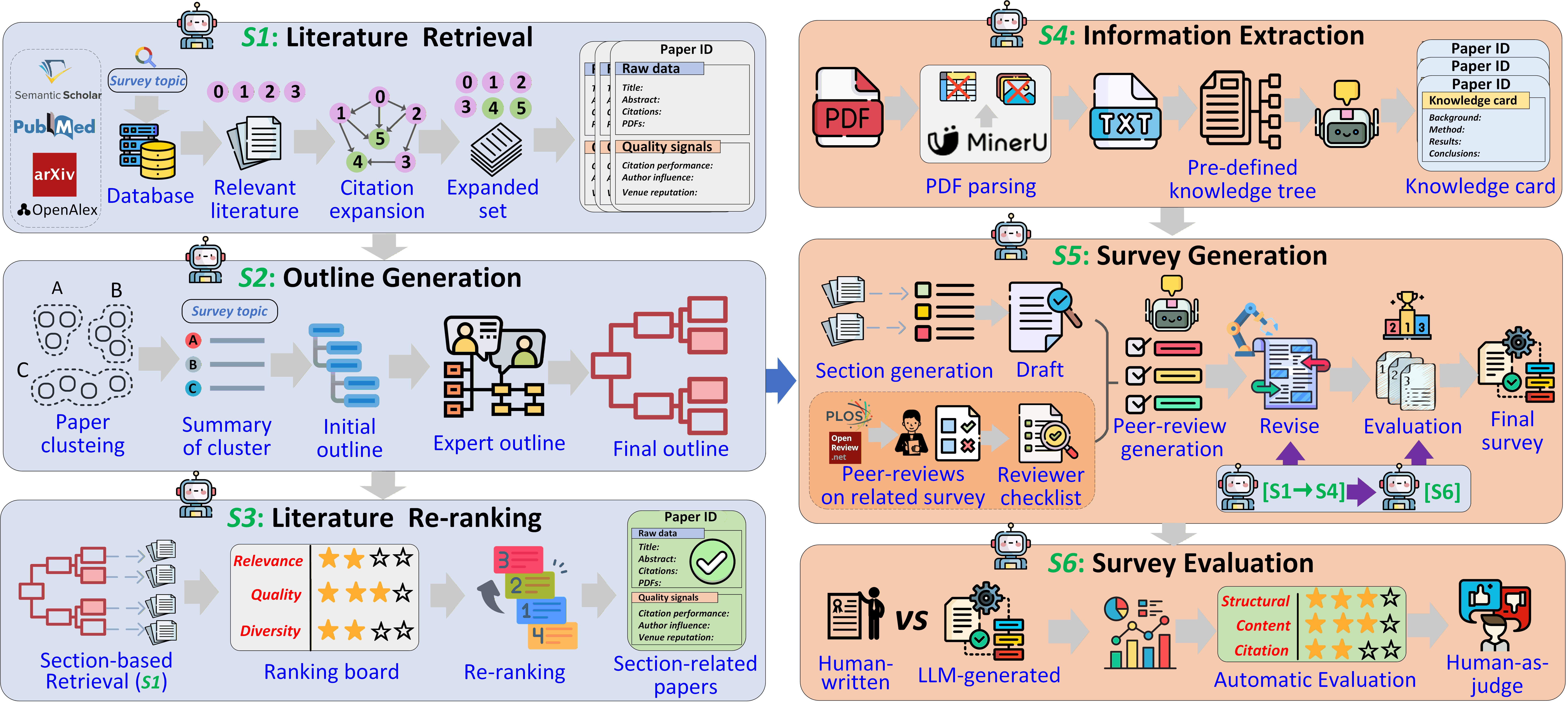} % 修改为你的图片路径
    \caption{The overall framework SurveyAgent-HKA, which includes two main stages: literature retrieval and survey generation. In the retrieval stage, relevant papers are collected from multiple sources \textbf{[S1]} and clustered to create an initial outline \textbf{[S2]}. Based on this outline, a section-focused retrieval is performed to collect more topic-specific papers, which are re-ranked to obtain the final reference set for each section \textbf{[S3]}. In the generation stage, the full texts of retrieved papers are used to generate the corresponding sections and compose the survey, which is then refined based on expert peer-review comments collected from related published surveys \textbf{[S4-S5]}. Finally, the generated survey is evaluated using both automatic metrics and human judgment \textbf{[S6]}.}
    \label{fig:2}
\end{figure}
\vspace{-10pt}

\begin{algorithm}[ht]
\caption{Algorithm of SurveyAgent-HKA.}
\label{algorithm1}
\begin{algorithmic}[1]
\State \textbf{Input:} Survey topic $T$, external database $D$, human reference outlines $\mathcal{O}_\text{human}$, peer reviews on survey $\mathcal{PR}$
\State \textbf{Output:} Final survey $S_\text{final}$

%================ First Round: Outline Generation =================%
\State $D_\text{initial} \gets \text{RetrieveFromDatabase}(T, D)$
\State $D'_\text{top} \gets \text{Top-n papers by semantic similarity from } D_\text{initial}$
\State Compute embeddings $\mathbf{v}_i \gets \text{Embedding}(\text{abstract}_i), \forall d_i \in D'_\text{top}$
\State Cluster embeddings: $\{\mathcal{K}_1, \dots, \mathcal{K}_K\} \gets \text{Cluster}(\{\mathbf{v}_i\})$
\State Summarize each cluster: $\mathbf{s}_i \gets \text{Summarize}(\mathcal{K}_i)$
\State Generate initial outline: $\mathcal{O}_\text{initial} \gets \text{GenerateOutline}(T, \mathbf{s}_i)$
\State Refine outline: $\mathcal{O}_\text{refined} \gets \text{RefineOutline}(\mathcal{O}_\text{initial}, \mathcal{O}_\text{human})$

%================ Second Round: Section Drafting =================%
\State $T_\text{section} \gets \text{ExtractSectionTitles}(\mathcal{O}_\text{refined})$

\For{each section $t \in T_\text{section}$}
    \State $D_\text{sec} \gets \text{RetrieveBySectionTitle}(t, D)$
    \State $D_\text{ranked} \gets \text{ReRank}(D_\text{sec})$
    \State Initialize $\mathcal{S}_i \gets \emptyset$
    \For{each paper $d_i \in D_\text{ranked}$}
        \State $\mathcal{S}_i \gets \mathcal{S}_i \cup \text{DraftSection}(t, d_i)$
    \EndFor
\EndFor

\State $S_\text{draft} \gets \text{AssembleSurvey}(\{\mathcal{S}_i\})$
%================ Third Round: Peer Review-Guided Refinement =================%
\State $\mathcal{PR}_\text{relevant} \gets \text{RetrievePeerReviewsByTopic}(T, \mathcal{PR})$
\State $\textit{Checklist} \gets \text{ExtractChecklist}(\mathcal{PR}_\text{relevant})$
\State $\textit{Comments} \gets \mathrm{LLM}(S_\text{draft}, \textit{Checklist}) \rightarrow \text{GenerateComments}$
\State $S_\text{revised} \gets \text{Refine}(S_\text{draft}, \textit{Comments})$
\State $S_\text{final} \gets \text{SelectOptimalSurvey}(S_{\text{revised}})$
\State \Return $S_\text{final}$
\end{algorithmic}
\end{algorithm}

\subsection{SurveyAgent-HKA for scientific survey generation}
\label{sec: Implementation}

As shown in  Figure \ref{fig:2}, the SurveyAgent-HKA consists of multiple agents responsible for performing different subtasks. For better understanding, the complete algorithm is summarized in Algorithm~\ref{algorithm1}. In this section, we describe the implementation of SurveyAgent-HKA for the scientific survey generation task. 

\subsubsection{Literature retrieval and citation-based expansion}
Given a user-input survey topic $T$, to increase retrieval efficiency and capture a broader set of relevant documents, we first use an LLM to perform query expansion to transform the user-provided topic into multiple queries (e.g., a set of keywords). Then, we first retrieve relevant literature from external academic databases (e.g., Semantic Scholar) to ensure that all retrieved papers are authentic and verifiable, which serve as grounding sources for the subsequent content generation.
Let $\mathcal{D} = \{d_1, d_2, \dots, d_n\}$ denote the papers collected from these sources. We use a text embedding model to encode each document and the expanded survey topic $T$, and then compute their cosine similarity as follows:

\begin{equation}
\operatorname{Sim}(T, d_i) = \cos(\mathbf{v}_T, \mathbf{v}_{d_i})
= \frac{\mathbf{v}_T \cdot \mathbf{v}_{d_i}}
{\|\mathbf{v}_T\| \, \|\mathbf{v}_{d_i}\|} \rightarrow \textbf{[RETRIEVER]}
\end{equation}
where $\mathbf{v}_T$ and $\mathbf{v}_{d_i}$ refer to the embeddings of the survey topic and the abstract of candidate papers.

We select the top-$n$ articles based on their embedding similarity to form the candidate set $D'$. However, relying solely on semantic similarity for retrieval is not robust enough and may miss key papers that are semantically distant from the query. For example, the paper that introduced the ``\textit{Transformer}'' \citep{vaswani2017attention} architecture may receive a low similarity score when queried with the topic
``\textit{Large Language Models}'', since neither its title nor abstract directly mentions the topic. Yet, it is widely cited in the publications of this field and recognized as a foundational work in this area. To address this, we perform a co-citation expansion to identify papers that are cited by at least two articles in $D'$ and add them to the candidate set.
Here, the assumption is that papers which are co-cited are likely to be topically related to those in $D'$, and are therefore considered indirectly relevant to the target topic $T$ because \( D \) is directly related to \( T \). Let $D_{\text{co}}$ denote the set of co-cited papers. The final expanded candidate set is defined as:
\begin{equation}
D_{\mathrm{ex}} = D' \cup D_{\mathrm{co}} \rightarrow \textbf{[RETRIEVER]}
\end{equation}

 Given that relying on a single data source may limit coverage or miss relevant literature, the agent is designed to query multiple academic databases (e.g., Semantic Scholar, arXiv) and aggregate the returned results using DOIs to remove duplicates.

\subsubsection{Dynamic survey outline generation and enhancement}
A well-structured outline is crucial for scientific survey writing as it not only helps organize the survey's content but also prevents common issues in one-shot LLM-based generation, where the output may be too short to cover essential topics. Although LLMs have demonstrated potential to generate outlines, they still do not meet the high standards required for scientific surveys \citep{shi2025scisage}. In this paper, we introduce a dynamic outline generation strategy to iteratively refine the outline based on the retrieved literature and LLM-based optimization. 
For each paper in $D_{\text{ex}}$, we encode its abstract into a feature vector using a text embedding model to capture its semantic content:
\begin{equation}
\mathbf{v}_i = \text{Embedding}(\text{abstract}_i), \quad \forall i \in D_{\text{ex}}
\end{equation}

We then use K-Means \citep{mcqueen1967some} to group papers into predefined clusters based on their semantic embeddings. 
Formally, let $\{\mathbf{v}_1, \mathbf{v}_2, \dots, \mathbf{v}_n\}$ be the set of paper embeddings. 
K-Means aims to partition these embeddings into $K$ clusters by minimizing the within-cluster sum of squared distances:
\begin{equation}
\{\mathcal{K}_1, \dots, \mathcal{K}_K\} = \arg\min_{\mathcal{K}_1, \dots, \mathcal{K}_K} \sum_{j=1}^{K} \sum_{\mathbf{v}_i \in \mathcal{K}_j} \|\mathbf{v}_i - \boldsymbol{\mu}_j\|^2,
\end{equation}
where $\boldsymbol{\mu}_j$ is the centroid of cluster $\mathcal{K}_j$, and the $K$ is determined based on the distribution of topics observed in published surveys and prior research in the field. 

Let \( \mathcal{K} = \{\mathcal{K}_1, \mathcal{K}_2, ..., \mathcal{K}_N\} \) represent the set of clusters. For each cluster $\mathcal{K}_i$, the abstracts of all included papers are fed into the LLM to generate a cluster summary $\mathbf{s}_i$, which captures the main themes and key topics of the cluster. Next, given the survey topic $T$ and the set of cluster summaries $\{\mathbf{s}_1, \dots, \mathbf{s}_k\}$, the LLM is prompted to generate a hierarchical outline that lays out the scope and main topics of sections, while also providing a short explanation of the purpose of each section. The generated initial outline is denoted as:
\begin{equation}
\mathcal{O}_{\text{initial}} = \text{LLM}(T, (\mathcal{K}_1, \mathbf{s}_1), (\mathcal{K}_2, \mathbf{s}_2), \dots, (\mathcal{K}_N, \mathbf{s}_N)),  \mathcal{F}_i)\rightarrow \textbf{[OUTLINER]}
\end{equation}
where \( \mathbf{s}_i \) is the summary of the cluster \textit{\( \mathcal{K}_i \)}.

% Based on the initial outline $\mathcal{O}_{\text{initial}}$, the LLM is further prompted to generate detailed subsections, expanding the outline into a hierarchical structure that specifies the scope and key content of each section. To maintain thematic and structural coherence, the survey topic $T$, the overall structure of $\mathcal{O}_{\text{initial}}$, and the prompt $\mathcal{P}$ are provided as input. The expanded hierarchical outline is denoted as:

% \begin{equation}
% \mathcal{O}_{\text{ex}} = \text{LLM}(\mathcal{O}_{\text{initial}}, T, \mathcal{F}_i) \rightarrow \textbf{[OUTLINER]}
% \end{equation}

To further refine the expanded outline and make it more aligned with human writing preferences, we collect a set of human-written survey outlines that are highly semantically similar to the focus topic as the optimization target. These human-authored outlines serve as the gold standard of high-quality, well-structured published surveys, guiding the LLM in improving the coherence and structure of the initial outline through few-shot learning. Let \( \mathcal{O}_{\text{ref\_structure}} \) denote the set of human-authored outlines. The refined outline can be expressed as:

\begin{equation}
\mathcal{O}_{\text{refined}} = \text{LLM}(\mathcal{O}_{\text{initial}}, \mathcal{O}_{\text{human\_written}}) \rightarrow \textbf{[OUTLINER]}
\end{equation}

As mentioned above, our dynamic outline generation method iteratively updates the initial outline by leveraging literature clustering to identify core themes, LLM-based expansion to enrich the outline with more detailed subsections, and human-authored outlines as few-shot learning to ensure the structure and language align with human writing standards. We then use this refined outline to guide the survey generation process.

\subsubsection{Topic-focused literature retrieval and re-ranking}
In survey writing, each section usually focuses on a specific subtopic, and therefore requires more targeted literature. However, our first-round search mainly concentrated on broad surveys, which may have missed some of these specialized works. To address this, we conduct a second-round retrieval using the section titles as queries to find literature that better matches the specific information needs of each section. The retrieval follows the same steps as the first round, but usually returns fewer documents since it is more targeted. Let $D_\text{sec}$ denote the set of documents retrieved for the corresponding section. While the papers in $D_\text{sec}$ are related to the section topic, not all are equally suitable for inclusion, as a survey should cover foundational and representative research in the field. Therefore, we further re-rank the $D_\text{sec}$ to identify the most representative and impactful works for the section. Specifically, based on the metadata of the candidate papers, we evaluate them from the following three aspects: 
\textbf{(1) }\textbf{topical relevance}, \textbf{(2)} \textbf{academic impact}, and \textbf{(3)} \textbf{recent popularity}. These aspects were chosen because they ensure a comprehensive selection of papers that are not only highly relevant to the topic but also influential within the academic community and current research trends, which are commonly emphasized factors in scientific survey writing outlines \citep{bib7, Wee03032016}. The descriptions of these aspects are summarized in Table~\ref{tab:paper_evaluation}.

\begin{table}[ht]
\centering
\caption{Evaluation criteria for candidate papers.}
\label{tab:paper_evaluation}
\begin{tabular}{@{}ll@{}}
\toprule
\textbf{Aspects} & \textbf{Description} \\ 
\midrule
Topical relevance & To ensure that the paper is closely related to the target topic. \\
Academic impact & To select papers that have greater scholarly influence within the research field. \\
Recent popularity & To identify papers that reflect current trends or emerging areas of interest in the field. \\
\bottomrule
\end{tabular}
\end{table}

For topical relevance, we use LLM-as-judge to assess the relevance between each \( d_i \) in \( D_{\mathrm{ex}} \) and the section topic \( T_\text{sec} \). The relevance score can be denoted as:
\begin{equation}
Score_t = \text{LLM}_{\text{judge}}(d_i, T_\text{sec})  \rightarrow \textbf{[RANKER]}
\end{equation}

For academic impact, we assess the academic influence of each paper from three dimensions: \textit{citation performance}, \textit{author influence}, and \textit{venue reputation}, as these are commonly used indicators for evaluating the innovation and impact of academic papers \citep{DONTHU2021285,hicks2015bibliometrics}. To balance the contributions of each dimension, we assign a weight to each dimension based on its relative importance in determining the overall academic impact as follows:
\begin{equation}
Score_a = \alpha \cdot \text{citation performance} + \beta \cdot \text{author influence} + \gamma \cdot \text{venue reputation} \rightarrow \textbf{[RANKER]}
\end{equation}
where \( \alpha, \beta, \gamma \) are the weights assigned to citation performance, author influence, and venue reputation, respectively.

For recent popularity, we define the score as the product of citations in the past three years (relative to the survey's publication time) and the logarithm of total citations:

\begin{equation}
Score_r(d_i) = (\text{citations in past 3 years}) \times \log(\text{total citations}) \rightarrow \textbf{[RANKER]}
\end{equation}

Lastly, the final score for each paper is defined as:
\begin{equation}
Score_{\text{final}} = \frac{Rank_t + Rank_a + Rank_r}{3} \rightarrow \textbf{[RANKER]}
\end{equation}

where \(Rank_t, Rank_a, Rank_r\) denote the ranks of the paper in topical relevance, academic impact, and recent popularity, respectively. The articles are then ranked in ascending order of \(Score_{\text{final}}\), with the top-K articles \(\mathcal{R}_{\mathbf{T}}\) selected as representative works that are most relevant to the section for section-level composition.

\subsubsection{Information extraction from candidate papers}

Previous studies typically rely on abstracts as the primary information source. However, we argue that abstracts are insufficient in coverage and often omit critical information such as methodological details and experimental findings. Therefore, we download the PDFs of the papers in $D_\text{sec}$  for information extraction. Specifically, for papers whose PDFs are successfully obtained, we use MinerU\footnote{\url{https://github.com/opendatalab/MinerU}} \citep{niu2025mineru25decoupledvisionlanguagemodel} to extract the main textual content from the PDFs and then clean the extracted text to remove noise and ensure consistency. During this process, non-textual elements such as figures and tables are removed. Next, we employ an LLM to extract essential information from each paper, following the same predefined template tree $\mathcal{E}$, introduced in SurveyX \citep{liang2025surveyx}. This template outlines a hierarchical structure that covers key elements such as the background, methodology, datasets, and results. Guided by this template, the LLM organizes the extracted content from each paper into a structured ``\textit{Knowledge Card}'', providing a standardized representation that can be used by downstream components. Finally, for papers whose PDFs cannot be downloaded, we follow common practice in prior studies and use abstracts as the primary source for downstream generation.

\begin{equation}
d_i  = 
\mathrm{}
\mathrm{Title}_i \oplus 
\begin{cases}
\mathrm{Knowledge\,Card}=\mathrm{LLM}_{}(\mathrm{Full Text}_i,\mathcal{E}), & \text{if PDF available} \\
\mathrm{Abstract}_i, & \text{otherwise}
\end{cases}
\Biggr) \rightarrow \mathbf{[EXTRACTOR]}
\end{equation}

\subsubsection{Reference-grounded survey drafting and peer review-guided refinement}
To ensure comprehensive coverage and improve generation efficiency,  instead of writing the entire survey in one go, we first write each subsection in parallel to focus on the specific topic in detail, ensuring that each subsection is both comprehensive and well-supported by claims from relevant references. Specifically, for each subsection \( \mathcal{F}_i \), we use the generated outline \( \mathcal{O}_{\text{refined}} \), the set of section-relevant documents \( D_\text{sec} \), and the prompt \( \mathcal{P} \) to guide the LLM for section drafting as follows:
\begin{equation}
\mathcal{S}_i = \mathrm{LLM}\bigl(\mathcal{O}_{\text{refined}}, D_\text{sec}, \mathcal{P} \bigr ) \rightarrow \mathbf{[WRITER]}
\end{equation}

Here, all statements and arguments in the generated content are grounded in  $D_\text{sec}$. Once all the subsections are written, we assemble them to form the complete raw survey \( S_{\text{draft}}\). 

To further improve the quality and scholarly reliability of \( S_{\text{draft}} \), we introduce a reviewer-guided refinement stage that leverages peer-review comments from published surveys as an external source of expert knowledge. These comments, which reflect expert judgment formed through rigorous academic evaluation, are retrieved by matching the target topic with existing surveys to ensure domain relevance. Formally, given the target survey topic \( T \) and a corpus of peer review comments \( \mathcal{PR} \) from published surveys, the reviewer comments retrieval can be defined as computing the semantic similarity between the target survey topic and the titles and abstracts of published surveys, and selecting the comments associated with the most similar surveys:
\begin{equation}
\mathcal{PR}_{T}
=
\arg\max_{\mathcal{PR}_i \in \mathcal{PR}}
\; \mathrm{Sim} (T, \mathcal{PR}_i)
\;\;\rightarrow\; \mathbf{[RETRIEVER]}
\end{equation}

Given \( \mathcal{PR}_{T} \), we prompt the LLM to identify the key aspects that reviewers frequently focus on when evaluating survey papers, such as coverage completeness, structural organization, and clarity of taxonomy. By aggregating these reviewer-identified aspects across surveys within the same domain, we construct a domain-specific review checklist, denoted as \( \textit{Checklist}_{T} \), and then the \textbf{[REVIEWER]} evaluates the draft survey \( S_{\text{draft}} \) based on the generated checklist \( \textit{Checklist}_{T} \), to identify potential issues and generate a list of targeted revision comments, denoted as \( \textit{Comments}_{T} \):
\begin{equation}
\textit{Comments}_{T} = \mathrm{LLM}(S_{\text{draft}}, \textit{Checklist}_{T}) \rightarrow \mathbf{[REVIEWER]}
\end{equation}

Once the revision \( \textit{Comments}_{T} \) is generated, it triggers the survey refinement process. These \( \textit{Comments}_{T} \) are used to guide the LLM to refine the draft survey to the revised survey, ensuring that the survey better reflects the knowledge and insights gained from experts, thus meeting the high academic rigor outlined in \( \mathit{Checklist}_T \).
This process can be denoted as:
\begin{equation}
S_{\text{revised}} = \mathrm{LLM}(S_{\text{draft}}, \textit{Comments}_{T}) \rightarrow \mathbf{[REVIEWER]}
\end{equation}

Here, based on the suggested revisions, we can invoke the previously defined agents to perform additional actions, such as literature retrieval and content supplementation. We trigger this process by setting a task flag signal in the output. The revised survey is then put through further automated evaluation, where different versions are assessed to select the optimal survey draft. As the final step, we refine the raw survey to polish it to proof-ready version:
\begin{equation}
S_{\text{final}} = \text{LLM}(\mathcal{S}_{\text{revised}},  \mathcal{P} )\rightarrow \mathbf{[REFINER]}
\end{equation}
where \( \mathcal{P} \) denotes a refinement prompt that includes instructions for maintaining consistency in formatting, standardizing references and citations, organizing sections and subsections logically, and ensuring correct grammar throughout the generated survey.

\section{Experiments}
\label{sec:experiments}
In this section, we present the experimental setup, including comparison baselines, datasets, and implementation details. We then describe the evaluation protocol as well as the prompt design for LLMs.
\subsection{Baselines}
We compare our model against several strong baselines for scientific survey generation, including:

\textbf{(1) Naive-RAG}  (2024)~\citep{lai2024instruct}: A standard retrieval-augmented framework that retrieves relevant documents through similarity search and directly uses an LLM for survey generation.

\textbf{(2) AutoSurvey}  (2024)~\citep{wang2024AutoSurvey}: An automated survey generation system that consists of multiple stages, including literature retrieval, LLM-based drafting, and iterative refinement.

\textbf{(3) SurveyX}  (2025)~\citep{liang2025surveyx}: An efficient and organized automated survey generation pipeline that divides the process into preparation and generation to ensure a streamlined workflow.

\textbf{(4) SurveyGen}  (2025)~\citep{bao-etal-2025-surveygen}: An RAG-based framework that explores the performance of LLMs in survey generation under different input settings.

\textbf{(5) SurveyForge}  (2025)~\citep{yan2025surveyforgeoutlineheuristicsmemorydriven}: A framework that improves survey generation by using human-written surveys as references, which is considered the current state-of-the-art (SOTA).

\textbf{(6) DeepResearch}  (2025)~\citep{deepresearch2025}: A deep semantic retrieval paradigm that leverages multi-stage reasoning to retrieve, aggregate, and synthesize information from various online sources to perform complex tasks, such as survey generation.

\subsection{Dataset}
\vspace{4pt}
\subsubsection{Dataset for survey generation}
\vspace{4pt}

We conduct experiments in two domains: \textbf{Computer Science} and \textbf{Medicine}, as they are representative fields with rapidly growing publications and are fast-moving in scientific research. For computer science, we select the same set of 20 surveys employed in AutoSurvey \citep{wang2024AutoSurvey}, which have also been used for comparison in subsequent works in SurveyX \citep{liang2025surveyx} and SurveyForge \citep{yan2025surveyforgeoutlineheuristicsmemorydriven}. These surveys are highly cited and cover a wide range of topics within the field, and they are also used as targets for survey generation in both the Naive-RAG \citep{lai2024instruct} and DeepResearch \citep{deepresearch2025}. For each survey, we retrieve the corresponding human-written version from arXiv and parse the PDF using MinerU \citep{niu2025mineru25decoupledvisionlanguagemodel} to extract the outline, content, and reference list as the gold standard; all identifying information, tables, and images are removed. For the medicine domain, we use SurveyGen \citep{bao-etal-2025-surveygen}, a dataset that includes 4.2K human-written scientific surveys with full-text content, structured outlines, and complete reference lists for all included surveys. We also select the same 10 surveys from the medicine domain for fair comparison. A comparison setting of the baseline models is provided in Table~\ref{tab:4}, and detailed information on these surveys is provided in Appendix~\ref{appendixA}.

\begin{table}[ht]
\centering
\caption{Selected survey examples for evaluation across domains and baselines.}
\label{tab:4}
\begin{tabular}{llll}
\toprule
\textbf{Domain} & \textbf{Data Source} & \textbf{\#Survey} & \textbf{Compared Baseline} \\
\midrule
Computer Science
& AutoSurvey \citep{wang2024AutoSurvey}  & 20 & Naive-RAG \citep{lai2024instruct} \\
& & & AutoSurvey \citep{wang2024AutoSurvey} \\
& & & SurveyX \citep{liang2025surveyx} \\
& & & SurveyForge \citep{yan2025surveyforgeoutlineheuristicsmemorydriven} \\
& & & DeepResearch \citep{deepresearch2025} \\

\midrule
Medicine
& SurveyGen \citep{bao-etal-2025-surveygen}  & 10 & SurveyGen \citep{bao-etal-2025-surveygen} \\
& & & Naive-RAG \citep{lai2024instruct} \\
& & & DeepResearch \citep{deepresearch2025} \\
\bottomrule
\end{tabular}
\end{table}

\vspace{4pt}
\subsubsection{Peer-review comments collection for survey papers}
\vspace{4pt}

\begin{table}[htbp]
\centering
\caption{Statistics of the collected peer-review comments, where each comment represents the complete feedback from an individual reviewer and may include multiple specific points.}
\label{tab:paper_comments}
\begin{tabular}{l l l l}
\toprule
\textbf{Domain} & \textbf{Venue} & \textbf{\#Survey} & \textbf{\#Peer-reviews} \\
\midrule
Computer Science
 & ICLR   & 12 & 45 \\
 & EMNLP  & 20 & 44 \\
&  F1000  & 29 & 83 \\
\midrule
Medicine
 & PLOS One & 188 & 1039 \\
 & F1000  & 231 & 467 \\

\bottomrule
\label{tab:peerreviewstats}
\end{tabular}
\end{table}

As mentioned earlier, peer-review comments on high-quality, published survey papers will serve as human feedback signals to refine and improve the generated surveys. For the computer science domain, we first collect peer-review comments from OpenReview\footnote{\url{https://openreview.net/}}, a public platform for peer reviews of most top-tier computer science conferences. Specifically, we download peer reviews for papers submitted to conferences that provide open peer review comments, such as ICLR and EMNLP, two leading conferences in the fields of AI and the broader computer science community. We filter the survey-type papers by checking whether the title of a submission contains keywords such as \textit{``a survey''}, \textit{``a systematic review''}, \textit{``a review''}, etc. Following this, we conduct a secondary verification using LLMs to judge whether the title and abstract accurately represent a survey-type paper according to the following criteria: \textbf{(i)} the title and abstract clearly indicate a systematic overview and summary of a specific research area, and \textbf{(ii)} the paper focuses on summarizing prior studies rather than introducing new techniques or empirical findings.  Next, we further extend our data collection by adding peer comments from F1000 Research\footnote{\url{https://f1000research.com/}}, a platform known for publishing open-access, peer-reviewed research in various scientific domains. We restrict the articles to be survey-type papers and focus on the computer science domain.

For the medical domain, we collect peer-review comments from DeepReview \citep{zhu-etal-2025-deepreview}, a large-scale benchmark that includes peer-review comments from the well-known medical journal PLOS One\footnote{\url{https://journals.plos.org/plosone/}}. We use the same two-stage process to identify published survey papers and include the corresponding peer-review comments in our collection. Similarly, we also supplement peer review comments from survey papers published in F1000 Research in the medical domain.  In total, we collected peer-review comments for 480 published survey papers that covered various topics and research areas in computer science and medicine. Table \ref{tab:peerreviewstats} summarizes the peer review comments collected in our study, while Appendix~\ref{app:peerreviews} provides two examples of these peer reviews.

\subsection{Evaluation metrics}
We design a comprehensive evaluation protocol that covers citation quality, structural consistency, and content quality, including automatic and human evaluation, to evaluate our proposed framework. For all evaluations, we use the collected human-written surveys as the ground truth.
\subsubsection{Automatic evaluation}
\textbf{\textbf{(1) Citation quality.}} To evaluate the reliability of the retrieved module, we select the \textit{Precision}, \textit{Recall}, and \textit{F1 score} of the retrieved papers by comparing them with the references selected by human authors.  Specifically, we calculate the pairwise textual similarity between the titles of the retrieved papers and the corresponding titles in the human-written surveys. Two titles are considered a correct match if their textual similarity exceeds 0.95\footnote{This threshold accounts for minor variations such as differences in mathematical symbols, formatting, or emojis.}. The metrics are defined as follows:

\begin{equation}
\textit{Precision} = \frac{\text{Number of correctly selected citations}}{\text{Total number of citations selected by the model}}
\end{equation}

\begin{equation}
\textit{Recall} = \frac{\text{Number of correctly selected citations}}{\text{Total number of citations in the human-selected set}}
\end{equation}

\begin{equation}
F_1 = 2 \times \frac{\text{Precision} \times \text{Recall}}{\text{Precision} + \text{Recall}}
\end{equation}

\textbf{(2) Structural consistency.} This metric evaluates how well the generated survey aligns with the structural organization of human-written surveys. Since our model generates surveys with a hierarchical section up to the second level, we denote first-level and second-level section headings as \textit{H1} and \textit{H2}, respectively, and evaluate their alignment using the corresponding F$_1$ scores, \textit{F1-H1} and \textit{F1-H2}. To address variations in heading phrasing, we consider a heading as matched when its semantic similarity with the corresponding human-written heading exceeds a predefined threshold of 0.8, a threshold determined through preliminary experiments and prior work to optimally capture valid semantic matches. We define the heading matching \textit{F$_1$} score as:
\begin{equation}
\textit{Match}_{\text{F1}} = \frac{2 \times (H_{\text{human}} \cap H_{\text{generated}}) }{\ H_{\text{human}}  +  H_{\text{generated}} }
\label{eq:smatch}
\end{equation}

where $ H_{\text{human}} $ and $ H_{\text{generated}} $ denote the sets of headings in the human-written and LLM-generated surveys, respectively, and $  H_{\text{human}} \cap H_{\text{generated}} $ represents the number of matched headings.

{\color{black} In addition to heading matching, we further employ LLM-as-judge as an auxiliary evaluator to assess the overall \textit{structural relevance} between the generated and human-written surveys from 1 (poor) to 5 (excellent). We evaluate each pair of outlines ten times using the same prompt and report the average score.} To ensure a fair and focused comparison, we exclude non-content sections such as ``\emph{References}'', ``\emph{Acknowledgements}'', ``\emph{Funding}'', and ``\emph{Appendices}'' from all structural consistency evaluations.

\textbf{(3) Content quality.} To evaluate content overlap between the generated and human-written surveys, we use established summarization metrics, including \textit{Semantic Similarity} and \textit{ROUGE} score \citep{lin2004rouge}. The former assesses how closely the meanings or concepts of the two surveys align, independent of exact word choice. The \textit{ROUGE} score quantifies the overlap by identifying common subsequences between the two surveys. Precision, recall, and F$_1$ score for \textit{ROUGE} score are computed as:

\begin{equation}
\textit{Precision}_{\text{n-gram}} = \frac{\text{N-grams}(generated) \cap \text{N-grams}(human)}{\text{N-grams}(generated)}
\end{equation}

\begin{equation}
\textit{Recall}_{\text{n-gram}} = \frac{\text{N-grams}(generated) \cap \text{N-grams}(human)}{\text{N-grams}(human)}
\end{equation}

\begin{equation}
\textit{F}_1 = \frac{2  \times  \text{Precision}_{\text{N-gram}}  \times  \text{Recall}_{\text{N-gram}}}{\text{Precision}_{\text{N-gram}} + \text{Recall}_{\text{N-gram}}}
\end{equation}

where $\text{n-grams}(generated)$ and $\text{n-grams}(human)$ denote the sets of n-grams in the generated survey $G$ and the human-written survey $H$, respectively. In this work, we report \textit{ROUGE-L}\footnote{Based on the official Python package, all ROUGE scores reported include 95\% confidence intervals.}, which computes the longest common subsequence (LCS) between the generated and ground-truth surveys. We report the balanced F$_1$ score to ensure a more robust evaluation of content overlap.

Given the potential bias of traditional metrics when evaluating uncontrollable generated text \citep{NguyenLuongKBS2025}, we also compute \textit{Key Point Recall (KPR)} \citep{qi2024long2rag}, which measures the proportion of key points in the original survey that are successfully covered by the generated survey. To do so, an LLM is used to extract key points from the human-written surveys, forming the reference set. We then use a question-answering approach to determine whether each key point is covered in the generated survey. Let $K_H$ denote the set of key points from the human-written survey and $K_G$ the set of key points present in the generated survey. The \textit{KPR} score is then defined as:

\begin{equation}
\textit{KPR} = \frac{1}{|K_H|} \sum_{i=1}^{|K_H|}
\begin{cases} 
1, & \text{if } k_i^H \in K_G, \\ 
0, & \text{otherwise.} 
\end{cases}
\end{equation}

\subsubsection{Human evaluation}
\textbf{Pairwise Comparison.} We perform human evaluation based on the same three aspects as previous studies \citep{wang2024AutoSurvey,liang2025surveyx,yan2025surveyforgeoutlineheuristicsmemorydriven}: (1) \textit{Coverage:} the extent to which the survey addresses the main aspects of the input topic; (2) \textit{Structure:} the clarity and coherence of the organization of sections and subsections; and (3) \textit{Relevance:} the relevance of the content to the user-provided survey topic. The criteria for the three metrics are summarized in Appendix \ref{app:criteria}. As survey evaluation requires domain expertise and significant time effort, we limit human evaluation to the computer science domain. We conduct two groups of pairwise comparisons. First, we conduct an internal comparison between the initial and final versions generated by our SurveyAgent-HKA to validate the benefits of improving the quality of the survey through multi-agent collaboration refinements. Second, we compare our generated survey with the corresponding human-written survey to explore the upper bound of our approach. For each group, we select 10 surveys generated by our model and pair each of them with its RAG-generated or a human-written survey on the same topic. We invite five PhD students with background in computer science, specializing in downstream applications of LLMs and each holding at least two peer-reviewed publications, to conduct the evaluation based on our predefined criteria. To minimize potential bias, the assigned topics are adjusted according to the annotator's domain expertise to ensure that all judges are knowledgeable about the topics they evaluate. The annotators are asked to compare each pair and judge which survey is better, comparable, or worse along each evaluation dimension and provide a brief explanation. To ensure evaluation reliability, annotators are provided with detailed guidelines and training prior to annotation, and all identifying information is removed to support a double-blind evaluation setup.

\textbf{User Study.} {\color{black} We further conduct a user study to assess the practical value of the generated surveys. We select 10 generated surveys from the computer science domain and invite 30 participants, including undergraduate, master’s, and PhD students. To ensure that the evaluations are informed by relevant background knowledge, participants may select one or more surveys based on their familiarity with the corresponding topics. For each selected survey, they rate its usefulness on a scale from 1 (poor) to 5 (excellent) across three aspects: \textbf{(1) Field Overview}, whether the survey helps them quickly understand the target topic; \textbf{(2) Useful References}, whether the survey helps them identify important papers in the field; \textbf{(3) Research Insights}, whether the survey provides useful information about research trends, limitations, or future directions. Participants are also encouraged to briefly describe any limitations they identify in each survey. All participants receive the same evaluation instructions, and responses are collected anonymously to reduce possible bias.}

\subsection{Implementation details}
\label{sec:implementation}
\vspace{4pt}
\textbf{\textbf{(1) Literature retrieval and re-ranking}}
\vspace{4pt}

\begin{table}[ht]
\centering
\caption{Key parameters for SurveyAgent-HKA implementation.}
\label{tab:parameters}
\begin{tabular}{lll}
\toprule
\textbf{Parameter} & \textbf{Set Value} & \textbf{Test Range} \\
\midrule
Top-n related papers (first-round) & 200 & (150,200,250) \\
Section-based retrieval & 60 & (40,60,80,100) \\
Co-cited papers added & 20 & (10,20,30) \\
Candidate papers per section & 80 & -- \\
Final retained papers per section & 50 & (30,50,70,90) \\
Number of clusters for outline generation & 6 & {\color{black}(4,5,6,7)} \\
Citation performance indicator & Total/3-year citations & -- \\
Venue reputation indicator  & Venue h-index & -- \\
Author impact indicator & H-index of first/last authors & -- \\
Re-ranking weight: Citation performance & 0.5 & (0.3,0.5,0.7) \\
Re-ranking weight: Author impact & 0.2 & (0.1,0.2,0.3) \\
Re-ranking weight: Venue reputation & 0.3 & (0.2,0.3,0.4) \\
Related surveys used for peer-review comments collection & 20 & -- \\
LLM for data processing and evaluation & GPT-4.1-mini &--\\
\bottomrule
\end{tabular}
\end{table}

We use \textbf{Semantic Scholar}\footnote{\url{https://api.semanticscholar.org/api-docs/datasets}} as our primary literature retrieval corpus, as it contains over 80 million peer-reviewed English scientific papers. To ensure domain-specific coverage, we supplement it with papers from \textbf{arXiv}\footnote{\url{https://arxiv.org/help/api}} for computer science, the most widely used preprint platform in the field, and papers from \textbf{PubMed}\footnote{\url{https://www.nlm.nih.gov/databases/download/pubmed_medline.html}}, which includes 36 million biomedical papers in the medical domain. For Semantic Scholar, we directly perform retrieval using its official API, while for arXiv, we employ the \texttt{bge-large-en-v1.5}\footnote{\url{https://huggingface.co/BAAI/bge-large-en-v1.5}} as an embedding model to encode 530,000 computer science papers as the retrieval database, following the same setup as AutoSurvey \citep{wang2024AutoSurvey}. During the first-round retrieval, for each topic, we calculate the semantic similarity between the survey topic and the abstracts of candidate papers from both databases and retrieve the top 200 papers with the highest similarity scores, including 150 from Semantic Scholar and 50 from arXiv or PubMed. The former was determined based on the observation of this quantity to ensure it covers the majority of the references typically included in survey papers. The latter was selected based on our experiments, which showed that the primary database to supplementary database ratio of 2:1 yielded the best results. We restrict all retrieved papers to those published prior to the release date of the target survey to ensure time consistency.

For section-based retrieval, we use section titles as search queries to identify the top-60 most relevant papers and remove any papers without DOI.  Next, through citation network analysis, we identify the top-20 most co-cited papers among the retrieved papers and include them in the candidate pool. The selected papers are then linked to the \textbf{OpenAlex}\footnote{\url{https://openalex.org/}} database via their DOIs to supplement quality-related signals for re-ranking purposes. For each paper, we quantify its academic impact using three indicators: \textbf{(1)} the total number of citations as a measure of citation performance, \textbf{(2)} the venue h-index\footnote{ The h-index is defined as the largest $h$
for which an author or venue has $h$ papers cited at least $h$ times.} as an indicator of venue reputation, and \textbf{(3)} the average h-index of the first and last authors \citep{donthu2021conduct}, which are typically the corresponding authors and have greater influence \citep{CORREAJR2017498}, as a measure of author impact. During the re-ranking stage, we assign weights of 0.5, 0.3, and 0.2 to citation count, author impact, and venue reputation, respectively, based on preliminary experiments that indicate their relative importance in ranking papers. Finally, we retain the top-50 papers per section as the final set of references for content generation. We set the mentioned parameters based on performance testing, with a summary presented in Table \ref{tab:parameters}.

\vspace{4pt}
\textbf{\textbf{(2) Outline generation and survey drafting}}
\vspace{4pt}

For outline generation, we first perform embedding-based clustering over the candidate papers retrieved in the first round to determine the main themes to be included in the outline. We set the number of clusters to 6, as we observed that most published surveys commonly organize their core content into around six major sections, which also aligns with the setting used in prior work \citep{wang2024AutoSurvey,yan2025surveyforgeoutlineheuristicsmemorydriven}. We use the same LLMs as the corresponding baselines for generating the outline and survey. Specifically, we use Claude-3.5-haiku-20241022 for comparisons with AutoSurvey \citep{wang2024AutoSurvey}, SurveyGen \citep{bao-etal-2025-surveygen}, and SurveyForge \citep{yan2025surveyforgeoutlineheuristicsmemorydriven}, and GPT-4o-mini for comparisons with SurveyX \citep{liang2025surveyx}. For each cluster, based on the title/abstracts of the papers within the group, we generate a summary and feed it into the  LLM to generate the initial survey structure. Then, we select the human-written outlines from the 10 most relevant surveys in the SurveyGen \citep{bao-etal-2025-surveygen} dataset as gold standards to perform few-shot learning for optimization, resulting in the final outline. 

For survey drafting, during the information extraction stage,  we build on SurveyX \citep{liang2025surveyx} and use the same knowledge extraction template, which is specifically tailored to reflect the typical structure of surveys in computer science. Similarly, for the biomedical domain, we design a knowledge template based on the widely recognized IMRaD structure (\textit{Introduction}, \textit{Methods}, \textit{Results}, and \textit{Discussion}) \citep{gasparyan2011writing,sollaci2004introduction}, which is commonly used in medical scientific writing. Both knowledge extraction templates are illustrated in Appendix \ref{appendixB}. In addition, we refine the surveys by leveraging peer review comments from published surveys that are related to the target topics. To balance extracting insights from human reviews while maintaining topic consistency, for each topic, we select 20 of the most relevant surveys and use the corresponding peer review comments as the source to extract reviewer checklists, which are then used to generate targeted revision comments for survey revision. To mitigate potential model bias, GPT-4.1-mini is used throughout all data processing and evaluation stages.

\vspace{4pt}
\textbf{\textbf{(3) Agents communication and stopping conditions}}
\vspace{4pt}

We design each agent as an LLM-powered module that handles specific subtasks with crafted prompts. In line with the process of writing a scientific survey, the sequence in which the agents perform their tasks follows the order: \textbf{[RETRIEVER]} → \textbf{[OUTLINER]} → \textbf{[RETRIEVER]} → \textbf{[RANKER]} → \textbf{[WRITER]} → \textbf{[REVIEWER]} → \textbf{[REFINER]}. Here, the two \textbf{[RETRIEVER]} stages serve different purposes. The first retrieval focuses on the survey-level scope, whereas the second targets section-related literature. It is worth noting that this sequence is flexible. For example, when the refinement stage requires additional references or content, the central router can re-invoke the subsequence \textbf{[RETRIEVER] → [RANKER] → [WRITER]} as needed. However, the overall sequence cannot be reversed; for instance, the outline generation cannot be called before any retrieval has occurred. The memory module is designed to store information that is likely to be reused in subsequent steps. This includes: (1) the literature retrieved in the first-round retrieval, which enable efficient reuse for the section-based retrieval round; (2) the initial outline, which serves as a blueprint to guide content generation; (3) the outputs from each step, including agent labels and generated content, which can be referenced in the follow-up steps; (4) execution records that track both current and historical agent activities. We define stop conditions as follows: the process is complete when all agent calls have been executed, and the optimization of the survey cannot exceed \textbf{three} iterations to balance cost and improvement gains.

\section{Results}
\label{sec:results}
In this section, we present the overall comparison results in Section~\ref{sec:51}. Sections~\ref{sec:52}, \ref{sec:53}, and \ref{sec:54} provide detailed analyses of literature retrieval, outline generation, and survey content quality, respectively. Section~\ref{sec:55} reports the ablation study, and Section \ref{sec:56} presents case studies.

\begin{table}[ht]
\centering
\renewcommand{\arraystretch}{1}
\setlength{\tabcolsep}{5pt}  % Adjust column spacing (default is 6pt)
\caption{Overall comparison of model performance; P and R represent precision and recall; H1 and H2 represent the match between the first and second-level headings with the human-written outline; Rel indicates the relevance score (1–5) rated by the LLM.  The best and second-best results are highlighted in \textbf{bold} and \underline{underlined}. }
\begin{tabular}{llccccccccc}
\toprule
\multirow{2}{*}{\textbf{Domain}}  & \multirow{2}{*}{\textbf{Model}}  
& \multicolumn{3}{l}{\textbf{Citation quality}} 
& \multicolumn{3}{l}{\textbf{Structural consistency}} 
& \multicolumn{3}{l}{\textbf{Content quality}} \\
\cmidrule(lr){3-5} \cmidrule(lr){6-8} \cmidrule(lr){9-11} 
& & \textbf{P$\uparrow$} & \textbf{R$\uparrow$} & \textbf{F1$\uparrow$} 
& \textbf{H1(F1)$\uparrow$} & \textbf{H2(F1)$\uparrow$} & \textbf{\color{black}Rel$\uparrow$} 
& \textbf{Similarity$\uparrow$} & \textbf{ROUGE-L$\uparrow$} & \textbf{KPR$\uparrow$} \\
 % Adds horizontal lines below the grouped columns
\midrule
\multirow{5}{*}{CS} 
& Naive-RAG  \citep{lai2024instruct}    & 4.36       & 10.24       & 6.12       & 29.35      & 12.07      & {\color{black}2.5}      & 80.06      & 8.54      & 38.04     \\
& AutoSurvey \citep{wang2024AutoSurvey} & 6.58       & 11.39       & 8.34       & 33.07      & 14.13      & {\color{black}2.7}       & 82.28      & 10.21     & 43.38     \\

& SurveyX \citep{liang2025surveyx}     & 9.44       & 15.15     & 11.63      & \underline{36.84}     & \underline{20.96}      & \textbf{\color{black}3.8}       & 83.11      & 13.72     & \underline{50.37}    \\

& SurveyForge \citep{yan2025surveyforgeoutlineheuristicsmemorydriven} 
                                        & \underline{11.57}     & \underline{17.62}      & \underline{13.97}      & 33.30     & 10.57      & {\color{black}3.0}        & 82.14      & 12.49     & 48.19     \\
& DeepResearch \citep{deepresearch2025}   & 6.72      & 14.78      & 9.24       & 34.37    & 15.12      & \color{black}3.2        & \textbf{83.75}      & \textbf{16.29}     & 45.63    \\
& \textbf{SurveyAgent-HKA}                                  & \textbf{14.95}      & \textbf{19.53}      & \textbf{16.94}      & \textbf{38.43}      & \textbf{22.17}      & \underline{\color{black}3.5}        & \underline{83.49}      & \underline{14.68 }    & \textbf{53.85}     \\
\midrule
\multirow{3}{*}{Med} 
& Naive-RAG \citep{lai2024instruct}    & 8.27       & 12.90      & 10.08              & 22.12      & 11.04      & \color{black}3.2       & 82.93       & 14.37     & 46.52     \\
& SurveyGen \citep{bao-etal-2025-surveygen}  & \underline{15.31} & \underline{19.82} & \underline{17.28}     & 25.07      & 13.68      & \underline{\color{black}3.4} & 83.42    & 15.25 & \underline{53.90} \\
& DeepResearch \citep{deepresearch2025}      & 11.15      & 16.04      & 13.16              & \underline{27.25} & \underline{18.72} & \color{black}3.0       & \textbf{85.07}     & \textbf{17.40} & 51.81 \\
& \textbf{SurveyAgent-HKA}               & \textbf{16.78} & \textbf{22.95} & \textbf{19.39}  & \textbf{30.22} & \textbf{19.26} & \textbf{\color{black} 3.9} & \underline{83.71 }            & \underline{15.96 }         & \textbf{57.07}  \\
\bottomrule
\label{tab:6}
\end{tabular}
\end{table}

\subsection{Overall comparison with baselines}
\label{sec:51}
We first report the overall comparison results with the baselines in Table \ref{tab:6}. As shown, our model achieves the best citation quality in both the computer science (14.95\% for Precision, 19.53\% for Recall, and 16.94\% for F1) and medical domains (16.78\% for Precision, 22.95\% for Recall, and 19.39\% for F1). These results are attributed to the fact that, compared to the baselines, our approach has the advantage of integrating multiple large-scale corpus sources such as Semantic Scholar, along with domain-specific databases like arXiv and PubMed, enabling access to a more comprehensive set of literature. Moreover, we designed a re-ranking module based on multiple indicators (e.g., \textit{academic impact}, \textit{recent popularity}, and \textit{content diversity}), which helps identify more representative and influential papers rather than relying solely on semantic similarity, which has been shown to yield higher consistency with human selection preferences. For structural consistency, our model not only achieves the highest structural match with human-written surveys on both domains but also attains the best scores in LLM-based structural relevance evaluation (3.5/5 for computer science and 3.9/5 for medicine). These results indicate that, instead of generating the outline in one-step LLM call, our dynamic outline generation process first clusters the retrieved literature related to the survey topic and produces cluster summaries. This approach allows the model to identify the main research themes within the topic, which provides a foundation for organizing the survey structure and ensures that the outline is comprehensive and logical. In addition, expanding the first-level outline into a second-level outline produces a chained effect, which also aligns with the strategy behind SurveyX \citep{liang2025surveyx} achieving the second-best performance. In terms of content quality, our model achieves higher key-point recall (53.85\% for computer science and 57.07\% for medicine), indicating that it captures the most important findings present in human-written surveys. Additionally, it ranks second only to the commercial model DeepResearch in terms of content overlap, which also outperforms all other baselines. {\color{black}The statistical significance tests for the main evaluation metrics are reported in Appendix~\ref{st}.}

\subsection{Further analysis of citation performance}
\label{sec:52}
\vspace{4pt}
% \text{\textbf{(1) Citation performance with different retrieval strategies}}
\subsubsection{Citation performance with different retrieval strategies}

\vspace{4pt}

\begin{figure}[pos=htbp]
    \centering
    \begin{minipage}{0.48\textwidth}
        \centering
        \includegraphics[width=\linewidth]{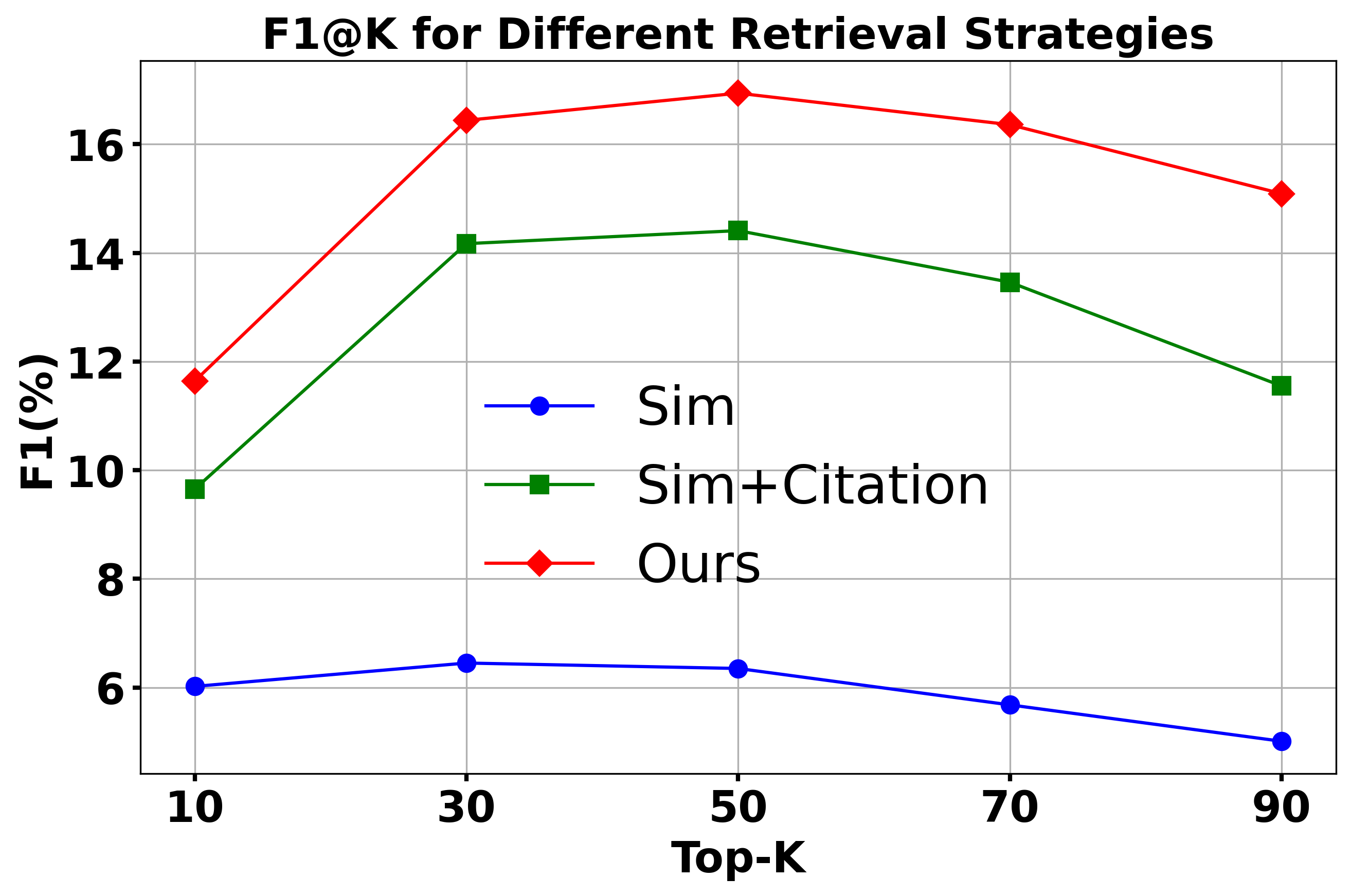} % 你的图像文件名
        \subcaption{Computer Science} \label{fig:3a}
    \end{minipage}%
    \begin{minipage}{0.48\textwidth}
        \centering
        \includegraphics[width=\linewidth]{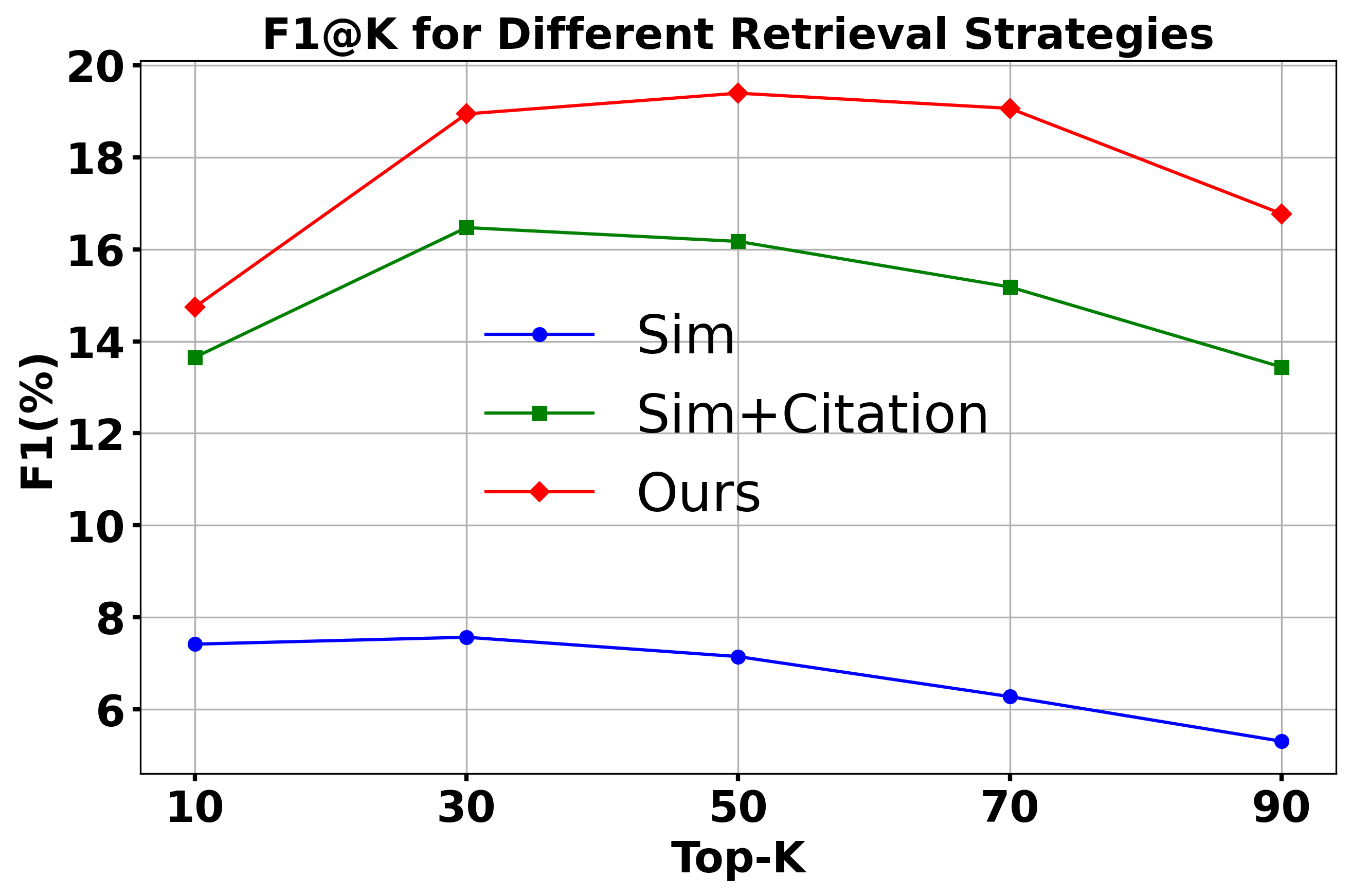} % 你的图像文件名
        \subcaption{Medicine} \label{fig:3b}
    \end{minipage}%
    \caption{Citation performance across different strategies, where K denotes the number of papers retrieved per section. In paper re-ranking, our method considers not only citation counts but also recent popularity, as well as the influence of authors and venues.}
    \label{fig:3}
\end{figure}

We further analyze the impact of different retrieval strategies on the results by examining different Top-K papers for each section. We compare two main approaches: \textbf{(i)} the similarity-only retrieval methods, represented by Naive-RAG \citep{lai2024instruct}, AutoSurvey \citep{wang2024AutoSurvey}, and SurveyX \citep{liang2025surveyx}, and \textbf{(ii)} the combined similarity and citation-based retrieval method like SurveyForge \citep{yan2025surveyforgeoutlineheuristicsmemorydriven}. As illustrated in Figure \ref{fig:3}, in both domains, purely similarity-based retrieval performs the worst, showing minimal performance improvement even as K increases. In contrast, citation-based methods excel at identifying highly cited, influential papers, which are typically regarded as crucial references in survey writing due to their academic significance. Notably, we observe that citation-based methods show clear performance improvements as K increases; however, beyond a certain threshold (K=50), the rate of improvement begins to slow down, as most highly-cited papers have already been included, and additional papers struggle to overcome the cut-off gains. In comparison, our approach consistently outperforms all other methods across a range of K values. This enhanced performance can be attributed to the dual consideration of our retrieval strategy, which not only incorporates citation counts but also accounts for the short-term relevance of papers. Through this multidimensional evaluation,  we are able to capture newly relevant literature that may not yet have accumulated a high citation count but remains valuable for recent topics of focus in the survey. This combination ensures continuous performance enhancement as K increases, setting our approach apart from the others.

\begin{figure}[pos=htbp]
    \centering
    \begin{minipage}{0.48\textwidth}
        \centering
        \includegraphics[width=\linewidth]{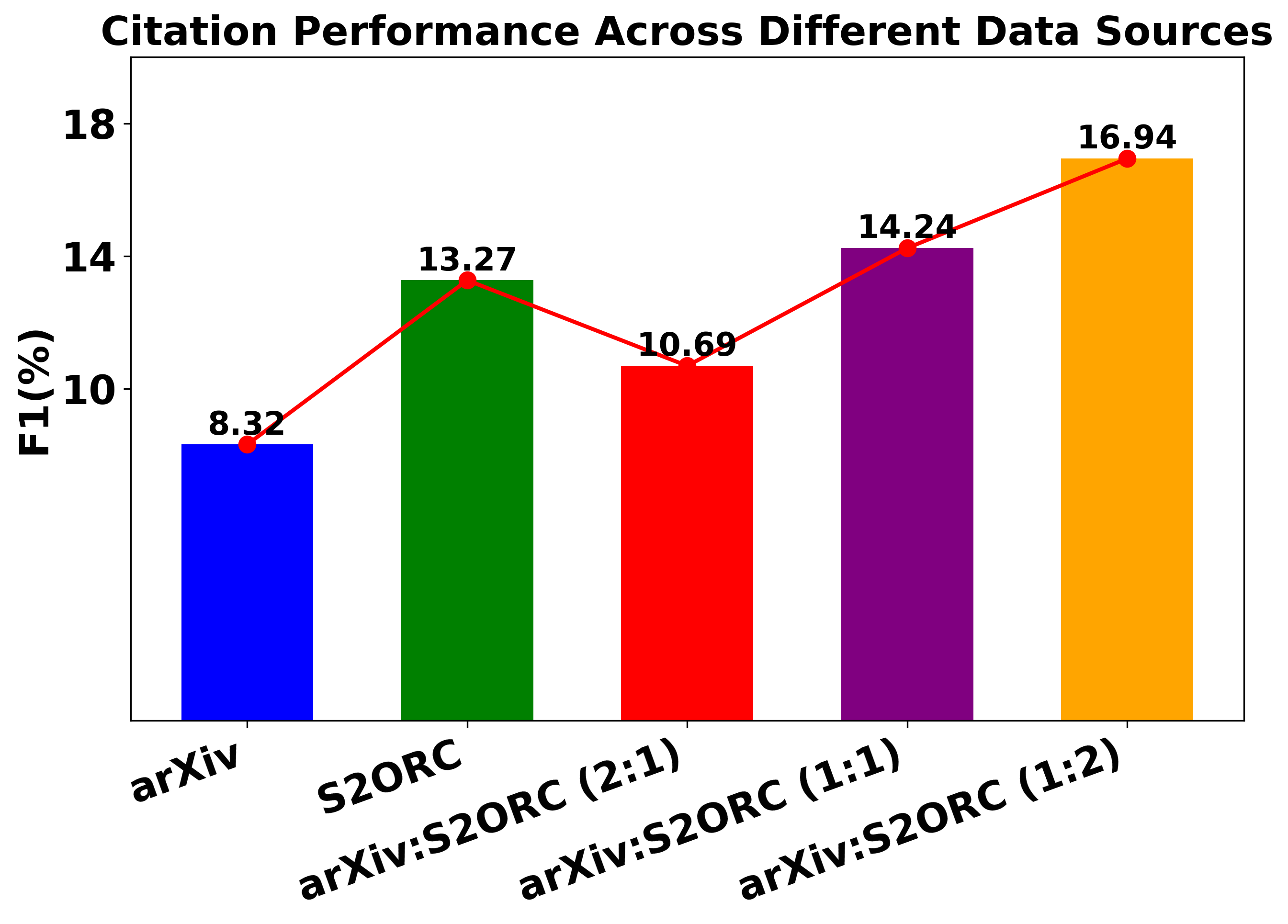} % 你的图像文件名
        \subcaption{Computer Science} \label{fig:4a}
    \end{minipage}%
    \begin{minipage}{0.48\textwidth}
        \centering
        \includegraphics[width=\linewidth]{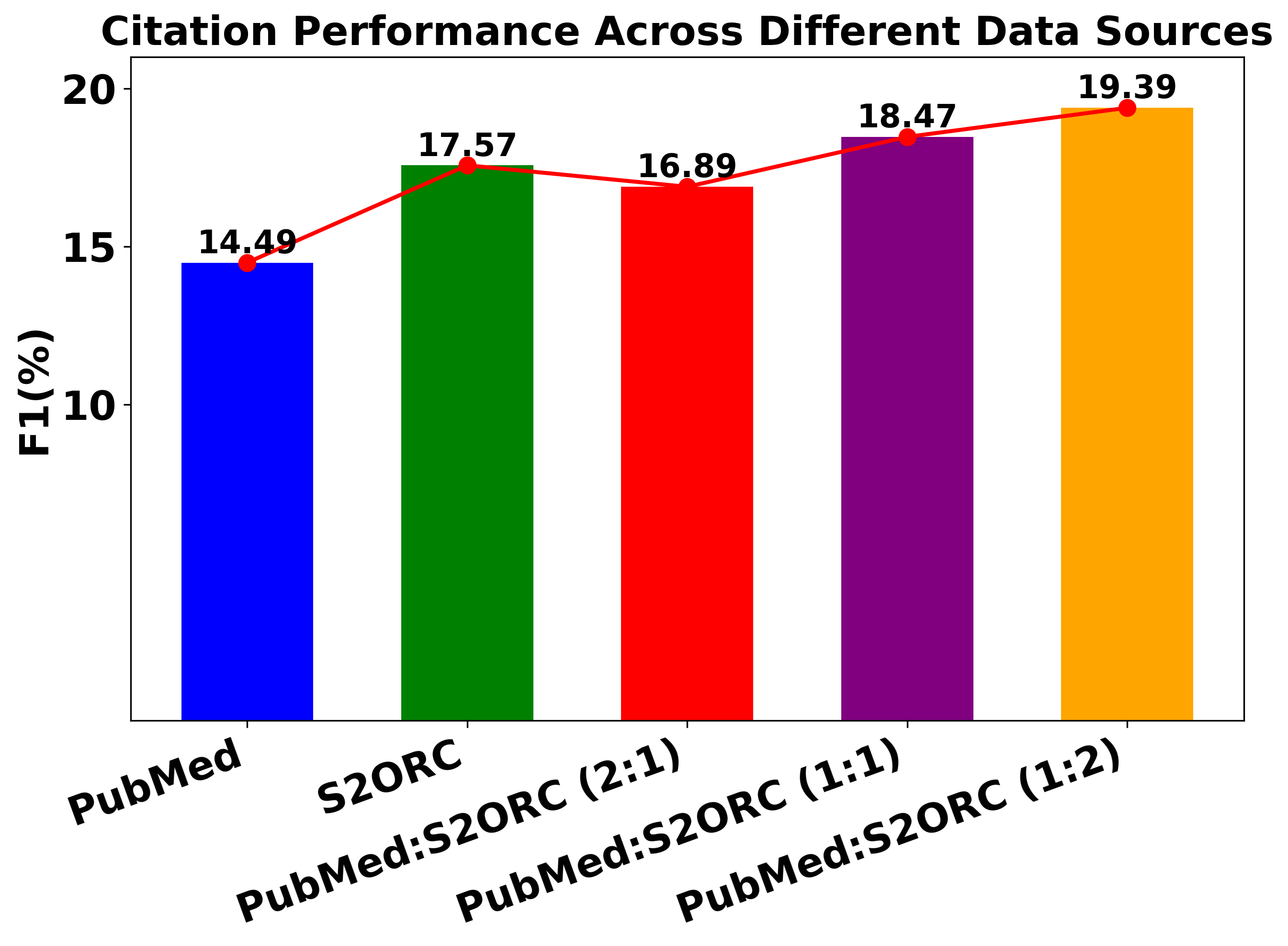} % 你的图像文件名
        \subcaption{Medicine} \label{fig:4b}
    \end{minipage}%
\caption{Impact of different data collection sources on literature retrieval performance, where arXiv and PubMed are used as supplementary data sources for computer science and medicine domains, respectively.}
    \label{fig:4}
\end{figure}

\subsubsection{Citation performance with different data sources}
\vspace{4pt}

In Figure \ref{fig:4}, we report the impact of different data sources on citation quality in literature retrieval. As mentioned previously, we use Semantic Scholar as the primary data source, while arXiv and PubMed serve as specialized databases for the computer science and medical fields, respectively. It can be observed that, for the computer science field, retrieval results from arXiv alone do not perform as well as those from Semantic Scholar. Although arXiv covers cutting-edge research, some papers have not yet undergone thorough peer review, which compromises their reliability, and certain publications that do not have preprints available on arXiv may also further contribute to this issue. For the medical field, the retrieval results of PubMed are also lower than those from Semantic Scholar, but the gap is smaller compared to the computer science field. This is mainly because PubMed is widely regarded as the authoritative database in the medical field, containing a large number of high-quality, rigorously peer-reviewed papers. Although Semantic Scholar has broader coverage, its advantage over PubMed in the medical domain is limited compared to the computer science field, but it remains useful for cross-source retrieval. In general, our findings indicate that the use of multiple data sources for literature retrieval can improve both search accuracy and recall, offering advantages over the use of a single database. In practice, the primary database should be given more weight, while secondary databases can play a supporting role, helping to improve the recall of relevant literature and enhance its overall quality.

\begin{figure}[pos=htbp]
    \centering
    % 第一行
    \begin{minipage}{0.49\textwidth}
        \centering
        \includegraphics[width=\linewidth]{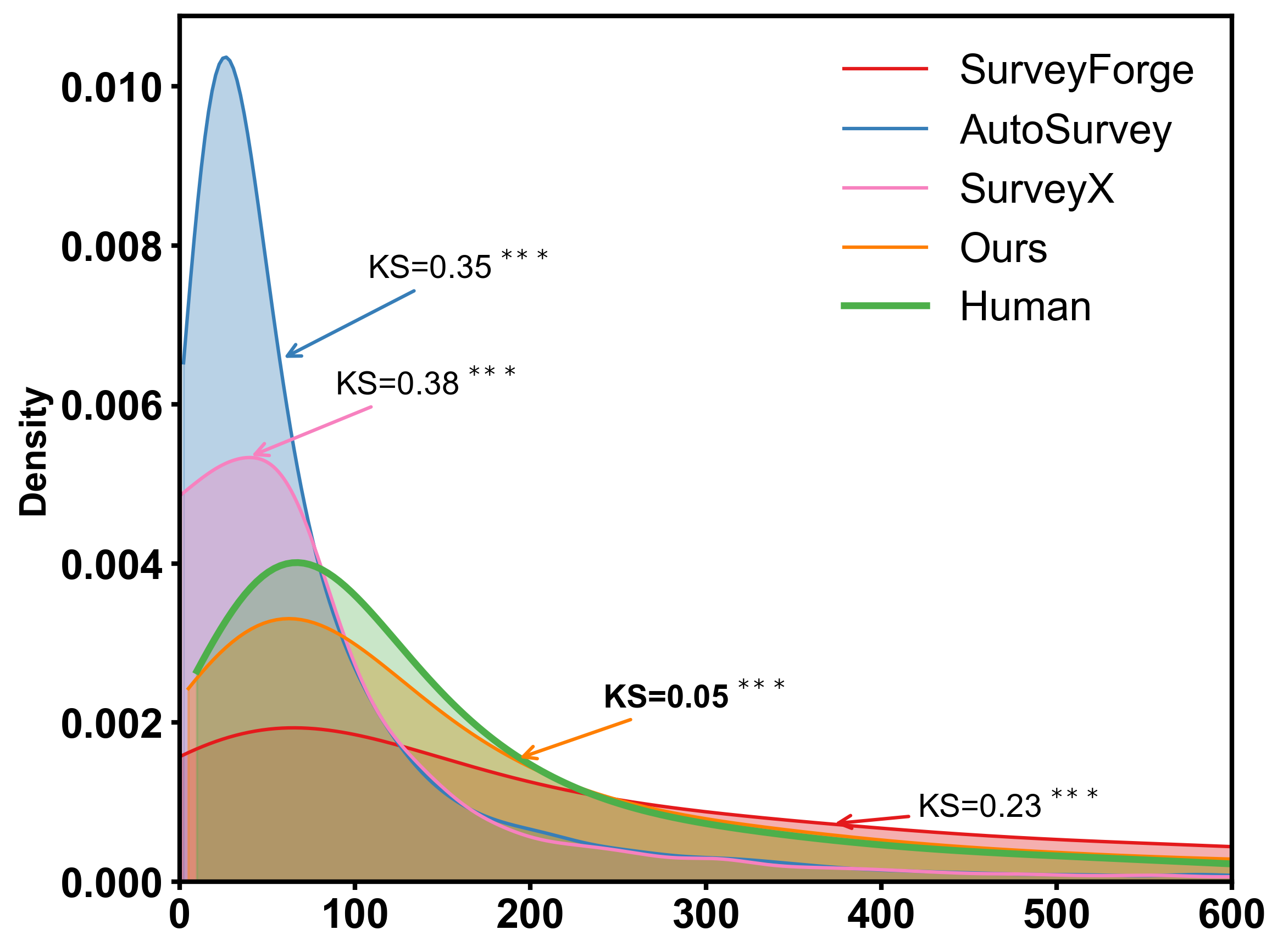}
        \subcaption{Citation counts} \label{fig:5a}
    \end{minipage}%
    \hspace{0.01\textwidth}% 调整行内间距
    \begin{minipage}{0.49\textwidth}
        \centering
        \includegraphics[width=\linewidth]{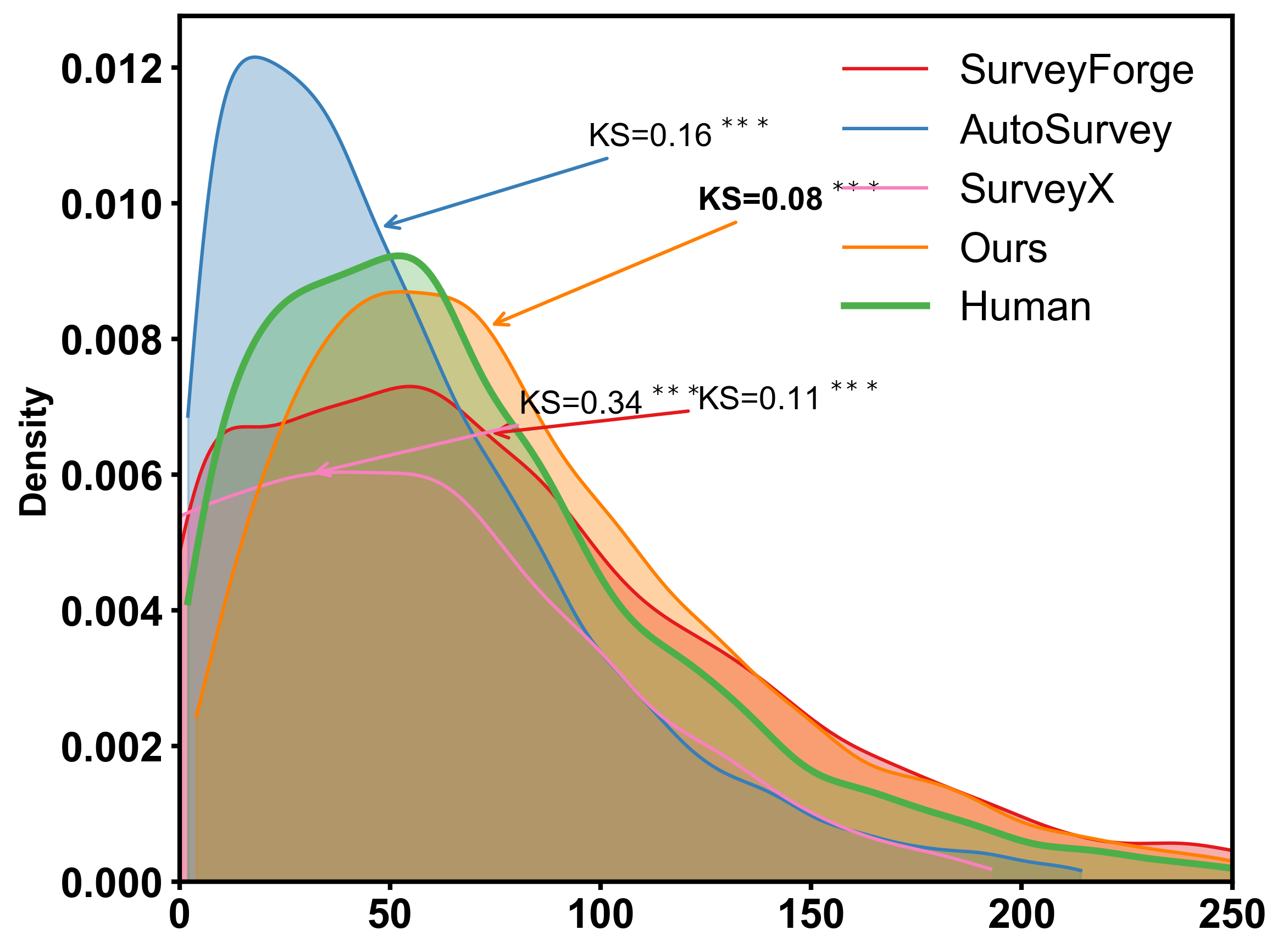}
        \subcaption{Author H-index} \label{fig:5b}
    \end{minipage}
    % 第二行
    \begin{minipage}{0.49\textwidth}
        \centering
        \includegraphics[width=\linewidth]{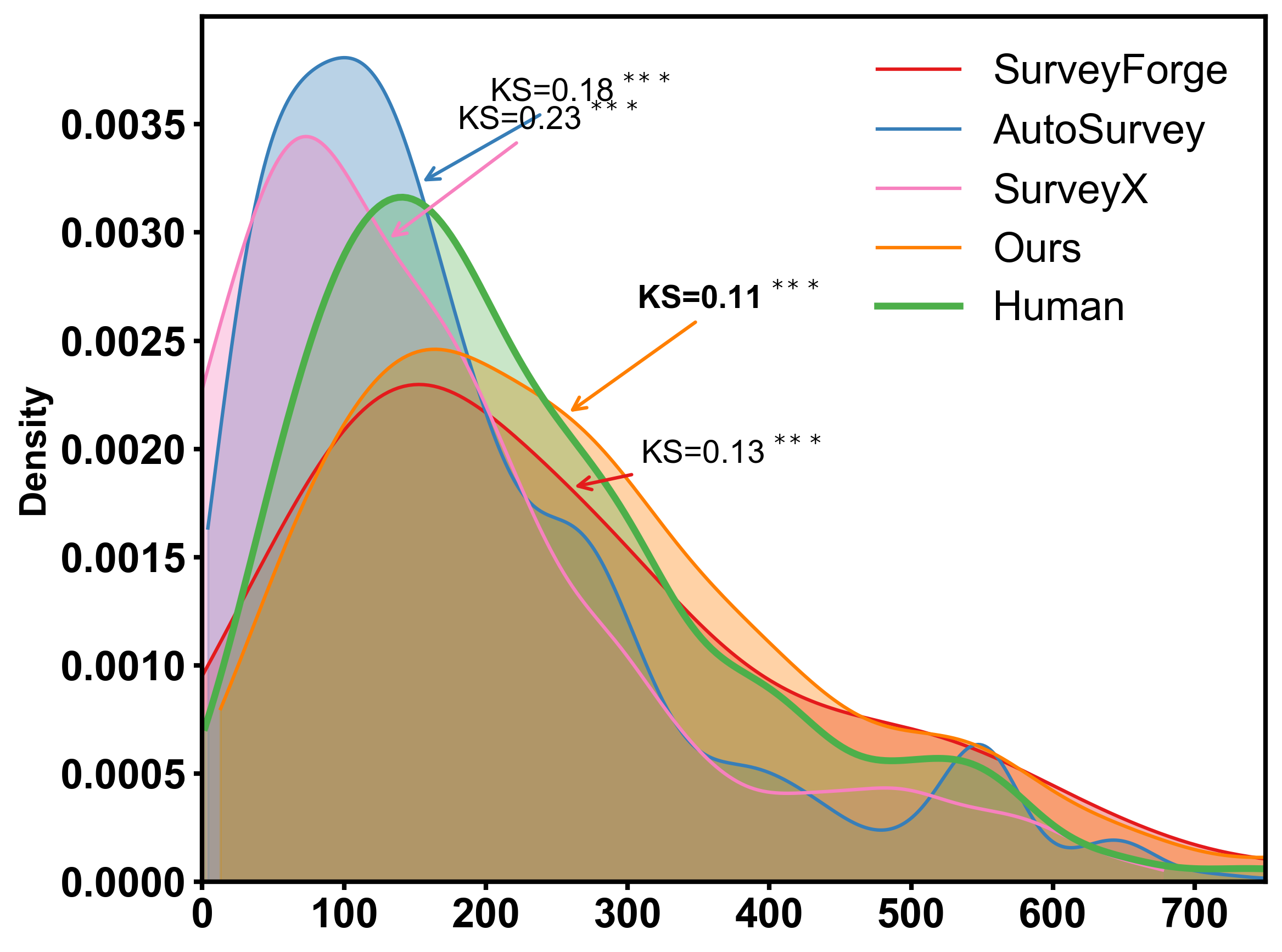}
        \subcaption{Journal H-index Sum} \label{fig:5c}
    \end{minipage}%
    \hspace{0.01\textwidth}
    \begin{minipage}{0.49\textwidth}
        \centering
        \includegraphics[width=\linewidth]{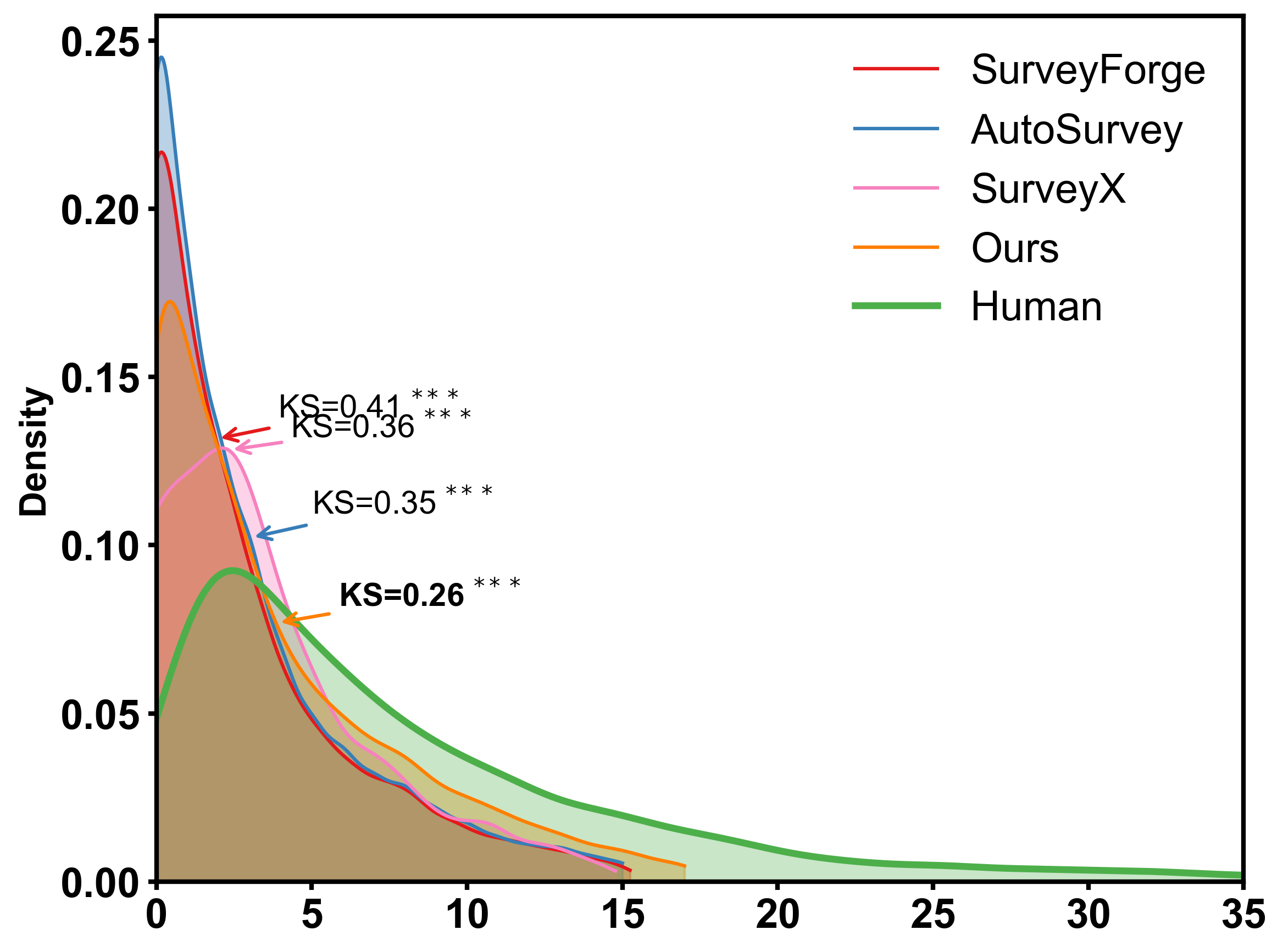}
        \subcaption{Time Gap} \label{fig:5d}
    \end{minipage}
    \caption{Distribution of retrieved references compared to ground truth, where the KS represents the Kolmogorov-Smirnov test, and a smaller KS value indicates a closer distribution to human-selected references. *** denotes a \textit{p}-value < 0.001. The time gap refers to the interval between the publication date of the retrieved reference and the corresponding survey.}
    \label{fig:5}
    \vspace{-10pt}
\end{figure}

\vspace{4pt}
\subsubsection{Distribution analysis of the retrieved references}
\vspace{4pt}

Taking computer science domain as an example, we investigate how the retrieved references are distributed across different models and compare them with the ground truth references in Figure \ref{fig:5}, focusing on the following four aspects: citation count (\ref{fig:5a}), author H-index (\ref{fig:5b}), journal H-index (\ref{fig:5c}), and time gap (\ref{fig:5d}).  As shown, our method outperforms the baseline models and aligns more closely with the ground truth distribution across all four dimensions, as evidenced by the lower KS values obtained.  Specifically, in terms of citation count, similarity-based models such as AutoSurvey \citep{wang2024AutoSurvey} and SurveyX \citep {liang2025surveyx} show a higher density in retrieving low-citation papers, which is also consistent with the distribution of these papers' publication years (\ref{fig:5d}), with more recent publications in arXiv being retrieved. Therefore, these papers tend to have lower citation counts, as they have not yet had time to accumulate higher citations. However, our model, which takes both citation count and recent popularity into the literature selection process, is more effective in identifying influential papers that have accumulated citations over time, while also including the latest research on hot topics. For author and journal impact, all models similarly reflect a consistent long-tail distribution, where a small number of influential authors and journals are heavily cited, while a larger proportion of less-cited works still contribute to the overall academic landscape. Overall, although all methods show some alignment with the human ground truth distribution, they still show significant differences compared to the references selected by humans. For example, humans writing surveys may choose papers that were published up to 30 years ago, but these papers may not be included in mainstream databases due to issues like the lack of a digital version or limited accessibility. Additionally, the significant differences observed across all models highlight the challenge of the task, as human selection behavior is a complex process and difficult to replicate.

\subsection{Further analysis of the generated survey outline}
\label{sec:53}

\begin{table}[htbp]
\centering
\caption{Performance comparison of generated outlines, where H1 denotes the number of first-level headings and H2 denotes the number of second-level headings. The best and second-best results are highlighted in \textbf{bold} and \underline{underlined}.}
\label{tab7}
\begin{tabular}{lcccccc}
\toprule
\textbf{Model} & \textbf{\makecell{\#H1\\LLM}} & \textbf{\makecell{\#H1\\Human}} & \textbf{\makecell{H1-Match\\(P/R/F1\%)}} & \textbf{\makecell{\#H2\\LLM}} & \textbf{\makecell{\#H2\\Human}} & \textbf{\makecell{H2-Match\\(P/R/F1\%)}} \\
\midrule
\multicolumn{7}{c}{\textbf{Computer Science}} \\
\midrule
Naive-RAG \citep{lai2024instruct} 
& 5.0 
& \multirow{6}{*}{7.4} 
& 36.40/24.59/29.35
& 17.6
& \multirow{6}{*}{18.2} 
& 25.59/24.73/25.14 \\
AutoSurvey \citep{wang2024AutoSurvey} 
& 5.3 
&  
& \underline{39.62}/28.38/33.07 
& 20.7
&  
& 27.54/31.32/29.31 \\ 
SurveyX \citep{liang2025surveyx} 
& 7.8 
&  
& 35.90/\textbf{37.85}/\underline{36.84}
& 14.6 
&  
& \textbf{36.61}/28.57/\underline{31.71}\\ 
SurveyForge \citep{yan2025surveyforgeoutlineheuristicsmemorydriven} 
& 8.2 
&  
& 31.71/35.06/33.30 
& 33.8
&  
& 22.78/\textbf{42.30}/29.62 \\
DeepResearch \citep{deepresearch2025} 
& 5.4 
&  
& \textbf{40.74}/29.73/34.37 
& 13.6
&  
& 27.21/25.33/23.27 \\
\textbf{SurveyAgent-HKA (Ours)} 
& 7.0
&  
& 39.57/\underline{37.35}/\textbf{38.43}
& 20.1
&  
& \underline{31.84}/\underline{35.17}/\textbf{33.42} \\
\midrule
\multicolumn{7}{c}{\textbf{Medicine}} \\
\midrule
Naive-RAG \citep{lai2024instruct}
& 5.2 
& \multirow{4}{*}{5.6} 
& 36.54/33.93/35.19 
& 16.3
& \multirow{4}{*}{15.6} 
& 26.99/28.20/27.59 \\
SurveyGen \citep{bao-etal-2025-surveygen}
& 6.5 
&  
& 35.39/41.07/38.02
& 17.5
&  
& 30.27/\underline{33.97}/\underline{32.02} \\
DeepResearch \citep{deepresearch2025} 
& 4.9
&  
& 40.82/35.71/38.10
& 12.7 
&  
& \textbf{33.89}/27.54/30.39 \\
\textbf{SurveyAgent-HKA (Ours)} 
& 6.3
&  
& \textbf{42.86}/\textbf{48.21}/\textbf{45.38}
& 16.9 
&  
& \underline{32.54}/\textbf{35.26}/\textbf{33.85} \\
\bottomrule
\label{tab:9}
\end{tabular}
\vspace{-10pt}
\end{table}

\vspace{4pt}
\subsubsection{Statistical analysis of the generated survey outline}
\vspace{4pt}

We first analyze the consistency between the numbers of survey outlines generated by different models and those written by humans, as summarized in Table \ref{tab:9}. As can be observed, in the computer science domain, the number of generated outlines varies notably across models. Specifically, the number of generated first-level sections normally ranges from 5 to 8, while the number of second-level subsections ranges from 13 to 20, which is comparable to the number of sections observed in human-generated outlines. Our model achieves the highest F1 match scores for both outline levels, with 38.53\% for first-level sections (H1) matching and 33.42\% for second-level (H2) matching, suggesting a better balance between coverage and precision when aligning with human-written outlines. In the medical domain, our model also demonstrates superior performance, achieving F1 match scores of 45.38\% for H1 matching and 33.85\% for H2 matching. In addition, when a model generates a smaller number of outline headings, such as DeepResearch~\citep{deepresearch2025}, it tends to achieve higher precision; however, it also results in limited coverage of human-written content, leading to lower recall and consequently lower F1 scores.  Across both domains,  generating second-level subsections is more challenging than first-level sections because the latter require capturing finer thematic distinctions or task-specific categorizations, while first-level sections typically correspond to high-level rhetorical functions shared across surveys (e.g., \textit{Introduction} and \textit{Conclusion}).  Furthermore, outlines in computer science tend to be more difficult to generate than those in medicine. A key contributing factor is the stronger degree of structural standardization in medical literature, where surveys frequently conform to well-established organizational paradigms—most notably the IMRaD format \citep{sollaci2004introduction, kumar2023improvingScientific,peppas2000physicochemical}, which facilitates more accurate outline generation by LLMs. Additionally, the widespread use of biomedical corpora in LLM training may also contribute to this phenomenon \citep{liu2025datasets}.

\begin{figure}[pos=htbp]
    \centering
    \begin{minipage}{0.5\textwidth}
        \centering
        \includegraphics[width=\linewidth]{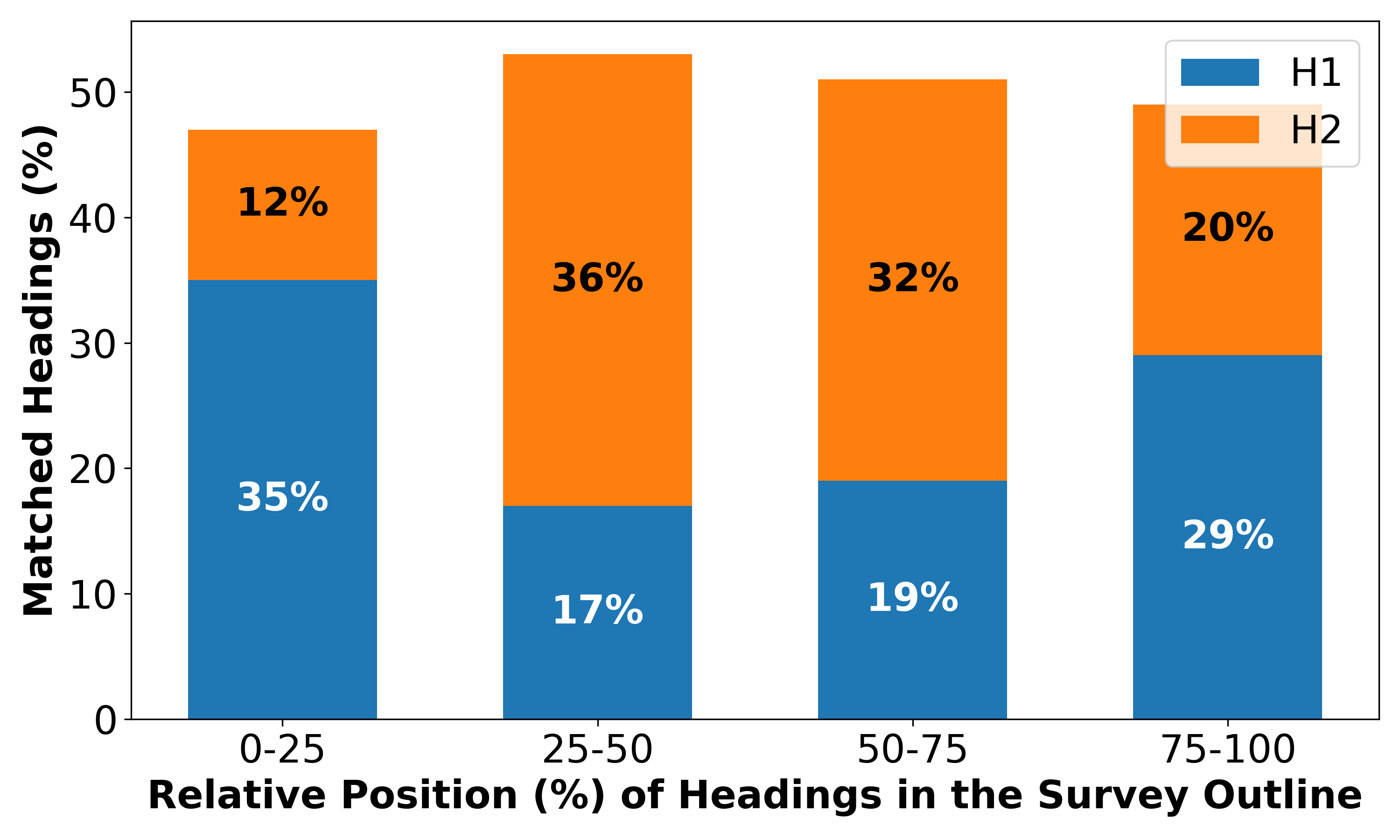}
        \subcaption{Computer Science} \label{fig:6a}
    \end{minipage}%
    \begin{minipage}{0.5\textwidth}
        \centering
        \includegraphics[width=\linewidth]{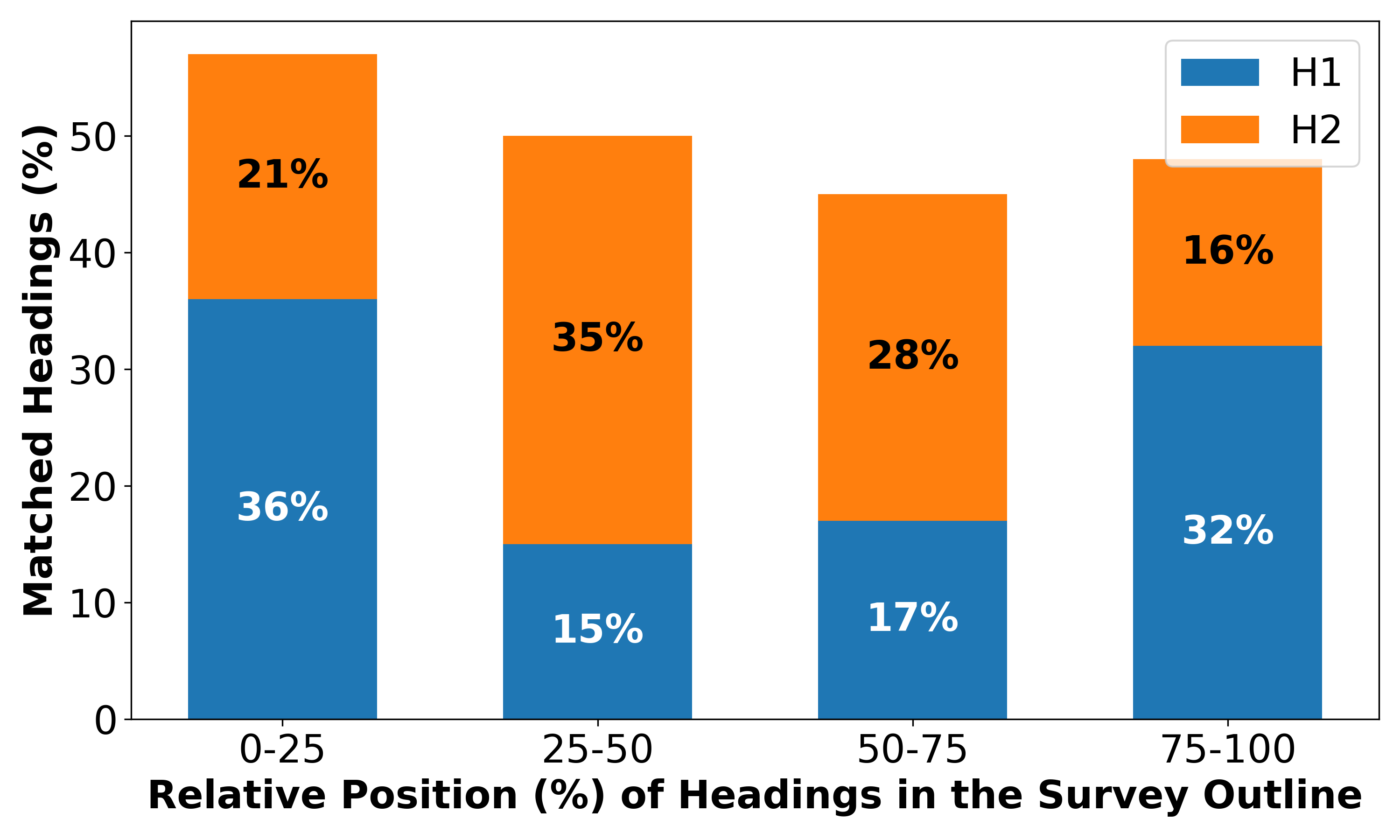} 
        \subcaption{Medicine} \label{fig:6b}
    \end{minipage}%
    \caption{Position distribution of matched section and subsection headings in the generated survey.}
    \label{fig:6}
\end{figure}

\begin{figure}[pos=htbp]
    \centering
    \begin{minipage}{0.5\textwidth}
        \centering
        \includegraphics[width=\linewidth]{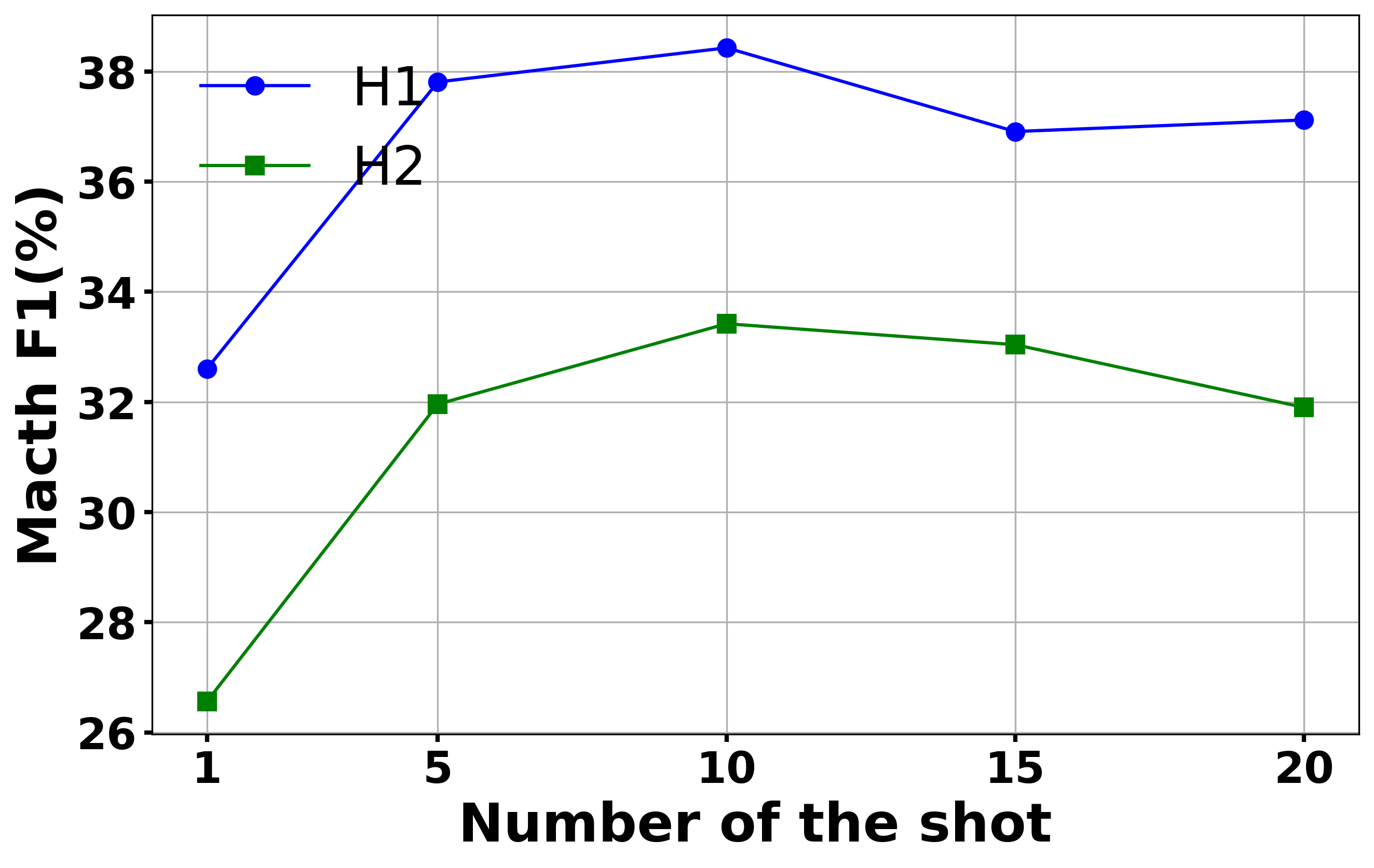}
        \subcaption{Computer Science} \label{fig:7a}
    \end{minipage}%
    \begin{minipage}{0.5\textwidth}
        \centering
        \includegraphics[width=\linewidth]{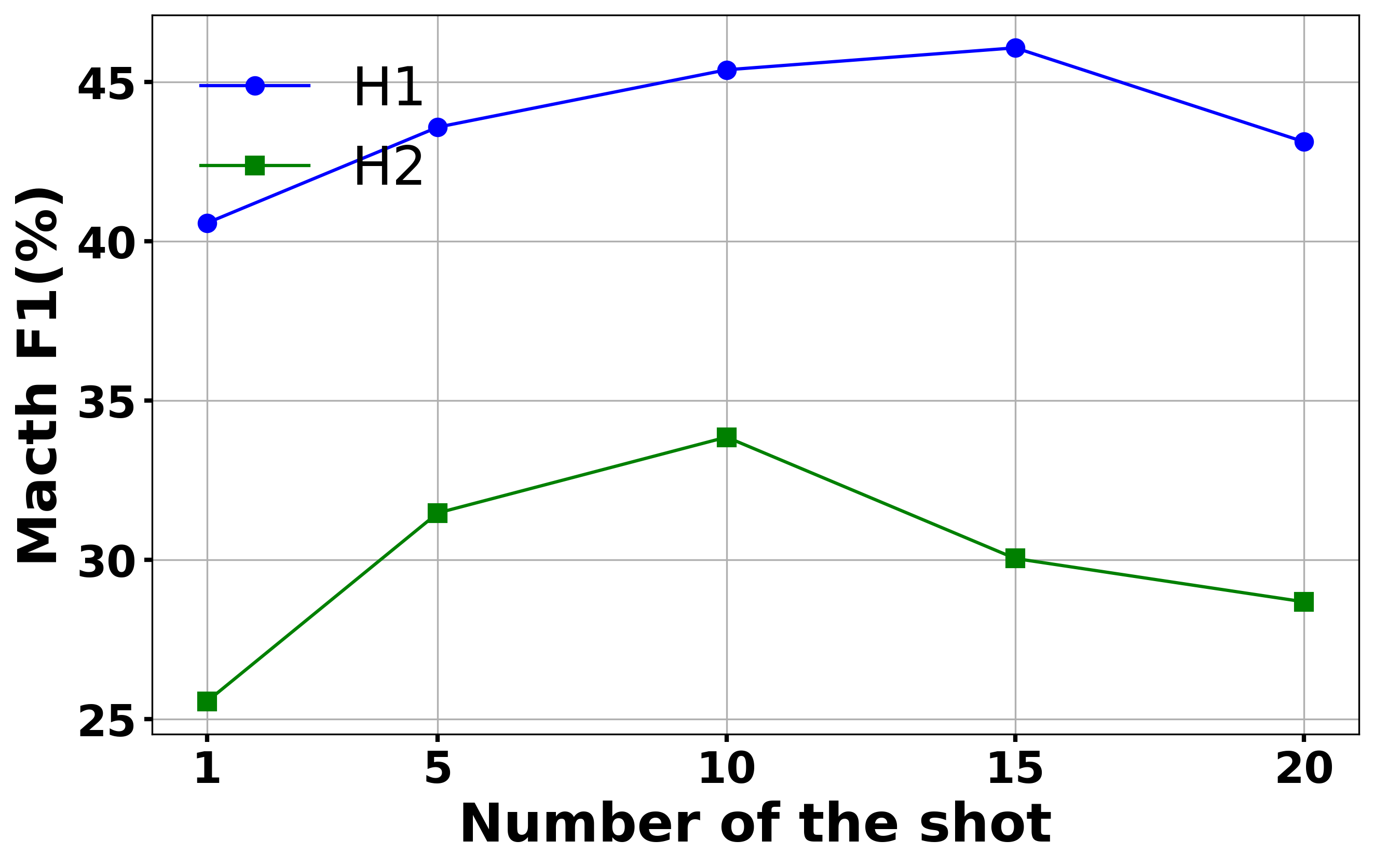} 
        \subcaption{Medicine} \label{fig:7b}
    \end{minipage}%
\caption{Structural consistency performance under different shot settings, where shot outlines are selected from related human-written surveys based on topical similarity.}
    \label{fig:7}   
\end{figure}
\vspace{-10pt}

\subsubsection{Distribution analysis of the matched section}
We further examined the distribution of matched headings that exceeded the predefined semantic similarity threshold. Specifically, we divided the relative position of the generated section within a survey into four equal segments: the beginning (0-25\%), the early middle (25-50\%), the late middle (50-75\%), and the ending section (75-100\%). As shown in Figure~\ref{fig:6}, most of the matched first-level sections (H1) generated by LLMs were clustered at the beginning and the end of the survey. This observation further validates our conclusion that first-level headings are generally easier for LLMs to generate, since the beginning and ending of surveys typically contain standardized sections, such as \textit{``Introduction''} and \textit{``Conclusion''}, that are largely shared across disciplines and thus more readily generated by LLMs. Moreover, H2 matches are more commonly observed in the middle sections, which tend to include more subtopics to provide detailed overview.

\vspace{4pt}
\subsubsection{Structural performance under different shot settings}
\vspace{4pt}

In the outline generation process, we provide LLMs with published human-written outlines on topics closely related to the target survey for few-shot learning. As shown in Figure \ref{fig:7}, as the few-shot examples increase, the performance of both H1 and H2 generation improves consistently cross both domains, with more pronounced improvements observed in computer science. The optimal performance peaks when the number of few-shot examples is set to 10. Although LLMs present limited controllability in text generation, this result indicates that human-written outlines offer effective structural guidance, enabling LLMs to better capture high-level organizational patterns as well as finer-grained hierarchical relationships. However, it is worth noting that when the number of few-shot examples exceeds a certain threshold, performance begins to decline, as additional examples may introduce irrelevant or redundant structural cues, which in turn reduce the model's capacity to identify the structural patterns needed for effective outline generation.

\subsection{Further analysis of the generated survey content}
\label{sec:54}

\vspace{4pt}
\subsubsection{Analysis of peer review comments on survey revision}
\vspace{4pt}

To examine the impact of peer review comments on survey generation, we use the initial draft without peer review as the baseline ($v_0$) and compare it with the final version refined by our SurveyAgent-HKA after generating a revision list from peer review comments of related surveys and applying the corresponding modifications. The comparison is conducted across three dimensions: citation quality, structural consistency, and content quality. As shown in Table \ref{tab:8},  the refined version shows improvements over the initial draft across all three dimensions. In the computer science domain, the citation quality F1 score increased by
3.86\%, with first- and second-level heading match F1 scores increased by 9.31\% and 6.19\%, respectively, while ROUGE and KPR scores increased by 2.81\% and 10.25\%.  In the medical domain, citation quality increased by 2.62\%, while first- and second-level heading matches improved by 4.92\% and 6.71\%, respectively, and ROUGE and KPR scores increased by 4.22\% and 8.48\%. These results indicate that after simulating the peer review process, when the model incorporates these simulated review suggestions, it reorganizes sections and adjusts the hierarchy of content, resulting in a more coherent and logically structured survey. In addition, such structural refinements and the supplementation of references can trigger additional literature retrieval, allowing the model to identify and add relevant studies that were missed in the prior round, further elevating the completeness of the generated survey.

\begin{table}[ht]
\centering
\renewcommand{\arraystretch}{1}
\setlength{\tabcolsep}{6pt}
\caption{Comparison of survey draft (V0) and final version in citation quality, structural consistency, and content quality. The best and second-best results are highlighted in \textbf{bold} and \underline{underlined.}}
\begin{tabular}{llccccccccc}
\toprule
\multirow{2}{*}{\textbf{Domain}}  & \multirow{2}{*}{\textbf{Model}}  
& \multicolumn{3}{l}{\textbf{Citation quality}} 
& \multicolumn{3}{l}{\textbf{Structural consistency}} 
& \multicolumn{3}{l}{\textbf{Content quality}} \\
\cmidrule(lr){3-5} \cmidrule(lr){6-8} \cmidrule(lr){9-11}
& & \textbf{P$\uparrow$} & \textbf{R$\uparrow$} & \textbf{F1$\uparrow$} 
& \textbf{H1(F1)$\uparrow$} & \textbf{H2(F1)$\uparrow$} & \textbf{\color{black}Rel$\uparrow$} 
& \textbf{Similarity$\uparrow$} & \textbf{ROUGE-L$\uparrow$} & \textbf{KPR$\uparrow$} \\
\midrule
\multirow{2}{*}{CS} 
& Survey$_{\rm v_0}$    
& 11.72 & 14.79 & 13.08
& 29.12 & 15.98 & \color{black} 2.9
& 83.04 & 11.87 & 43.60 \\
& Survey$_{\rm final}$                                    
& \textbf{14.95} & \textbf{19.53} & \textbf{16.94} 
& \textbf{38.43} & \textbf{22.17} & \textbf{\color{black}3.5} 
& \textbf{83.49} & \textbf{14.68} & \textbf{53.85} \\
\midrule
\multirow{2}{*}{Med} 
& Survey$_{\rm v_0}$    
& 14.61 & 19.69 & 16.77
& 25.30 & 12.55 & \color{black} 3.6
& 81.92 & 11.74 & 48.59 \\
& Survey$_{\rm final}$      
& \textbf{16.78} & \textbf{22.95} & \textbf{19.39}
& \textbf{30.22} & \textbf{19.26} & \textbf{\color{black}3.9}
& \textbf{83.71} & \textbf{15.96} & \textbf{57.07} \\
\bottomrule
\label{tab:8}
\end{tabular}
\vspace{-10pt}
\end{table}

\begin{table}[ht]
\centering
\renewcommand{\arraystretch}{1}
\setlength{\tabcolsep}{6pt}
\caption{Comparison of survey draft and refined version (in computer science domain) in key metrics: citation quality (F1), structural consistency (H1-, H2-F1), and content quality (KPR).  The best and second-best results are highlighted in \textbf{bold} and \underline{underlined}.}
\begin{tabular}{lcccc}
\toprule
\textbf{Versions} 
& \textbf{Citation F1$\uparrow$} 
& \textbf{H1(F1)$\uparrow$} & \textbf{H2(F1)$\uparrow$} 
& \textbf{KPR$\uparrow$} \\
\midrule
V0 
& 13.08 & 29.12 & 15.98 & 43.60 \\
V1                                  
& 14.36 & 33.85 & 19.17 & 48.15 \\
V2    
& \textbf{17.25} & \underline{36.21} & 20.55 & 50.34 \\
V3     
& \underline{16.94} & \textbf{38.43} & \textbf{22.17} & \textbf{53.85} \\
V4       
& 16.12 & 32.29 & \underline{21.73} & \underline{51.56} \\
\bottomrule
\label{tab:11}
\vspace{-10pt}
\end{tabular}
\end{table}

\subsubsection{Analysis of the revision stopping rounds}
Taking the computer science domain as an example, we further analyze how the stopping condition for revision rounds impacts generated survey performance. As shown in Table \ref{tab:11}, the initial draft (V0) starts with relatively low scores across all metrics, with a citation F1 of 13.08\%, H1 (F1) of 29.12\%, H2 (F1) of 15.98\%, and KPR of 43.60\%. Early revisions (V1-V2) contribute notable gains across multiple dimensions, and the third revision (V3) achieves the highest overall performance, indicating that iterative refinement effectively enhances the survey's quality and coherence. Beyond this point, additional revisions yield only marginal improvements, and the performance tends to converge.  One possible explanation is that, as the survey becomes more polished, the model has fewer substantial errors to correct, and excessive iterations primarily reinforce existing content rather than introduce meaningful improvements. Additionally, extra revision rounds incur higher token usage and time costs. Therefore, we set the stopping condition at three iterations to balance performance gains and computational efficiency.

\vspace{4pt}
\subsection{\textcolor{black}{Parameter sensitivity analysis}}
\vspace{4pt}
{\color{black}We perform a parameter sensitivity analysis to justify the parameter choices, including the ranking weights, which determine the priority of candidate references during re-ranking; the number of clusters, which affects the structural organization of the survey; and the similarity thresholds used to match reference titles and section headings during evaluation. As shown in Figure  \ref{fig:Parameter}, in the citation re-ranking module, assigning a larger weight to citation performance leads to better results than emphasizing author impact or venue influence. This suggests that citation-based indicators provide a more direct signal for identifying influential and representative references. For topic clustering, KPR generally increases with the number of clusters, as more clusters enable the system to cover a broader range of subtopics. However, when the number of clusters exceeds 6, the improvement in KPR becomes marginal, while the additional retrieval costs increase. Therefore, following prior studies, we set the number of clusters to 6 in the main experiments. For reference matching, the results indicate that a similarity threshold of 0.95 yields the highest title-matching accuracy. In contrast, when the threshold is set to 1.00, the matching rate drops noticeably. Through manual inspection, we find that exact matching is highly sensitive to minor variations, such as spelling differences, emojis, or formatting inconsistencies. For title matching, a threshold of 0.85 provides an optimal trade-off between correctly identifying matched titles and avoiding missed matches, leading to the highest F1 score.

\begin{figure}[pos=htbp]
    \centering
    \includegraphics[width=\linewidth]{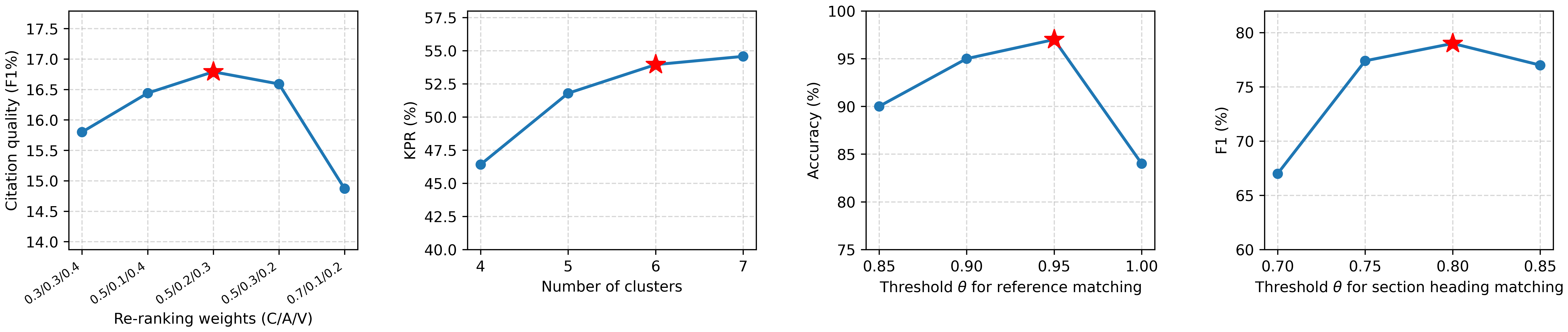}
    \caption{{\color{black}Parameter sensitivity analysis. In the re-ranking module, $C$, $A$, and $V$ denote citation performance, author impact, and venue influence, respectively. The results of reference matching and section heading matching are measured based on human evaluation of 200 matched samples.}}
    \label{fig:Parameter}
    \vspace{-10pt}
\end{figure}}

\subsection{Human evaluation results}
\subsubsection{Comparison across LLM-generated and Human-written surveys}

\begin{figure}[pos=htbp]
    \centering
    \begin{minipage}{0.49\textwidth}
        \centering
        \includegraphics[width=\linewidth]{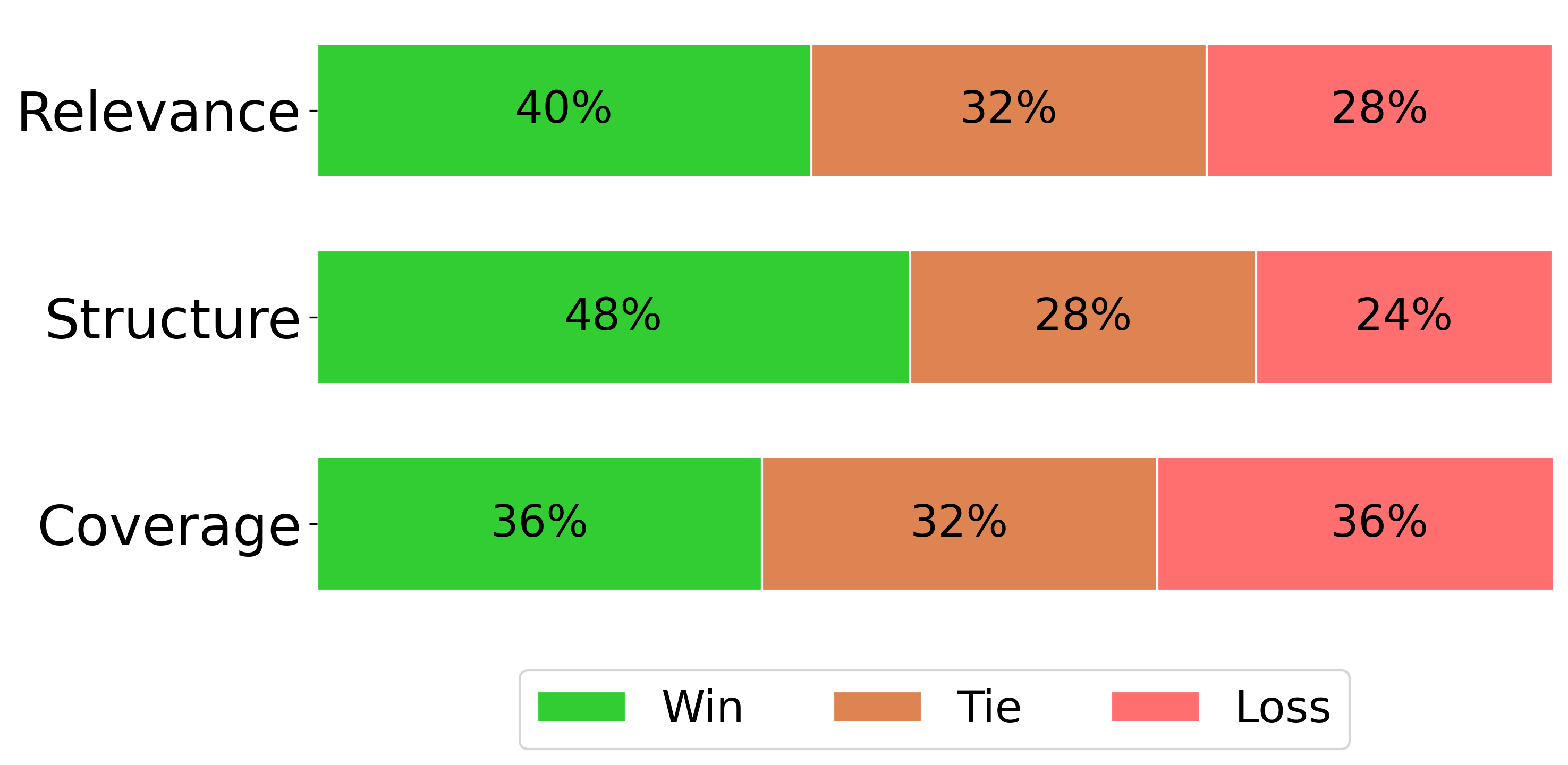} 
        \subcaption{SurveyAgent-HKA (Final) \textit{\textbf{VS}} SurveyAgent-HKA (V0)} \label{fig:8a}
    \end{minipage}%
    \begin{minipage}{0.49\textwidth}
        \centering
        \includegraphics[width=\linewidth]{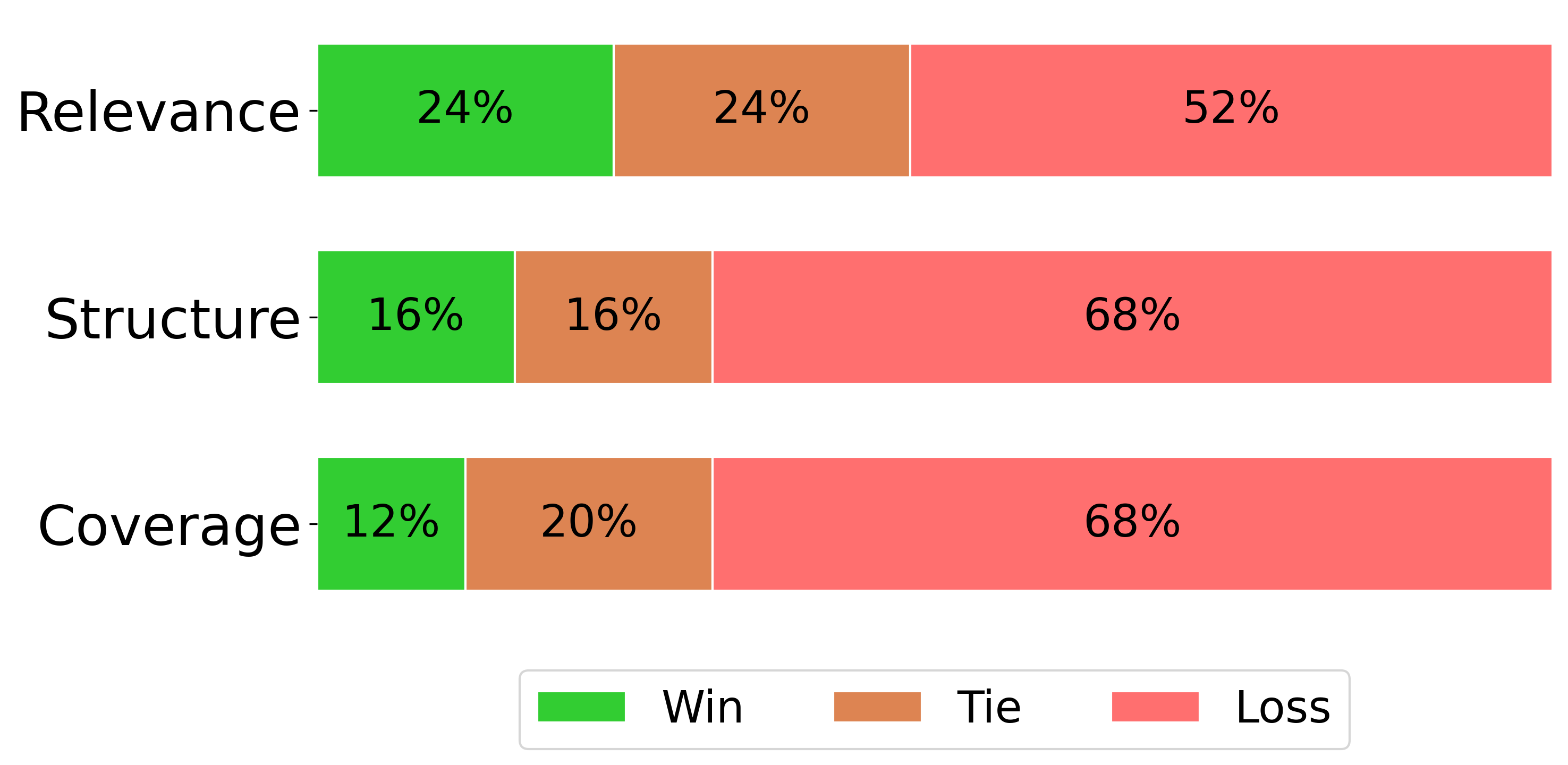}
        \subcaption{SurveyAgent-HKA  \textit{\textbf{VS}} Human-written } \label{fig:8b}
    \end{minipage}%
\caption{Human evaluation results in two groups.}
    \label{fig:8}
\vspace{-10pt}
\end{figure}

The human evaluation results are shown in Figure \ref{fig:8}, which includes two groups. The first group compares the final
version with the initial version generated by our SurveyAgent-HKA framework (\ref{fig:8a}), and the second group compares
our model with the human-written survey (\ref{fig:8b}). It can be observed that, in the internal model comparison, the final
version outperforms the initial version in all three dimensions with improvements in structure and coverage exceeding 40\%. This further highlights the importance of dynamic updates and iteration in improving the structure and quality of
the survey generated by SurveyAgent-HKA. However, when compared to the human-written ground truth, the human-written surveys still have a clear advantage, indicating that while our framework can automatically generate basic surveys, there is still a considerable gap for further enhancement to match the quality of human-written surveys. To quantify the differences, we provide some detailed human evaluation results in Table \ref{tab:10}.

% token消耗结果

\begin{table}[htbp]
\centering
\caption{Detailed human evaluation examples of Human-written versions and SurveyAgent-HKA generated surveys.}
\begin{tabular}{l l l>{\raggedright\arraybackslash}p{0.4\textwidth}} 
\toprule
\textbf{Survey topics} & \textbf{Reference} & \textbf{Win} & \textbf{Explanation} \\
\midrule
LLM-Generated Texts Detection &  \citep{wu2023survey} & Human & LLM version is somewhat limited in scope,  while the human version has broader analysis of recent developments and emerging technologies. \\
LLMs for Recommendation & \citep{zhao2024recommender} & Human &  LLM version shows clear signs of AI generation, contains redundant statements, while the human version provides a clear method comparison that helps in understanding the field. \\
Acceleration for LLMs & \citep{xu2023survey} & SurveyAgent-HKA & LLM version has a better structural design, while human-written ones are too flat, and the definitions are not clear. \\
LLMs for Information Retrieval  & \citep{zhu2025large} & Human &  Human-written version provides a complete introduction to query, retrieval, and ranking-related work; the LLM version lacks work related to ranking and future discussion. \\
LLMs for Software Engineering & \citep{hou2024large} & Human &  Human version delivers a trend analysis and solid summary of different topics, while the LLM version only lists the related work. \\
\bottomrule
\end{tabular}
\label{tab:10}
\end{table}

\begin{figure}[pos=htbp]
    \centering
    \includegraphics[width=\linewidth]{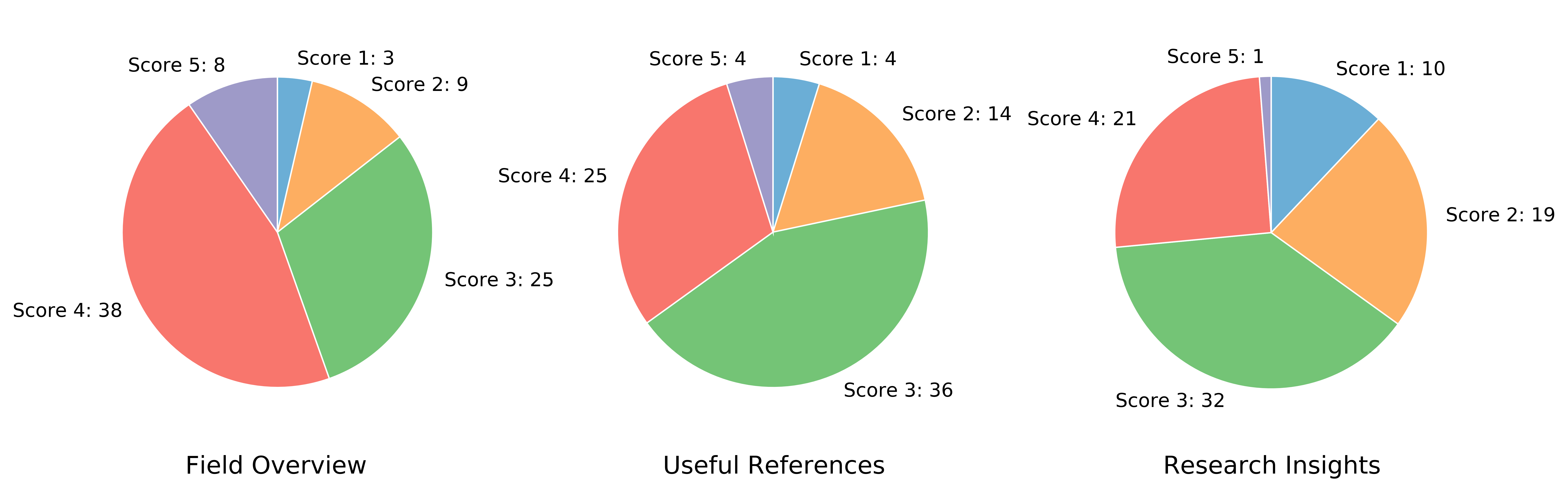}
    \caption{{\color{black}Score distributions of the user study across three evaluation aspects.}}
    \label{fig:user_study}
    \vspace{-10pt}
\end{figure}

\subsubsection{\color{black}Practical evaluation through user study}
{\color{black}
In the user study, 10 generated surveys were distributed, yielding 83 valid responses. As shown in Figure~\ref{fig:user_study}, ratings across all three dimensions are predominantly concentrated between 3 and 4, indicating that participants generally found the generated surveys useful. The relatively strong ratings for \textit{Field Overview} and \textit{Useful References} suggest that the surveys are effective in helping users quickly grasp the scope of a research field and identify relevant literature. In comparison, \textit{Research Insights} receives slightly lower ratings, indicating that the system is less consistent in synthesizing open challenges, identifying research gaps, and developing future directions. These findings suggest that the generated surveys can serve as useful starting points for literature exploration and survey drafting, but still require expert involvement to strengthen critical analysis and high-level research synthesis.
}

\begin{table*}[t]
\centering
\begingroup
\color{black}
\arrayrulecolor{black}

\caption{\textcolor{black}{Human validation of LLM-based judgments for survey classification, structural evaluation, and content evaluation. Survey classification is treated as a binary task, structural consistency is rated on a 5-point scale, and content evaluation verifies the LLM judgments used to measure key point recall (KPR).}}
\label{tab:llm_human_consistency}

\small
\setlength{\tabcolsep}{4pt}
\renewcommand{\arraystretch}{1.2}

\begin{tabularx}{\textwidth}{
p{3.0cm}
p{0.5cm}
p{0.7cm}
p{5.8cm}
p{3.0cm}
X
}
\toprule
\textbf{LLM-based evaluation} &
\textbf{N} &
\textbf{Phds} &
\textbf{Human task} &
\textbf{Metric} &
\textbf{Result} \\
\midrule

Survey classification &
20  &
1 &
Verify the LLM-assigned survey labels &
Accuracy &
Acc = 90\% \\

Structural evaluation &
20 &
2 &
Score the consistency between outline pairs &
Mean Absolute Error &
\begin{tabular}[t]{@{}l@{}}
MAE$_1$ = 0.70\\
MAE$_2$ = 0.95
\end{tabular} \\

Content evaluation &
30 &
2 &
Judge whether key points are covered &
\begin{tabular}[t]{@{}l@{}}
Cohen's Kappa\\
Accuracy
\end{tabular} &
\begin{tabular}[t]{@{}l@{}}
$\kappa = 0.82$\\
Acc = 84\%
\end{tabular} \\

\bottomrule
\end{tabularx}

\vspace{2pt}
\begin{minipage}{0.98\textwidth}
\footnotesize
\end{minipage}

\arrayrulecolor{black}
\endgroup
\end{table*}

\subsubsection{\textcolor{black}{Consistency between LLM-as-judge and human evaluations}}

{\color{black}
Since LLMs are involved in several evaluation stages, we further conduct human validation to examine the reliability of their judgments in survey classification, structural evaluation, and content evaluation. Since evaluating scientific surveys requires domain knowledge, we conduct the human validation in the computer science domain. The annotators are PhD students with computer science backgrounds and peer-reviewed publications in related venues. The validation settings and results are summarized in Table~\ref{tab:llm_human_consistency}. For survey classification, one annotator manually verifies 20 papers identified as surveys by the LLM, resulting in an accuracy of 90\%.  For structural evaluation, two annotators independently assess 20 outline pairs using the same 5-point criteria. We separately calculate the MAE between the ten-run average LLM scores and each annotator's
scores, obtaining MAE values of 0.70 and 0.95, indicating that their evaluations differ by less than one point on average. For content evaluation, two annotators independently determine whether 30 key points extracted from five human-written surveys are covered by the corresponding generated surveys. Their judgments achieve a Cohen's $\kappa$ of 0.82, while the LLM-based evaluator obtains an accuracy of 84\% against the adjudicated human labels. Overall, these results indicate that the LLM-based evaluations show reasonable agreement with human judgments and can provide scalable evaluation signals in our setting.}

\subsection{Cost and efficiency analysis}
\begin{table}[ht]
\centering
\caption{Average cost and time efficiency in SurveyAgent-HKA.}
\label{tab:cost_analysis}
\begin{tabular}{lllll}
\toprule
\textbf{Stage} & \textbf{Input Tokens} & \textbf{Output Tokens} & \textbf{Time (min)} & \textbf{Cost} \\
\midrule
Retrieval & 0 & 0 & 4 & \$0 \\
Data processing & 1.1M & 0.18M & 8 & \$0.75 (GPT-4.1-mini) \\ 
Writing   & 2.5M & 0.21M & 12 & \$0.50 (GPT-4o-mini) / \$0.89 (Claude-3.5-Haiku) \\
\bottomrule
\label{tab:12}
\vspace{-10pt}
\end{tabular}
\end{table}

The average cost and efficiency in the SurveyAgent-HKA are presented in Table \ref{tab:12}. We divide the process into three phases: literature retrieval, data processing, and survey writing. It can be seen that the retrieval phase is time-efficient, taking only 4 minutes for retrieval, citation expansion, and quality signal supplement. Additionally, it does not require LLM calls, as it relies on APIs or full-text databases for data acquisition. 
During the data processing phase, which involves tasks such as rating the relevance of literature, knowledge extraction from downloadable PDF full-text papers, retrieving relevant review comments, generating peer-review checklists, and other related tasks, the average cost is \$0.75. In the Writing stage, SurveyAgent-HKA generates a survey of length comparable to that of human-written surveys, with an average input length of 2.5M tokens and an average output of 0.21M tokens, which also includes the intermediate polishing process. This phase is completed in approximately 12 minutes (variations based on API call). Specifically, the cost is \$0.50 when using GPT-4o-mini and increases to \$0.89 when using Claude-3.5-Haiku. Overall, our method can assist researchers in generating a comprehensive survey while remaining within a reasonable budget in both time and financial cost.

\subsection{Ablation study}

\label{sec:55}
\vspace{4pt}
\subsubsection{Ablation experiments on citation quality}
\vspace{4pt}

We report the ablation experiments on citation quality in Table \ref{tab:13}, focusing on two key components: citation expansion and re-ranking. It is evident that removing either component leads to significant drops in citation performance. Specifically, removing citation expansion results in a drop in F1-score by 3.10\%  in Computer Science and 2.72\% in Medicine. Similarly, removing re-ranking causes a larger decline in F1-score, with a decrease of 4.83\%  in computer science and 4.56\% in medicine. We believe that citation quality is fundamental to the overall quality of the generated content, as well-supported citations ensure that the survey is both reliable and comprehensive. Therefore, the above results highlight that expanding citations to include a wider range of relevant work, while re-ranking them to select the most important sources, is crucial for our SurveyAgent-HKA framework.

% % 引文质量实验
\begin{table}[ht]
\centering
\renewcommand{\arraystretch}{1}
\caption{Ablation experiments results on citation quality. The best and second-best results are highlighted in \textbf{bold} and \underline{underlined}.}
\begin{tabular}{lcccccc}
\toprule

\multirow{2}{*}{\textbf{Model}}   & \multicolumn{3}{l}{\textbf{CS}} & \multicolumn{3}{l}{\textbf{Med}} \\
\cmidrule(lr){2-4} \cmidrule(lr){5-7}
\textbf{} & \textbf{P$\uparrow$} & \textbf{R$\uparrow$} & \textbf{F1$\uparrow$} & \textbf{P$\uparrow$} & \textbf{R$\uparrow$} & \textbf{F1$\uparrow$} \\
\midrule
\textbf{SurveyAgent-HKA} & \textbf{14.95} & \textbf{19.53}  & \textbf{16.94}  &  \textbf{16.78}   & \textbf{22.95}  &  \textbf{19.39}  \\
\hspace{1em}w/o Citation expansion & 12.03 & 16.29 & 13.84 & 15.02 & 18.73  &16.67   \\
\hspace{1em}w/o Re-ranking &  10.17  &  14.95  &  12.11  & 12.90 &  17.44 &  14.83  \\

\bottomrule
\label{tab:13}
\vspace{-10pt}
\end{tabular}
\end{table}

% % 结构质量实验
\begin{table}[ht]
\centering
\caption{Ablation experiments results on structural consistency. The best and second-best results are highlighted in \textbf{bold} and \underline{underlined}.}
\renewcommand{\arraystretch}{1}
\begin{tabular}{lcccc}
\toprule
\multirow{2}{*}{\textbf{Model}}  & \multicolumn{2}{l}{\textbf{CS}} & \multicolumn{2}{l}{\textbf{Med}} \\
\cmidrule(lr){2-3} \cmidrule(lr){4-5}
\textbf{} & \textbf{F1-H1$\uparrow$} & \textbf{F1-H2$\uparrow$} & \textbf{F1-H1$\uparrow$} & \textbf{F1-H2$\uparrow$} \\
\midrule
\textbf{SurveyAgent-HKA} & \textbf{38.43} & \textbf{22.17} & \textbf{30.22} & \textbf{19.26}  \\
\hspace{1em}w/o Topic cluster & 35.09 & 18.54 & 28.07  & 17.73  \\
\hspace{1em}w/o Few-shot learning & 31.41 & 15.09 & 25.86 & 14.32 \\
\bottomrule
\label{tab:14}
\end{tabular}
\vspace{-10pt}
\end{table}

\subsubsection{Ablation experiments on structural consistency}
The ablation experiments on structural consistency are shown in Table \ref{tab:14}, examining the impact of topic clustering and few-shot learning on the quality of the generated survey outlines. When topic clustering is removed, the first-level section F1-score decreases by 3.34\% in computer science and 2.15\% in medicine, highlighting the importance of clustering in organizing the primary sections of the survey, as they provide the main topic covered by the retrieved candidate literature. Meanwhile, the second-level section F1-score also drops by 3.63\% in computer science and 1.53\% in medicine. Similarly, removing few-shot learning leads to a more significant decline, with a decrease of 7.02\% in computer science and 4.36\% in medicine. This highlights the important role of human-written outlines, which provide a clear understanding of how different sections of a survey should be organized logically, helping LLMs learn the structure of scientific surveys and generate outlines that better align with human-written structures.

\begin{table}[ht]
\centering
\caption{Ablation experiments results on content quality. The best and second-best results are highlighted in \textbf{bold} and \underline{underlined}.}
\renewcommand{\arraystretch}{1}
\begin{tabular}{lcccccc}
\toprule
\multirow{2}{*}{\textbf{Model}}  & \multicolumn{3}{l}{\textbf{CS}} & \multicolumn{3}{l}{\textbf{Med}} \\
\cmidrule(lr){2-4} \cmidrule(lr){5-7}
\textbf{} & \textbf{Similarity$\uparrow$} & \textbf{ROUGE-L$\uparrow$} & \textbf{KPR$\uparrow$} & \textbf{Similarity$\uparrow$} & \textbf{ROUGE-L$\uparrow$} & \textbf{KPR$\uparrow$} \\
\midrule
\textbf{SurveyAgent-HKA}            & \textbf{83.49} &  \textbf{14.68}  &  \textbf{53.85}  &  \textbf{83.71}  & \textbf{15.96} &  \textbf{57.07} \\
\hspace{1em}w/o Knowledge extraction  &  83.23  &  12.08     &   50.31   &    83.26   &     14.15  &   54.64   \\
\hspace{1em}w/o Peer-review feedback  &  83.04   &  11.87     &  43.60     &   81.92    &   11.74   &  48.59    \\

\bottomrule
\label{tab:15}
\vspace{-10pt}
\end{tabular}
\end{table}

\begin{figure}[pos=htbp]
    \centering
    \includegraphics[width=\textwidth, trim=0.2cm 2cm 2cm 0cm, clip]{figure11.pdf} 
    \caption{Example of the generated survey by our SurveyAgent-HKA, where the section marked with {\textcolor[rgb]{0.0, 0.7, 0.0}{green}} color represents the overlap with the ground truth survey \citep{zhu2025large}, and the {\textcolor[rgb]{1.0, 0.0, 0.0}{red}} references represent the same literature selected by humans.}
\label{fig:9}
\vspace{-10pt}
\end{figure}

\vspace{4pt}
\subsubsection{Ablation experiments on content quality}
\vspace{4pt}

We further evaluate the impact of knowledge extraction and peer-review feedback on content quality, as shown in Table \ref{tab:15}. When knowledge extraction is removed, there is a slight decrease in performance, with the KPR score dropping to 50.31\% for computer science and 54.64\% for medicine. Additionally, Similarity and ROUGE scores also declined. This decrease can be attributed to the fact that knowledge extraction, which pulls content directly from full-text papers using predefined templates, offers a richer and more detailed understanding of the literature compared to abstracts. However, due to copyright restrictions, we are only able to access a limited number of full-text PDFs, which limits the scope of knowledge extraction. Removing peer-review feedback also leads to a more significant decline, with KPR scores decreasing to 43.60\% in computer science and 48.59\% in medicine. This drop underscores the crucial role of peer reviews, particularly from field experts, who offer valuable insights into the weaknesses of scientific surveys, such as missing topics, methodological flaws, or unaddressed areas that could strengthen the quality of the draft survey. To our knowledge, SurveyAgent-HKA is the first model to optimize survey generation by using existing feedback from human experts to simulate the real-world peer review process, which sets it apart as an innovative approach to improving content quality in this task.  {\color{black}To further examine whether the observed improvements are solely due to additional retrieval resources, we also conduct a resource-controlled experiment, with detailed results reported in Appendix~\ref{sec:controlled}}

\subsection{Case study}
\label{sec:56}
In this case study, we explore the effectiveness of SurveyAgent-HKA in generating the survey on the topic of ``\textit{Large Language Models for Information Retrieval}''. As shown in Figure \ref{fig:9}, the generated survey showcases a clear and logical structure, with each section well designed to ensure a balanced flow of sections and seamless transitions from one topic to the next. Unlike the previous work AutoSurvey \citep{wang2024AutoSurvey}, which tends to have an equal number of sub-chapters for each section, our system offers a more focused and detailed exploration of key topics rather than a rigid distribution of subsections. In particular, our survey begins with a historical overview of \textit{``Information Retrieval''} and the second section provides a detailed examination of traditional IR models and the rise of LLMs, focusing on their distinct advantages over classical models. In the third section, the survey focuses on the application of LLMs in IR and is subdivided into five main sub-chapters; three of them overlap with sections in the human-written survey, highlighting that SurveyAgent-HKA effectively identifies key research areas that correspond with those selected by human experts. Furthermore, out of the 11 references in the \textit{``Introduction''} section, 2 match the references chosen by human experts, and we observed that each statement is clearly related to the cited literature. However, compared to the sections summarizing past research, many subsections in the generated survey are less detailed than those in human-written surveys, especially in areas requiring summary and in-depth analysis, such as challenges and future directions. These sections involve more than basic summarization and require thorough analysis and a deep understanding of the field, capabilities that, at this point, may be beyond the capabilities of LLMs \citep{zhu2024llms}.

\begin{table*}[htbp]
{\color{black}
\centering
\caption{\textcolor{black}{Residual error analysis of generated surveys.}}
\label{tab:error_distribution}

\small
\renewcommand{\arraystretch}{1.18}
\setlength{\tabcolsep}{3.5pt}

\begin{tabularx}{\textwidth}{
    >{\raggedright\arraybackslash}p{1.30cm}
    >{\raggedright\arraybackslash}p{1.90cm}
    >{\centering\arraybackslash}p{1.00cm}
    >{\raggedright\arraybackslash}p{4.25cm}
    >{\raggedright\arraybackslash}X
}
\toprule
\textbf{Dimension} &
\textbf{Error type} &
\textbf{Counts} &
\textbf{Common issues} &
\textbf{Cause analysis} \\
\midrule

Citation
& Missing key references
& 12
& Foundational, representative, or recently published studies are missed.
& Human--algorithm ranking mismatch;\newline
  Incomplete database coverage;\newline
  Query--paper semantic mismatch. \\

\cmidrule(lr){2-5}

& Citation--claim mismatch
& 8
& Citations do not support the associated claims or are placed with inappropriate statements.
& Restricted access to full-text PDFs;\newline
  Loss of supporting evidence during PDF parsing;\newline
  Incorrect source attribution during writing;\newline
  LLM hallucinations during generation. \\

\midrule

Structure
& Missing key topics/sections
& 13
& Important research directions or specialized subtopics are missing from the generated outline.
& Fixed clusters obscure minor topics;\newline
  Related but distinct topics are merged;\newline
  Inaccurate topic expansion during outline generation. \\

\cmidrule(lr){2-5}

& Homogenized section structure
& 6
& Different sections follow similar organizational patterns and lack topic-specific structures.
& Uniform outline prompts across topics;\newline
  Homogenization in LLM outputs. \\

\midrule

Content
& Insufficient research insights/discussion
& 16
& The surveys provide limited discussion of open challenges and future directions.
&  Paper-level extraction limits cross-paper synthesis;\newline
  Compressed evidence weakens critical analysis;\newline
  Generic feedback lacks topic specificity. \\

\cmidrule(lr){2-5}

& Limited cross-paper comparison
& 14
& Studies are summarized individually, with limited comparison of their methods, results, strengths, and limitations.
& Limited extraction of cross-paper relations;\newline
  Visual information is excluded;\newline
  Insufficient comparison-oriented guidance;\newline
  Separate generation of individual study summaries.\\

\cmidrule(lr){2-5}

& Non-academic language
& 5
& The writing is sometimes overly repetitive or insufficiently precise for academic communication.
& Limited academic-style guidance;\newline
  Repetition introduced during iterative rewriting;\newline
  Insufficient language-focused refinement. \\

\bottomrule
\end{tabularx}

\vspace{2pt}
\begin{minipage}{0.98\textwidth}
\footnotesize
\end{minipage}
}
\end{table*}

\subsection{\color{black}{Error analysis}}
{{\color{black}
To better understand the remaining limitations of SurveyAgent-HKA, we conduct a manual error analysis of the generated surveys, as shown in Table~\ref{tab:error_distribution}. The fine-grained error taxonomy was derived from recurring issues identified in the written explanations provided during the pairwise human evaluation and limitations described by participants in the user study and was further refined through a pilot inspection.  The most frequent problems concern research insights and cross-paper comparison, suggesting that the system is more reliable at summarizing individual studies than at producing deeper synthesis. These limitations are also related to earlier stages of the pipeline. Relevant studies may be missed because they are not indexed in the selected databases or are not adequately identified by the search queries and ranking process. Even when a paper is successfully retrieved, copyright restrictions may prevent access to its full-text PDF. As a result, the framework can make fuller use of open-access papers, whereas information from inaccessible papers is often limited to titles and abstracts. Such information loss also affects the writing stage, making it difficult to provide detailed comparisons across studies or develop sufficiently critical discussions. Therefore, fully automated survey writing remains difficult in practice; human input is needed earlier in the workflow, particularly for literature selection, evidence verification, and outline refinement, since errors introduced at these stages are difficult to correct during writing.}

\vspace{4pt}
\section{Discussion}
\label{sec:disscussion}
\subsection{Implications}
Our work has several practical implications. First, automatic survey generation can help reduce the burden of information overload. For early-stage researchers, in particular, a high-quality survey is an essential resource for understanding research development and trends in their field, and it can provide insights for future research directions. Our method integrates the entire process of literature retrieval, paper ranking, knowledge extraction, survey generation, and evaluation, improving the efficiency of writing such long-form survey documents. 

The value of our framework is twofold. First, it provides an efficient solution for survey generation in domains where large amounts of literature need to be synthesized. By using multiple LLM-powered agents, researchers can generate draft surveys that summarize key studies, identify major trends, and highlight research gaps. This can help researchers stay up to date with fast-growing scientific literature. {\color{black}Second, our work is also complementary to other methods that aim to automate or assist scientific review writing. For example, network-analysis and NLP-based methods have been used to provide a landscape view of a research field by identifying major topics, topic evolution, and influential research areas from large-scale literature collections \citep{brito2023network}. Such methods do not directly generate full survey articles, but they can provide valuable upstream support for survey generation. The topic structures, citation networks, and impact signals produced by these methods could be integrated into our retrieval and ranking stages to improve paper selection and outline construction. Recent AI-assisted review writing systems further use RAG and modular LLM agents to generate citation-rich review drafts from manually curated papers or citation-network-derived corpora \citep{silva2025ai}. This study supports our motivation by showing that LLM-based scientific writing tools can assist researchers in producing initial review drafts. However, it also highlights the need for more reliable and human-controllable review generation. SurveyAgent-HKA extends this line of work by moving beyond draft generation from a predefined literature set and introducing an end-to-end framework that integrates cross-platform retrieval, literature re-ranking, human-written outline guidance, and peer-review-based revision. }

In real-world implementation, our model is based on publicly accessible databases that provide full-data downloads and support online API retrieval. All of the tools used are also publicly available. Moreover, Table~\ref{tab:12} demonstrates the effectiveness of our model in terms of time and financial cost. Considering that our writing process relies on LLM APIs, there are almost no data annotation or training costs. Finally, the dataset and evaluation protocols introduced in this study establish a foundation for future research on automated survey generation, enabling systematic comparisons across models and domains, and facilitating progress toward more reliable AI-related scientific writing practices.

\subsection{Limitations} We acknowledge several limitations in this study. 

\textbf{Limited visual content and evaluation scale.} The generated surveys focus on textual content and do not include figures, tables, or other visual information, which are common elements in published surveys. However, we maintained consistency with the experimental setups of previous works to ensure a fair comparison. In addition, due to the high API cost of long-form generation and evaluation, our experiments are conducted on a relatively small dataset. {\color{black} Although the framework demonstrates consistent effectiveness in both computer science and medicine, its applicability to other disciplines, publication venues, survey lengths, and emerging research areas remains to be further examined. Since our evaluation relies on human-written surveys as gold standards, a rigorous length- or venue-aware evaluation would require additional matched human surveys and comparable annotations, which are beyond the scope of this study.} 

\textbf{Potential data contamination.} {\color{black}We cross-referenced the earliest public release dates of the evaluation surveys with the knowledge cutoff points reported from the backbone LLMs, as detailed in the Appendix~\ref{appendixA}. According to the official documentation, GPT-4o has a knowledge cutoff of October 2023, while Claude-3.5-Sonnet has a cutoff of April 2024. Therefore, some human-written surveys may have appeared in the models' pretraining data. Potential prior exposure could influence generation by enabling the backbone LLM to recall the references, structure, or conclusions of a target survey. It may also bias the LLM-based evaluation of structural consistency and key point coverage through familiarity with the reference survey. To reduce the risk of direct information leakage during our experiments, we excluded each target survey and its alternative versions, including preprints and extended versions, from the literature retrieval results and from the human-written surveys used for outline enhancement. Moreover, all compared methods were developed after 2024 and evaluated on the same survey set using the same backbone LLMs. Therefore, any contamination-related advantage would be largely shared across the compared methods, potentially affecting the absolute scores but being less likely to fully explain their relative performance differences.}

\textbf{Gap from expert-written surveys.} 
{\color{black}Finally, despite promising results, generated surveys still lag behind expert-written ones, particularly in identifying key references, maintaining balanced and coherent outlines, and providing in-depth cross-paper comparisons and research insights. The generated surveys should therefore be regarded as supportive materials for literature exploration rather than substitutes for expert-written surveys. Human verification and further refinement remain necessary.} To support future research in this area, we will release the dataset and code used in this study, enabling other researchers to build upon our work and push the boundaries of automated scientific survey generation.

\section{Conclusion and future work}
\label{sec:conclusion}

We proposed SurveyAgent-HKA, a multi-agent framework for automated survey generation that combines LLMs and human knowledge augmentation. In SurveyAgent-HKA, we first define a unified data collection pipeline to retrieve relevant literature from multiple data sources related to the survey topic. Then, we generate a topic-related survey structure based on the retrieved literature and use human-crafted outlines to guide the LLM for outline optimization. Building upon the refined outline, we further gather literature relevant to subtopics and employ citation network analysis along with a re-ranking module to identify key, influential papers for each section. Furthermore, to better utilize the retrieved references, we extract the structured knowledge typically required for writing survey papers and use it to generate well-supported sections that contribute to the raw survey draft. Afterward, we gather peer review comments from relevant published surveys, identify common issues highlighted by experts, and use this human knowledge to generate revision suggestions and iteratively refine the draft into the final survey. Experiments in both the computer science and biomedical domains show that our approach outperforms mainstream baselines in terms of citation quality, structural consistency, and content quality, and ablation studies further confirm the contribution of each module designed in our framework. In an era of rapidly expanding publications, our work provides a timely solution to assist researchers in quickly generating surveys on a given topic, while also offering valuable insights for broader academic collaboration scenarios, such as related work generation and other knowledge synthesis tasks.

In future work, we plan to incorporate visual information, such as images and tables, into the survey generation process to help researchers gain deeper insights into the field. Additionally, conducting experiments on larger datasets and across other disciplines to further validate the robustness of our approach is also being considered. Furthermore, we intend to develop a web-based system that will serve as an easy-to-use platform for users to generate customized surveys that meet their specific needs.

\section*{Declaration of competing interest}
\noindent The authors declare that they have no competing interests.
\vspace{-10pt}
\section*{Data availability}
\noindent The code and dataset will be released at \href{https://github.com/tongbao96/code-for-SurveyAgent-HKA}{https://github.com/tongbao96/code-for-SurveyAgent-HKA} after review.
\vspace{-10pt}

\section*{Acknowledgment}
\noindent This paper was supported by the  Jiangsu Social Science Fund (No.26XZA001). 

\clearpage

\newpage
\thispagestyle{empty} 

\appendix
\section{Appendix}
\subsection{Survey examples}
\label{appendixA}

To ensure a fair comparison, we selected 30 surveys that are the same as those in previous studies for comparison. Among them, the surveys in the computer science field are from AutoSurvey \citep{wang2024AutoSurvey}, and those in the medical field are from SurveyGen \citep{bao-etal-2025-surveygen}. The specific surveys are shown in Table \ref{tab:16}.

\begin{table}[ht]
\centering
\caption{Detail of the selected survey papers.}
\resizebox{\textwidth}{!}{ % Scale the table to fit the text width
\begin{tabular}{llrr}
\toprule
\textbf{Topic} & \textbf{Survey Title} & \textbf{Citations} & \textbf{\color{black}Publication Year} \\
\midrule
\multicolumn{4}{c}{\textbf{Computer Science}} \\
\midrule
In-context Learning & A survey for in-context learning & 323  & \color{black}2022.12 \\
LLMs for Recommendation & A Survey on Large Language Models for Recommendation & 55  & \color{black}2023.05 \\
LLM-Generated Texts Detection & A Survey of Detecting LLM-Generated Texts & 42  & \color{black} 2023.10 \\
Explainability for LLMs & Explainability for Large Language Models & 25  & \color{black}2023.09\\
Evaluation of LLMs & A Survey on Evaluation of Large Language Models & 183  & \color{black}2023.07\\
LLMs-based Agents & A Survey on Large Language Model based Autonomous Agents & 101  & \color{black}2023.08\\
LLMs in Medicine & A Survey of Large Language Models in Medicine & 234  & \color{black}2023.11\\
Domain Specialization of LLMs & Domain Specialization as the Key to Make Large Language Models Disruptive & 14  & \color{black}2023.05\\
Challenges of LLMs in Education & Practical and Ethical Challenges of Large Language Models in Education & 53  & \color{black}2023.03\\
Alignment of LLMs & Aligning Large Language Models with Human & 53  & \color{black}2023.07 \\
ChatGPT & A Survey on ChatGPT and Beyond & 144  & \color{black}2023.04\\
Instruction Tuning for LLMs & Instruction Tuning for Large Language Models & 45  & \color{black}2023.08\\
LLMs for Information Retrieval & Large Language Models for Information Retrieval & 22  & \color{black}2023.08\\
Safety in LLMs & Towards Safer Generative Language Models: Safety Risks, Evaluations, and Improvements & 17  & \color{black}2023.02 \\
Chain of Thought & A Survey of Chain of Thought Reasoning & 13  & \color{black}2023.09\\
Hallucination in LLMs & A Survey on Hallucination in Large Language Models & 116  & \color{black}2023.11\\
Bias and Fairness in LLMs & Bias and Fairness in Large Language Models & 12 & \color{black}2023.09 \\
Large Multi-Modal Language Models & Large-scale Multi-Modal Pre-trained Models & 61  & \color{black}2023.02\\
Acceleration for LLMs & A Survey on Model Compression and Acceleration for Pretrained Language Models & 22  & \color{black}2022.11 \\
LLMs for Software Engineering & Large Language Models for Software Engineering & 49  & \color{black}2023.08\\
\midrule
\multicolumn{4}{c}{\textbf{Medicine}} \\
\midrule
Impact of Sugars on Body Weight & Dietary Sugars and Body Weight: Systematic Review and Meta-analyses of Randomised Controlled Trials and Cohort Studies & 1583  & \color{black}2012.01 \\
Benefits of Medicinal Plant Compounds & Flavonoids and Other Phenolic Compounds from Medicinal Plants for Pharmaceutical and Medical Aspects: An Overview & 1408  & \color{black}2018.08\\
Patient Involvement in Medical Research & Patient Engagement in Research: A Systematic Review & 1135 & \color{black}2014.02 \\
Delirium on Clinical Outcome & Impact of Delirium on Clinical Outcome in Critically Ill Patients: A Meta-analysis & 793 & \color{black}2013.03 \\
Chinese Medicine for SARS-CoV-2  & Traditional Chinese Medicine in the Treatment of Patients Infected with 2019-new Coronavirus (SARS-CoV-2): A Review and Perspective & 769  & \color{black}2020.03\\
Cognitive Reserve and Dementia Risk & Education and Dementia in the Context of the Cognitive Reserve Hypothesis: A Systematic Review with Meta-analyses and Qualitative Analyses & 740  & \color{black}2012.06\\
Effectiveness of ADHD Medications & Comparative Efficacy and Tolerability of Medications for Attention-deficit Hyperactivity Disorder in Children, Adolescents, and Adults: A Systematic Review & 705  & \color{black}2018.08 \\
Air Pollutions in Stroke Development & Short Term Exposure to Air Pollution and Stroke: Systematic Review and Meta-analysis & 705  & \color{black}2015.03 \\
Clinical Perspectives on ALS & Amyotrophic Lateral Sclerosis: A Clinical Review & 657  & \color{black}2020.06 \\
Maternal Smoking on Birth Defects & Maternal Smoking in Pregnancy and Birth Defects: A Systematic Review Based on 173,687 Malformed Cases and 11.7 Million Controls& 626 & \color{black}2011.07 \\

\bottomrule
\label{tab:16}
\end{tabular}
}
\end{table}

\begin{figure}[htbp]
    \centering
    \begin{minipage}{0.45\textwidth}
        \centering
        \includegraphics[width=\linewidth]{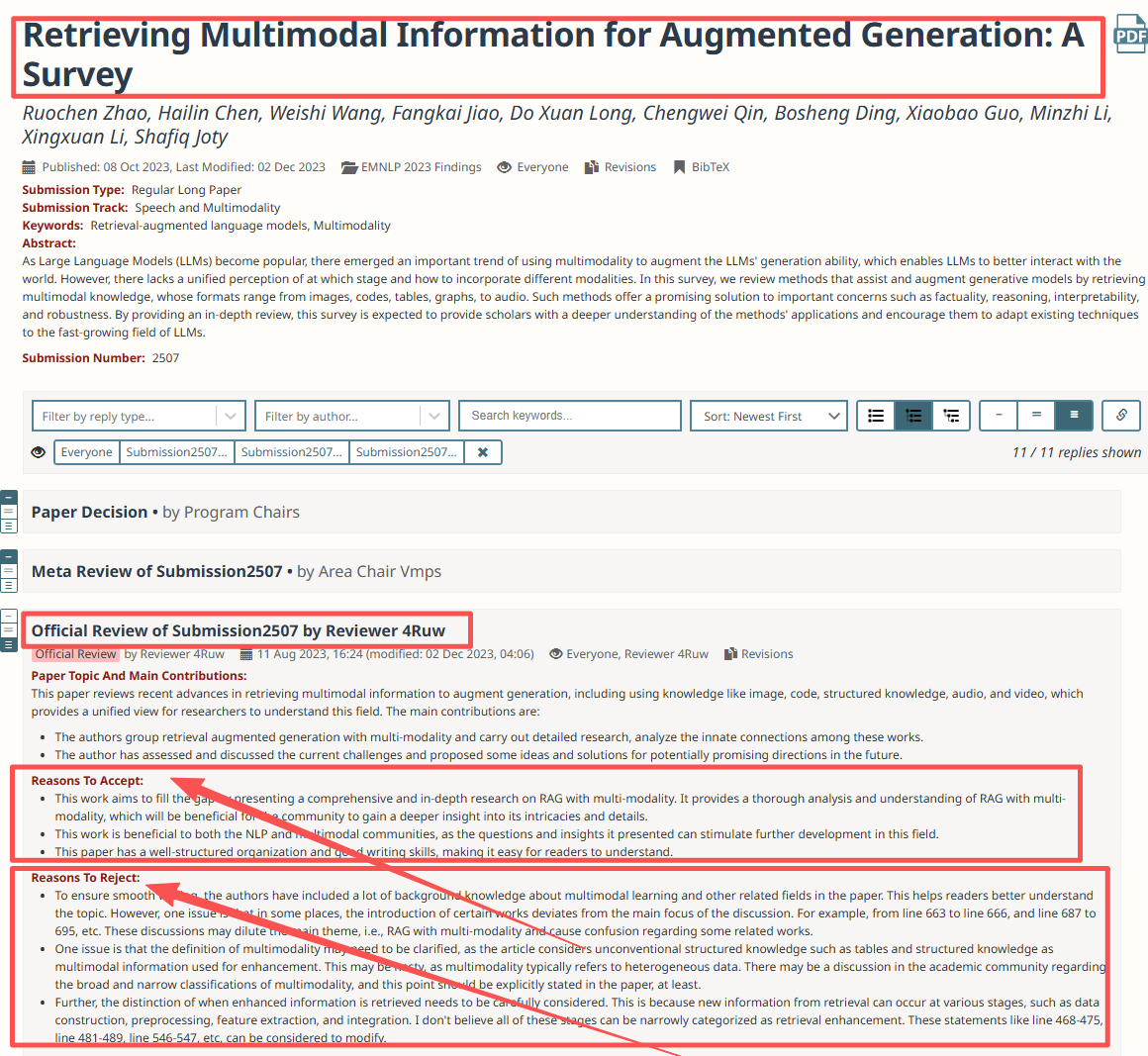} 
        \subcaption{Peer reviewer comments from the OpenReview platform.} \label{fig:10a}
    \end{minipage}%
    \begin{minipage}{0.45\textwidth}
        \centering
        \includegraphics[width=\linewidth]{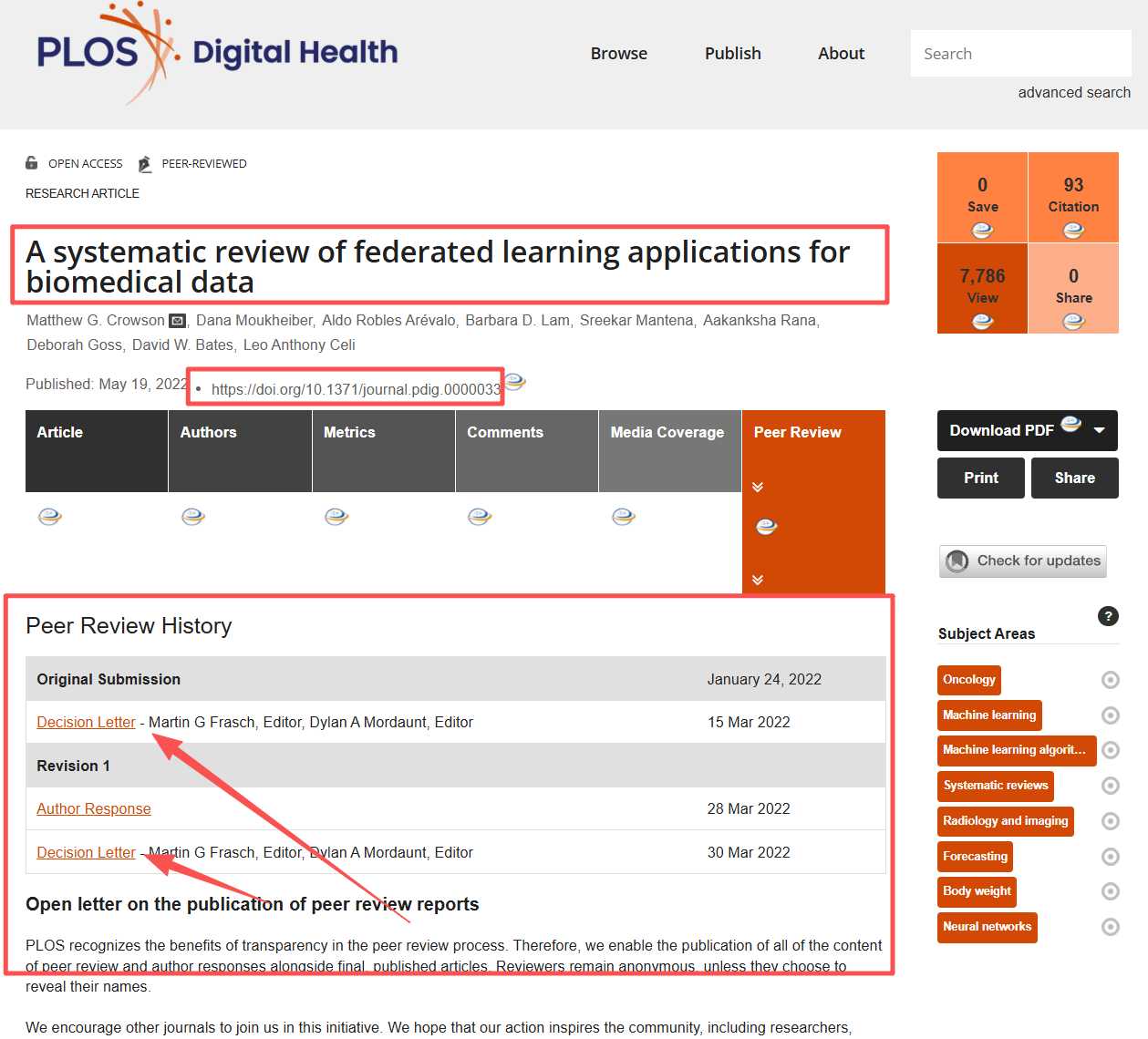} 
        \subcaption{Peer reviewer comments from PLOS One.} \label{fig:10b}
    \end{minipage}%
    \caption{Examples of peer reviewer comments on survey paper from OpenReview and PLOS One, where the former is the peer review platform for most major leading computer science conferences, and the latter is a well-known open-access medical journal.} 
    \label{fig:10}
\end{figure}

\subsection{Peer review examples}
\label{app:peerreviews}
We provide two peer review examples from our dataset in Figure \ref{fig:10}.

\subsection{Human evaluation criteria}
\label{app:criteria}

We provide the human evaluation criteria in Table \ref{tab:criteria}.

\begin{table}[ht]
\centering
\caption{Criteria for human evaluation.}
\begin{tabular}{lp{12.5cm}} 
\toprule
\textbf{Dimension} & \textbf{Guidelines} \\
\midrule
\textbf{Coverage} & \textbf{Win}: The survey comprehensively covers all key aspects of the topic, addressing all major points better than the other survey. \\
 & \textbf{Tie}: Both surveys cover similar key aspects, with only minor differences in coverage. \\
 & \textbf{Loss}: The survey misses important key aspects or addresses fewer points compared to the other survey. \\
\midrule
\textbf{Structure} & \textbf{Win}: The structure is well-organized with clear and logical sections, making it more readable and easier to follow than the other survey. \\
 & \textbf{Tie}: Both surveys have similar structural clarity, with only minor differences in organization. \\
 & \textbf{Loss}: The structure is incomplete, with unclear organization or sections, making it harder to follow than the other survey. \\
\midrule
\textbf{Relevance} & \textbf{Win}: The content is highly relevant and aligns well with the survey topic, offering a more focused and precise exploration than the other survey. \\
 & \textbf{Tie}: Both surveys are equally relevant to the topic, with only minor gaps in content alignment. \\
 & \textbf{Loss}: The content is less relevant to the topic or covers less relevant aspects compared to the other survey. \\
\bottomrule
\end{tabular}
\label{tab:criteria}
\end{table}

\begin{table}[h!]
\centering
\caption{Knowledge extraction trees for paper in Computer Science and Medicine domains.}
\resizebox{\textwidth}{!}{ % Scale the table to fit the text width
\begin{tabular}{ll}
\hline
\textbf{Attribute} & \textbf{Details} \\
\hline
\multicolumn{2}{c}{\textbf{Computer Science}} \\
\hline
\textbf{1. Background} & \textbf{a. Definition:} Specific description of the problem. \\
& \textbf{b. Key Obstacle:} Main difficulty, main challenge. \\
\hline
\textbf{2. Idea} & \textbf{a. Intuition:} Idea was inspired by what. \\
& \textbf{b. Opinion:} What's the idea? \\
& \textbf{c. Innovation:} What's the main difference compared to previous method, or where is the primary improvement. \\
\hline
\textbf{3. Method} & \textbf{a. Method Definition:} Given the problem, what's the definition of the method. \\
& \textbf{b. Method Description:} In one sentence, describe the method. \\
& \textbf{c. Method Steps:} Procedures of method. \\
& \textbf{d. Principle:} Why this method is effective. \\
\hline
\textbf{4. Experiments} & \textbf{a. Experiment Setting:} Including dataset, baseline, evaluation metrics. \\
& \textbf{b. Experiment Results:} Summarize the main results and findings in one paragraph. \\
\hline
\textbf{5. Discussion} & \textbf{a. Advantage:} What's the advantages of this paper. \\
& \textbf{b. Limitation:} What's the disadvantages of this paper. \\
& \textbf{c. Future Work:} Based on the advantages and disadvantages, what and where can be improved in the future. \\
\hline
\multicolumn{2}{c}{\textbf{Medicine}} \\
\hline
\textbf{1. Introduction} & \textbf{a. Disease/Condition Overview:} Briefly describe the disease or condition under study. \\
& \textbf{b. Key Challenge/Research Gap:} What is the key issue or research gap being addressed? \\
\hline
\textbf{2. Methods} & \textbf{a. Study Design:} What is the study design (e.g., clinical trial, cohort study)? \\
& \textbf{b. Sample/Participants:} What are the inclusion/exclusion criteria for the study sample? \\
& \textbf{c. Intervention/Procedure:} What treatments, interventions, or procedures were used in the study? \\
\hline
\textbf{3. Results} & \textbf{a. Main Findings:} Summarize the key results and findings of the study. \\
& \textbf{b. Statistical Significance:} What were the statistical results? \\
\hline
\textbf{4. Discussion} & \textbf{a. Strengths and Limitations:} What are the strengths and limitations of the study? \\
& \textbf{b. Future Directions:} What are the suggested future research directions or areas for improvement? \\
\hline
\end{tabular}
\label{tab:knowledgetree}
}
\end{table}

\subsection{Knowledge extraction trees for the retrieved papers}
\label{appendixB}
Table~\ref{tab:knowledgetree} presents knowledge extraction trees for the retrieved papers. Unlike abstracts, which provide a brief overview, these knowledge extraction trees offer a more comprehensive representation of the paper content, particularly the details such as datasets, results, and other important information that are usually not included in abstracts. Moreover, they condense the full text into a more structured and concise format,  which helps reduce token cost while retaining the most relevant content for survey generation.

\subsection{\textcolor{black}{Statistical significance against baselines}}
\label{st}
{\color{black}
We conduct paired significance tests at the topic level. For each topic, we compute the performance difference between SurveyAgent-HKA and the baseline, and then apply the Wilcoxon signed-rank test to these paired differences. 
As shown in Table~\ref{tab:significance_test}, SurveyAgent-HKA achieves statistically significant improvements over most baselines across both domains. In the CS domain, the improvements over Naive-RAG, AutoSurvey, and SurveyForge are significant across all four key metrics. Compared with the stronger SurveyX baseline, SurveyAgent-HKA still obtains significant gains in Citation F1 and KPR, while the improvements in H1(F1) and H2(F1) are positive but not statistically significant. In the Medicine domain, SurveyAgent-HKA significantly outperforms Naive-RAG across all metrics, and shows significant improvements over SurveyGen in structural consistency and KPR. These results indicate that the observed gains are not merely caused by random variation, although some comparisons with stronger baselines remain non-significant under the limited evaluation scale.
}

\begin{table*}[t]
\centering
\begingroup
\color{black}
\arrayrulecolor{black}

\caption{\textcolor{black}{Statistical significance test of SurveyAgent-HKA against baselines across domains and metrics. 
Each cell reports $\Delta$ and the corresponding significance level, where $\Delta$ denotes the average topic-level improvement of SurveyAgent-HKA over the compared method.}}
\label{tab:significance_test}
\scriptsize
\setlength{\tabcolsep}{4pt}
\renewcommand{\arraystretch}{1.12}

\begin{threeparttable}
\begin{adjustbox}{max width=\textwidth}
\begin{tabular}{llcccc}
\toprule
\textbf{Domain} 
& \textbf{Compared Method} 
& \textbf{Citation F1} 
& \textbf{H1(F1)} 
& \textbf{H2(F1)} 
& \textbf{KPR} \\
\midrule

\multirow{4}{*}{CS}
& Naive-RAG  \citep{lai2024instruct}
& +10.82 ($p<0.01$)$^{**}$ 
& +9.08 ($p<0.01$)$^{**}$ 
& +10.10 ($p<0.01$)$^{**}$ 
& +15.81 ($p<0.01$)$^{**}$ \\

& AutoSurvey \citep{wang2024AutoSurvey}
& +8.59 ($p<0.01$)$^{**}$ 
& +5.36 ($p<0.01$)$^{**}$ 
& +8.04 ($p<0.01$)$^{**}$ 
& +10.47 ($p<0.01$)$^{**}$ \\

& SurveyForge \citep{yan2025surveyforgeoutlineheuristicsmemorydriven}
& +2.97 ($p<0.01$)$^{**}$ 
& +5.13 ($p<0.01$)$^{**}$ 
& +11.60 ($p<0.01$)$^{**}$ 
& +5.66 ($p<0.01$)$^{**}$ \\

& SurveyX \citep{liang2025surveyx}
& +5.30 ($p<0.01$)$^{**}$ 
& +1.59 (\textit{n.s.}) 
& +1.21 (\textit{n.s.}) 
& +3.48 ($p<0.05$)$^{*}$ \\

\midrule

\multirow{2}{*}{Med}
& Naive-RAG  \citep{lai2024instruct}
& +9.31 ($p<0.01$)$^{**}$ 
& +8.10 ($p<0.01$)$^{**}$ 
& +8.22 ($p<0.01$)$^{**}$ 
& +10.55 ($p<0.01$)$^{**}$ \\

& SurveyGen \citep{bao-etal-2025-surveygen}
& +2.11 (\textit{n.s.}) 
& +5.15 ($p<0.01$)$^{**}$ 
& +5.58 ($p<0.01$)$^{**}$ 
& +3.17 ($p<0.05$)$^{*}$ \\

\bottomrule
\end{tabular}
\end{adjustbox}
\end{threeparttable}

\endgroup
\end{table*}

\subsection{\color{black}Resource-controlled comparison}
\label{sec:controlled}
{\color{black}
We further conduct a resource-controlled comparison to examine whether the performance gains of SurveyAgent-HKA can be attributed only to richer
retrieval resources. In this setting, candidate papers were retrieved
exclusively from arXiv for Computer Science and S2ORC for Medicine, and only
their metadata, including titles and abstracts, were used for subsequent
generation. All other configurations remained the same as in the original
experiments.
As shown in Table~\ref{tab:resource_controlled_results}, the single-source
and abstract-only setting leads to performance drops compared with the full
system, confirming the contribution of multi-source retrieval and full-text
evidence. Under the same resource setting, the controlled version still
outperforms the baselines on most metrics. In Computer Science, the controlled version achieves citation quality
comparable to SurveyForge, with a slightly lower F1 score of 13.46\% versus
13.97\%. However, it achieves better structural consistency and content quality, suggesting that the proposed human-knowledge-guided planning and revision components still contribute beyond resource advantages.
}

\begin{table*}[t]
\centering
\begingroup
\color{black}
\arrayrulecolor{black}

\renewcommand{\arraystretch}{1}
\setlength{\tabcolsep}{4pt}

\caption{\textcolor{black}{
Performance comparison under the resource-controlled and full settings.
The best and second-best results are highlighted in \textbf{bold} and
\underline{underlined}, respectively. In the arXiv-only setting, the venue reputation weight is reassigned to citation performance during the re-ranking because all candidates share the same venue index. }}
\label{tab:resource_controlled_results}

\begin{tabular}{llccccccccc}
\toprule
\multirow{2}{*}{\textbf{Domain}} &
\multirow{2}{*}{\textbf{Model}} &
\multicolumn{3}{c}{\textbf{Citation quality}} &
\multicolumn{3}{c}{\textbf{Structural consistency}} &
\multicolumn{3}{c}{\textbf{Content quality}} \\
\cmidrule(lr){3-5}
\cmidrule(lr){6-8}
\cmidrule(lr){9-11}

& &
\textbf{P$\uparrow$} &
\textbf{R$\uparrow$} &
\textbf{F1$\uparrow$} &
\textbf{H1 (F1)$\uparrow$} &
\textbf{H2 (F1)$\uparrow$} &
\textbf{Rel.$\uparrow$} &
\textbf{Similarity$\uparrow$} &
\textbf{ROUGE-L$\uparrow$} &
\textbf{KPR$\uparrow$} \\
\midrule

\multirow{5}{*}{CS}
& Naïve-RAG \citep{lai2024instruct}
& 4.36 & 10.24 & 6.12
& 29.35 & 12.07
& 2.5 & 80.06 & 8.54 & 38.04 \\

& AutoSurvey \citep{wang2024AutoSurvey}
& 6.58 & 11.39 & 8.34
& 33.07 & 14.13
& 2.7 & 82.28 & 10.21 & 43.38 \\

& SurveyForge \citep{yan2025surveyforgeoutlineheuristicsmemorydriven}
& \underline{11.57} & \underline{17.62} & \underline{13.97}
& 33.30 & 10.57
& 3.0 & 82.14 & 12.49 & 48.19 \\

% Separate our methods from the baselines

& \textbf{Ours (Controlled)}
& 11.09
& 17.13
& 13.61
& \underline{36.76}
& \underline{17.95}
& \underline{3.2}
& \underline{82.45}
& \underline{12.82}
& \underline{50.67} \\

& \textbf{Ours (Full)}
& \textbf{14.95}
& \textbf{19.53}
& \textbf{16.94}
& \textbf{38.43}
& \textbf{22.17}
& \textbf{3.5}
& \textbf{83.49}
& \textbf{14.68}
& \textbf{53.85} \\

\midrule

\multirow{4}{*}{Med}
& Naïve-RAG \citep{lai2024instruct}
& 8.27 & 12.90 & 10.08
& 22.12 & 11.01
& 3.2 & 82.93 & 14.37 & 46.52 \\

& SurveyGen  \citep{bao-etal-2025-surveygen} 
& 15.31 & 19.82 & 17.28
& 25.07 & 13.68
& 3.4 & \underline{83.42} & 15.25 & 53.90 \\

% Separate our methods from the baselines
& \textbf{Ours (Controlled)}
& \underline{15.56}
& \underline{20.18}
& \underline{17.57}
& \underline{28.94}
& \underline{18.57}
& \underline{3.8}
& 83.37
& \underline{15.53}
& \underline{55.13} \\

& \textbf{Ours (Full)}
& \textbf{16.78}
& \textbf{22.95}
& \textbf{19.39}
& \textbf{30.22}
& \textbf{19.26}
& \textbf{3.9}
& \textbf{83.71}
& \textbf{15.96}
& \textbf{57.07} \\

\bottomrule
\end{tabular}

\arrayrulecolor{black}
\vspace{-10pt}
\endgroup
\end{table*}

\subsection{Prompts used in this study}
\label{appendixc}
Prompt design is key to ensuring that LLMs perform tasks reliably. In practice, structured prompts have proven to be effective in improving model performance, as they guide the LLM more clearly towards desired outcomes, reducing ambiguity and increasing task accuracy \citep{zhang-etal-2025-prompt-design, atreja2024prompt}. Therefore, we also employ a structured prompting strategy to reduce variability in model behavior and to ensure consistent adherence to task objectives. The prompt is constructed using four elements: \textbf{(1) Instruction}: Specifies which task and stage in survey generation the LLM is expected to perform, and clarifies the objective of the task. In the instruction, we can add human-informed empirical observations to better guide the LLM in completing the task. For example, survey papers in computer science typically cover methodological advances, datasets, and evaluation metrics, which could be part of the instruction to provide high-level structural guidance during generation. \textbf{(2) Context Information}: Provides the context information required for completing the task, such as conceptual definitions or evaluation criteria, to help the LLM better respond within a well-defined semantic framework. \textbf{(3) Input Data}: Provides the specific data that the LLM needs to analyze under the guidance of the given instruction. This may also include gold-standard examples that help the model better infer the task. \textbf{(4) Output Format}: Defines the structure and layout of the model's response, ensuring that outputs follow a standardized format and remain consistent across different instances. At all stages, our prompts strictly adhere to the predefined guidelines and are adjusted according to specific task requirements. For example, during the outline generation stage, the LLM is instructed to output strictly in JSON format and provide brief explanations for the settings of each section. However, during the survey generation stage, such explanations are not required. Overall, this structured prompt design enables greater control over the model's outputs and minimizes variability across tasks, which helps ensure that the generated content remains practical and consistent. While we acknowledge that different prompts may yield varying results, this issue falls outside the scope of this study. Due to space limitations, the prompt is not included in the main paper and can be found in the supplementary files
attached.

\printcredits

% %% Loading bibliography style file
% \bibliographystyle{elsarticle-num-names}
% \clearpage
% % Ensure the reference list starts on a clean page
% \newpage
% \thispagestyle{empty}  % Remove the header/footer
% % Loading bibliography database
% \bibliography{cas-refs}

\begin{thebibliography}{79}
\expandafter\ifx\csname natexlab\endcsname\relax\def\natexlab#1{#1}\fi
\providecommand{\url}[1]{\texttt{#1}}
\providecommand{\href}[2]{#2}
\providecommand{\path}[1]{#1}
\providecommand{\DOIprefix}{doi:}
\providecommand{\ArXivprefix}{arXiv:}
\providecommand{\URLprefix}{URL: }
\providecommand{\Pubmedprefix}{pmid:}
\providecommand{\doi}[1]{\href{http://dx.doi.org/#1}{\path{#1}}}
\providecommand{\Pubmed}[1]{\href{pmid:#1}{\path{#1}}}
\providecommand{\bibinfo}[2]{#2}
\ifx\xfnm\relax \def\xfnm[#1]{\unskip,\space#1}\fi
%Type = Article
\bibitem[{Lawrence(2001)}]{bib1}
\bibinfo{author}{S.~Lawrence},
\newblock \bibinfo{title}{Free online availability substantially increases a paper's impact},
\newblock \bibinfo{journal}{Nature} \bibinfo{volume}{411} (\bibinfo{year}{2001}) \bibinfo{pages}{521}. \DOIprefix\doi{10.1038/35079151}.
%Type = Article
\bibitem[{Chu and Evans(2021)}]{bib2}
\bibinfo{author}{J.~S.~G. Chu}, \bibinfo{author}{J.~A. Evans},
\newblock \bibinfo{title}{Slowed canonical progress in large fields of science},
\newblock \bibinfo{journal}{Proceedings of the National Academy of Sciences} \bibinfo{volume}{118} (\bibinfo{year}{2021}) \bibinfo{pages}{e2021636118}. \DOIprefix\doi{10.1073/pnas.2021636118}.
%Type = Article
\bibitem[{Park et~al.(2023)Park, Leahey, and Funk}]{bib3}
\bibinfo{author}{M.~Park}, \bibinfo{author}{E.~Leahey}, \bibinfo{author}{R.~J. Funk},
\newblock \bibinfo{title}{Papers and patents are becoming less disruptive over time},
\newblock \bibinfo{journal}{Nature} \bibinfo{volume}{613} (\bibinfo{year}{2023}) \bibinfo{pages}{138--144}. \DOIprefix\doi{10.1038/s41586-022-05543-x}.
%Type = Article
\bibitem[{Lok(2010)}]{bib4}
\bibinfo{author}{C.~Lok},
\newblock \bibinfo{title}{Literature mining: Speed reading},
\newblock \bibinfo{journal}{Nature} \bibinfo{volume}{463} (\bibinfo{year}{2010}) \bibinfo{pages}{416--418}. \DOIprefix\doi{10.1038/463416a}.
%Type = Article
\bibitem[{Palmatier et~al.(2018)Palmatier, Houston, and Hulland}]{bib5}
\bibinfo{author}{R.~W. Palmatier}, \bibinfo{author}{M.~B. Houston}, \bibinfo{author}{J.~Hulland},
\newblock \bibinfo{title}{Review articles: purpose, process, and structure},
\newblock \bibinfo{journal}{Journal of the Academy of Marketing Science} \bibinfo{volume}{46} (\bibinfo{year}{2018}) \bibinfo{pages}{1--5}. \DOIprefix\doi{10.1007/s11747-017-0563-4}.
%Type = Article
\bibitem[{Torraco(2016)}]{bib6}
\bibinfo{author}{R.~J. Torraco},
\newblock \bibinfo{title}{Writing integrative literature reviews: Using the past and present to explore the future},
\newblock \bibinfo{journal}{Human Resource Development Review} \bibinfo{volume}{15} (\bibinfo{year}{2016}) \bibinfo{pages}{404--428}. \DOIprefix\doi{10.1177/1534484316671606}.
%Type = Article
\bibitem[{Paul and Criado(2020)}]{PAUL2020101717}
\bibinfo{author}{J.~Paul}, \bibinfo{author}{A.~R. Criado},
\newblock \bibinfo{title}{The art of writing literature review: What do we know and what do we need to know?},
\newblock \bibinfo{journal}{International Business Review} \bibinfo{volume}{29} (\bibinfo{year}{2020}) \bibinfo{pages}{101717}. \DOIprefix\doi{https://doi.org/10.1016/j.ibusrev.2020.101717}.
%Type = Article
\bibitem[{Torraco(2005)}]{bib7}
\bibinfo{author}{R.~J. Torraco},
\newblock \bibinfo{title}{Writing integrative literature reviews: Guidelines and examples},
\newblock \bibinfo{journal}{Human Resource Development Review} \bibinfo{volume}{4} (\bibinfo{year}{2005}) \bibinfo{pages}{356--367}. \DOIprefix\doi{10.1177/1534484305278283}.
%Type = Article
\bibitem[{Wee and Banister(2016)}]{Wee03032016}
\bibinfo{author}{B.~V. Wee}, \bibinfo{author}{D.~Banister},
\newblock \bibinfo{title}{How to write a literature review paper?},
\newblock \bibinfo{journal}{Transport Reviews} \bibinfo{volume}{36} (\bibinfo{year}{2016}) \bibinfo{pages}{278--288}. \DOIprefix\doi{10.1080/01441647.2015.1065456}.
%Type = Inproceedings
\bibitem[{Brown et~al.(2020)Brown, Mann, Ryder, Subbiah, Kaplan, Dhariwal, Neelakantan, Shyam, Sastry, Askell, Agarwal, Herbert-Voss, Krueger, Henighan, Child, Ramesh, Ziegler, Wu, Winter, Hesse, Chen, Sigler, Litwin, Gray, Chess, Clark, Berner, McCandlish, Radford, Sutskever, and Amodei}]{brown2020language}
\bibinfo{author}{T.~Brown}, \bibinfo{author}{B.~Mann}, \bibinfo{author}{N.~Ryder}, \bibinfo{author}{M.~Subbiah}, \bibinfo{author}{J.~D. Kaplan}, \bibinfo{author}{P.~Dhariwal}, \bibinfo{author}{A.~Neelakantan}, \bibinfo{author}{P.~Shyam}, \bibinfo{author}{G.~Sastry}, \bibinfo{author}{A.~Askell}, \bibinfo{author}{S.~Agarwal}, \bibinfo{author}{A.~Herbert-Voss}, \bibinfo{author}{G.~Krueger}, \bibinfo{author}{T.~Henighan}, \bibinfo{author}{R.~Child}, \bibinfo{author}{A.~Ramesh}, \bibinfo{author}{D.~Ziegler}, \bibinfo{author}{J.~Wu}, \bibinfo{author}{C.~Winter}, \bibinfo{author}{C.~Hesse}, \bibinfo{author}{M.~Chen}, \bibinfo{author}{E.~Sigler}, \bibinfo{author}{M.~Litwin}, \bibinfo{author}{S.~Gray}, \bibinfo{author}{B.~Chess}, \bibinfo{author}{J.~Clark}, \bibinfo{author}{C.~Berner}, \bibinfo{author}{S.~McCandlish}, \bibinfo{author}{A.~Radford}, \bibinfo{author}{I.~Sutskever}, \bibinfo{author}{D.~Amodei},
\newblock \bibinfo{title}{Language models are few-shot learners},
\newblock in: \bibinfo{booktitle}{Advances in Neural Information Processing Systems}, volume~\bibinfo{volume}{33}, \bibinfo{year}{2020}, pp. \bibinfo{pages}{1877--1901}. \URLprefix \url{https://proceedings.neurips.cc/paper_files/paper/2020/file/1457c0d6bfcb4967418bfb8ac142f64a-Paper.pdf}.
%Type = Article
\bibitem[{Wu et~al.(2025)Wu, Ma, Luo, Li, Shi, Chang, Lin, Luo, Pei, Du, Zhao, and Gong}]{wu2024automated}
\bibinfo{author}{S.~Wu}, \bibinfo{author}{X.~Ma}, \bibinfo{author}{D.~Luo}, \bibinfo{author}{L.~Li}, \bibinfo{author}{X.~Shi}, \bibinfo{author}{X.~Chang}, \bibinfo{author}{X.~Lin}, \bibinfo{author}{R.~Luo}, \bibinfo{author}{C.~Pei}, \bibinfo{author}{C.~Du}, \bibinfo{author}{Z.-J. Zhao}, \bibinfo{author}{J.~Gong},
\newblock \bibinfo{title}{Automated literature research and review generation method based on large language models},
\newblock \bibinfo{journal}{National Science Review}  (\bibinfo{year}{2025}). \DOIprefix\doi{10.1093/nsr/nwaf169}.
%Type = Inproceedings
\bibitem[{Lai et~al.(2025)Lai, Wu, Wang, Hu, and Zheng}]{lai2024instruct}
\bibinfo{author}{Y.~Lai}, \bibinfo{author}{Y.~Wu}, \bibinfo{author}{Y.~Wang}, \bibinfo{author}{W.~Hu}, \bibinfo{author}{C.~Zheng},
\newblock \bibinfo{title}{Instruct large language models to generate scientific literature survey step by step},
\newblock in: \bibinfo{booktitle}{Natural Language Processing and Chinese Computing}, \bibinfo{address}{Singapore}, \bibinfo{year}{2025}, pp. \bibinfo{pages}{484--496}. \DOIprefix\doi{10.1007/978-981-97-9443-0_43}.
%Type = Inproceedings
\bibitem[{Wang et~al.(2024)Wang, Guo, Yao, Zhang, Zhang, Wu, Zhang, Dai, Zhang, Wen, Ye, Zhang, and Zhang}]{wang2024AutoSurvey}
\bibinfo{author}{Y.~Wang}, \bibinfo{author}{Q.~Guo}, \bibinfo{author}{W.~Yao}, \bibinfo{author}{H.~Zhang}, \bibinfo{author}{X.~Zhang}, \bibinfo{author}{Z.~Wu}, \bibinfo{author}{M.~Zhang}, \bibinfo{author}{X.~Dai}, \bibinfo{author}{M.~Zhang}, \bibinfo{author}{Q.~Wen}, \bibinfo{author}{W.~Ye}, \bibinfo{author}{S.~Zhang}, \bibinfo{author}{Y.~Zhang},
\newblock \bibinfo{title}{AutoSurvey: Large language models can automatically write surveys},
\newblock in: \bibinfo{booktitle}{Advances in Neural Information Processing Systems}, volume~\bibinfo{volume}{37}, \bibinfo{year}{2024}, pp. \bibinfo{pages}{115119--115145}. \URLprefix \url{https://proceedings.neurips.cc/paper_files/paper/2024/file/d07a9fc7da2e2ec0574c38d5f504d105-Paper-Conference.pdf}. \DOIprefix\doi{10.52202/079017-3655}.
%Type = Misc
\bibitem[{Liang et~al.(2025)Liang, Yang, Wang, Tang, Zheng, Song, Lin, Yang, Niu, Wang, Tang, Xiong, Mao, and li}]{liang2025surveyx}
\bibinfo{author}{X.~Liang}, \bibinfo{author}{J.~Yang}, \bibinfo{author}{Y.~Wang}, \bibinfo{author}{C.~Tang}, \bibinfo{author}{Z.~Zheng}, \bibinfo{author}{S.~Song}, \bibinfo{author}{Z.~Lin}, \bibinfo{author}{Y.~Yang}, \bibinfo{author}{S.~Niu}, \bibinfo{author}{H.~Wang}, \bibinfo{author}{B.~Tang}, \bibinfo{author}{F.~Xiong}, \bibinfo{author}{K.~Mao}, \bibinfo{author}{Z.~li}, \bibinfo{title}{Surveyx: Academic survey automation via large language models}, \bibinfo{year}{2025}. \URLprefix \url{https://arxiv.org/abs/2502.14776}. \href{http://arxiv.org/abs/2502.14776}{{\tt arXiv:2502.14776}}.
%Type = Misc
\bibitem[{Izacard et~al.(2022)Izacard, Lewis, Lomeli, Hosseini, Petroni, Schick, Dwivedi-Yu, Joulin, Riedel, and Grave}]{izacard2022few}
\bibinfo{author}{G.~Izacard}, \bibinfo{author}{P.~Lewis}, \bibinfo{author}{M.~Lomeli}, \bibinfo{author}{L.~Hosseini}, \bibinfo{author}{F.~Petroni}, \bibinfo{author}{T.~Schick}, \bibinfo{author}{J.~Dwivedi-Yu}, \bibinfo{author}{A.~Joulin}, \bibinfo{author}{S.~Riedel}, \bibinfo{author}{E.~Grave}, \bibinfo{title}{Atlas: Few-shot learning with retrieval augmented language models}, \bibinfo{year}{2022}. \URLprefix \url{https://arxiv.org/abs/2208.03299}. \href{http://arxiv.org/abs/2208.03299}{{\tt arXiv:2208.03299}}.
%Type = Inproceedings
\bibitem[{Borgeaud et~al.(2022)Borgeaud, Mensch, Hoffmann, Cai, Rutherford, Millican, Van Den~Driessche, Lespiau, Damoc, Clark, De~Las~Casas, Guy, Menick, Ring, Hennigan, Huang, Maggiore, Jones, Cassirer, Brock, Paganini, Irving, Vinyals, Osindero, Simonyan, Rae, Elsen, and Sifre}]{borgeaud2022improving}
\bibinfo{author}{S.~Borgeaud}, \bibinfo{author}{A.~Mensch}, \bibinfo{author}{J.~Hoffmann}, \bibinfo{author}{T.~Cai}, \bibinfo{author}{E.~Rutherford}, \bibinfo{author}{K.~Millican}, \bibinfo{author}{G.~B. Van Den~Driessche}, \bibinfo{author}{J.-B. Lespiau}, \bibinfo{author}{B.~Damoc}, \bibinfo{author}{A.~Clark}, \bibinfo{author}{D.~De~Las~Casas}, \bibinfo{author}{A.~Guy}, \bibinfo{author}{J.~Menick}, \bibinfo{author}{R.~Ring}, \bibinfo{author}{T.~Hennigan}, \bibinfo{author}{S.~Huang}, \bibinfo{author}{L.~Maggiore}, \bibinfo{author}{C.~Jones}, \bibinfo{author}{A.~Cassirer}, \bibinfo{author}{A.~Brock}, \bibinfo{author}{M.~Paganini}, \bibinfo{author}{G.~Irving}, \bibinfo{author}{O.~Vinyals}, \bibinfo{author}{S.~Osindero}, \bibinfo{author}{K.~Simonyan}, \bibinfo{author}{J.~Rae}, \bibinfo{author}{E.~Elsen}, \bibinfo{author}{L.~Sifre},
\newblock \bibinfo{title}{Improving language models by retrieving from trillions of tokens},
\newblock in: \bibinfo{booktitle}{Proceedings of the 39th International Conference on Machine Learning}, volume \bibinfo{volume}{162} of \textit{\bibinfo{series}{Proceedings of Machine Learning Research}}, \bibinfo{year}{2022}, pp. \bibinfo{pages}{2206--2240}. \URLprefix \url{https://proceedings.mlr.press/v162/borgeaud22a.html}.
%Type = Misc
\bibitem[{Gao et~al.(2024)Gao, Xiong, Gao, Jia, Pan, Bi, Dai, Sun, Wang, and Wang}]{gao2023retrieval}
\bibinfo{author}{Y.~Gao}, \bibinfo{author}{Y.~Xiong}, \bibinfo{author}{X.~Gao}, \bibinfo{author}{K.~Jia}, \bibinfo{author}{J.~Pan}, \bibinfo{author}{Y.~Bi}, \bibinfo{author}{Y.~Dai}, \bibinfo{author}{J.~Sun}, \bibinfo{author}{M.~Wang}, \bibinfo{author}{H.~Wang}, \bibinfo{title}{Retrieval-augmented generation for large language models: A survey}, \bibinfo{year}{2024}. \URLprefix \url{https://arxiv.org/abs/2312.10997}. \href{http://arxiv.org/abs/2312.10997}{{\tt arXiv:2312.10997}}.
%Type = Inproceedings
\bibitem[{Jiang et~al.(2023)Jiang, Xu, Gao, Sun, Liu, Dwivedi-Yu, Yang, Callan, and Neubig}]{jiang-etal-2023-active}
\bibinfo{author}{Z.~Jiang}, \bibinfo{author}{F.~Xu}, \bibinfo{author}{L.~Gao}, \bibinfo{author}{Z.~Sun}, \bibinfo{author}{Q.~Liu}, \bibinfo{author}{J.~Dwivedi-Yu}, \bibinfo{author}{Y.~Yang}, \bibinfo{author}{J.~Callan}, \bibinfo{author}{G.~Neubig},
\newblock \bibinfo{title}{Active retrieval augmented generation},
\newblock in: \bibinfo{booktitle}{Proceedings of the 2023 Conference on Empirical Methods in Natural Language Processing}, \bibinfo{address}{Singapore}, \bibinfo{year}{2023}, pp. \bibinfo{pages}{7969--7992}. \DOIprefix\doi{10.18653/v1/2023.emnlp-main.495}.
%Type = Inproceedings
\bibitem[{Bao et~al.(2025)Bao, Nayeem, Rafiei, and Zhang}]{bao-etal-2025-surveygen}
\bibinfo{author}{T.~Bao}, \bibinfo{author}{M.~T. Nayeem}, \bibinfo{author}{D.~Rafiei}, \bibinfo{author}{C.~Zhang},
\newblock \bibinfo{title}{{S}urvey{G}en: Quality-aware scientific survey generation with large language models},
\newblock in: \bibinfo{booktitle}{Proceedings of the 2025 Conference on Empirical Methods in Natural Language Processing}, \bibinfo{year}{2025}, pp. \bibinfo{pages}{2712--2736}. \DOIprefix\doi{10.18653/v1/2025.emnlp-main.136}.
%Type = Misc
\bibitem[{Yan et~al.(2025)Yan, Feng, Yuan, Xia, Wang, Zhang, and Bai}]{yan2025surveyforgeoutlineheuristicsmemorydriven}
\bibinfo{author}{X.~Yan}, \bibinfo{author}{S.~Feng}, \bibinfo{author}{J.~Yuan}, \bibinfo{author}{R.~Xia}, \bibinfo{author}{B.~Wang}, \bibinfo{author}{B.~Zhang}, \bibinfo{author}{L.~Bai}, \bibinfo{title}{Surveyforge: On the outline heuristics, memory-driven generation, and multi-dimensional evaluation for automated survey writing}, \bibinfo{year}{2025}. \URLprefix \url{https://arxiv.org/abs/2503.04629}. \href{http://arxiv.org/abs/2503.04629}{{\tt arXiv:2503.04629}}.
%Type = Inproceedings
\bibitem[{Tang et~al.(2025)Tang, Duan, and Cai}]{tang2024llms}
\bibinfo{author}{X.~Tang}, \bibinfo{author}{X.~Duan}, \bibinfo{author}{Z.~Cai},
\newblock \bibinfo{title}{Large language models for automated literature review: An evaluation of reference generation, abstract writing, and review composition},
\newblock in: \bibinfo{booktitle}{Proceedings of the 2025 Conference on Empirical Methods in Natural Language Processing}, \bibinfo{address}{Suzhou, China}, \bibinfo{year}{2025}, pp. \bibinfo{pages}{1602--1617}. \DOIprefix\doi{10.18653/v1/2025.emnlp-main.83}.
%Type = Article
\bibitem[{Paul and Criado(2020)}]{paul2020art}
\bibinfo{author}{J.~Paul}, \bibinfo{author}{A.~R. Criado},
\newblock \bibinfo{title}{The art of writing literature review: What do we know and what do we need to know?},
\newblock \bibinfo{journal}{International Business Review} \bibinfo{volume}{29} (\bibinfo{year}{2020}) \bibinfo{pages}{101717}. \DOIprefix\doi{https://doi.org/10.1016/j.ibusrev.2020.101717}.
%Type = Article
\bibitem[{Kanellos et~al.(2021)Kanellos, Vergoulis, Sacharidis, Dalamagas, and Vassiliou}]{kanellos2019impact}
\bibinfo{author}{I.~Kanellos}, \bibinfo{author}{T.~Vergoulis}, \bibinfo{author}{D.~Sacharidis}, \bibinfo{author}{T.~Dalamagas}, \bibinfo{author}{Y.~Vassiliou},
\newblock \bibinfo{title}{Impact-based ranking of scientific publications: A survey and experimental evaluation},
\newblock \bibinfo{journal}{IEEE Transactions on Knowledge and Data Engineering} \bibinfo{volume}{33} (\bibinfo{year}{2021}) \bibinfo{pages}{1567--1584}. \DOIprefix\doi{10.1109/TKDE.2019.2941206}.
%Type = Inproceedings
\bibitem[{Jha et~al.(2015)Jha, Finegan-Dollak, King, Coke, and Radev}]{jha-etal-2015-content}
\bibinfo{author}{R.~Jha}, \bibinfo{author}{C.~Finegan-Dollak}, \bibinfo{author}{B.~King}, \bibinfo{author}{R.~Coke}, \bibinfo{author}{D.~Radev},
\newblock \bibinfo{title}{Content models for survey generation: A factoid-based evaluation},
\newblock in: \bibinfo{booktitle}{Proceedings of the 53rd Annual Meeting of the Association for Computational Linguistics and the 7th International Joint Conference on Natural Language Processing (Volume 1: Long Papers)}, \bibinfo{address}{Beijing, China}, \bibinfo{year}{2015}, pp. \bibinfo{pages}{441--450}. \DOIprefix\doi{10.3115/v1/P15-1043}.
%Type = Inproceedings
\bibitem[{Hu and Wan(2014)}]{hu-wan-2014-automatic}
\bibinfo{author}{Y.~Hu}, \bibinfo{author}{X.~Wan},
\newblock \bibinfo{title}{Automatic generation of related work sections in scientific papers: An optimization approach},
\newblock in: \bibinfo{booktitle}{Proceedings of the 2014 Conference on Empirical Methods in Natural Language Processing ({EMNLP})}, \bibinfo{address}{Doha, Qatar}, \bibinfo{year}{2014}, pp. \bibinfo{pages}{1624--1633}. \DOIprefix\doi{10.3115/v1/D14-1170}.
%Type = Inproceedings
\bibitem[{Kasanishi et~al.(2023)Kasanishi, Isonuma, Mori, and Sakata}]{kasanishi-etal-2023-scireviewgen}
\bibinfo{author}{T.~Kasanishi}, \bibinfo{author}{M.~Isonuma}, \bibinfo{author}{J.~Mori}, \bibinfo{author}{I.~Sakata},
\newblock \bibinfo{title}{{S}ci{R}eview{G}en: A large-scale dataset for automatic literature review generation},
\newblock in: \bibinfo{booktitle}{Findings of the Association for Computational Linguistics: ACL 2023}, \bibinfo{address}{Toronto, Canada}, \bibinfo{year}{2023}, pp. \bibinfo{pages}{6695--6715}. \DOIprefix\doi{10.18653/v1/2023.findings-acl.418}.
%Type = Misc
\bibitem[{{OpenAI}(2025)}]{deepresearch2025}
\bibinfo{author}{{OpenAI}}, \bibinfo{title}{DeepResearch}, \bibinfo{howpublished}{\url{https://openai.com/index/introducing-deep-research/}}, \bibinfo{year}{2025}. \bibinfo{note}{Accessed: 2025-07-07}.
%Type = Misc
\bibitem[{{ScholarAI}(2025)}]{scholarai2025}
\bibinfo{author}{{ScholarAI}}, \bibinfo{title}{Scholarai}, \bibinfo{howpublished}{\url{https://scholarai.io/}}, \bibinfo{year}{2025}. \bibinfo{note}{Accessed: 2025-07-07}.
%Type = Misc
\bibitem[{Wen et~al.(2025)Wen, Cao, Wang, Guo, Yang, and Liu}]{wen2025interactivesurveyllmbasedpersonalizedinteractive}
\bibinfo{author}{Z.~Wen}, \bibinfo{author}{J.~Cao}, \bibinfo{author}{Z.~Wang}, \bibinfo{author}{B.~Guo}, \bibinfo{author}{R.~Yang}, \bibinfo{author}{S.~Liu}, \bibinfo{title}{Interactivesurvey: An llm-based personalized and interactive survey paper generation system}, \bibinfo{year}{2025}. \URLprefix \url{https://arxiv.org/abs/2504.08762}. \href{http://arxiv.org/abs/2504.08762}{{\tt arXiv:2504.08762}}.
%Type = Inproceedings
\bibitem[{Fok et~al.(2025)Fok, Siu, and Weld}]{10.1145/3706598.3714047}
\bibinfo{author}{R.~Fok}, \bibinfo{author}{A.~Siu}, \bibinfo{author}{D.~S. Weld},
\newblock \bibinfo{title}{Toward living narrative reviews: An empirical study of the processes and challenges in updating survey articles in computing research},
\newblock in: \bibinfo{booktitle}{Proceedings of the 2025 CHI Conference on Human Factors in Computing Systems}, \bibinfo{year}{2025}, pp. \bibinfo{pages}{1--10}.
%Type = Misc
\bibitem[{Qiu et~al.(2025)Qiu, Chen, Su, Yen, and Shen}]{qiu2025completingsystematicreviewhours}
\bibinfo{author}{R.~Qiu}, \bibinfo{author}{S.~Chen}, \bibinfo{author}{Y.~Su}, \bibinfo{author}{P.-Y. Yen}, \bibinfo{author}{H.-W. Shen}, \bibinfo{title}{Completing a systematic review in hours instead of months with interactive ai agents}, \bibinfo{year}{2025}. \URLprefix \url{https://arxiv.org/abs/2504.14822}. \href{http://arxiv.org/abs/2504.14822}{{\tt arXiv:2504.14822}}.
%Type = Misc
\bibitem[{Nguye et~al.(2025)Nguye, Nguyen, T., Dang, Dong, and Le}]{nguye2025surveyg}
\bibinfo{author}{M.-A. Nguye}, \bibinfo{author}{M.-D. Nguyen}, \bibinfo{author}{H.~L.~N. T.}, \bibinfo{author}{K.~H. Dang}, \bibinfo{author}{N.~T. Dong}, \bibinfo{author}{D.~D. Le}, \bibinfo{title}{Surveyg: A multi-agent llm framework with hierarchical citation graph for automated survey generation}, \bibinfo{year}{2025}. \URLprefix \url{https://arxiv.org/abs/2510.07733}. \href{http://arxiv.org/abs/2510.07733}{{\tt arXiv:2510.07733}}.
%Type = Article
\bibitem[{Asai et~al.(2026)Asai, He, Shao, Shi, Singh, Chang, Lo, Soldaini, Feldman, D’Arcy et~al.}]{asai2026synthesizing}
\bibinfo{author}{A.~Asai}, \bibinfo{author}{J.~He}, \bibinfo{author}{R.~Shao}, \bibinfo{author}{W.~Shi}, \bibinfo{author}{A.~Singh}, \bibinfo{author}{J.~C. Chang}, \bibinfo{author}{K.~Lo}, \bibinfo{author}{L.~Soldaini}, \bibinfo{author}{S.~Feldman}, \bibinfo{author}{M.~D’Arcy}, et~al.,
\newblock \bibinfo{title}{Synthesizing scientific literature with retrieval-augmented language models},
\newblock \bibinfo{journal}{Nature}  (\bibinfo{year}{2026}) \bibinfo{pages}{1--7}.
%Type = Inproceedings
\bibitem[{Singh et~al.(2025)Singh, Chang, Haddad, Naik, Hwang, Kinney, Weld, Downey, and Feldman}]{singh2025ai2}
\bibinfo{author}{A.~Singh}, \bibinfo{author}{J.~C. Chang}, \bibinfo{author}{D.~Haddad}, \bibinfo{author}{A.~Naik}, \bibinfo{author}{J.~D. Hwang}, \bibinfo{author}{R.~Kinney}, \bibinfo{author}{D.~S. Weld}, \bibinfo{author}{D.~Downey}, \bibinfo{author}{S.~Feldman},
\newblock \bibinfo{title}{Ai2 scholar qa: Organized literature synthesis with attribution},
\newblock in: \bibinfo{booktitle}{Proceedings of the 63rd Annual Meeting of the Association for Computational Linguistics (Volume 3: System Demonstrations)}, \bibinfo{year}{2025}, pp. \bibinfo{pages}{513--523}.
%Type = Inproceedings
\bibitem[{Weng et~al.(2025)Weng, Zhu, Bao, Zhang, Wang, Zhang, and Yang}]{weng2025cycleresearcher}
\bibinfo{author}{Y.~Weng}, \bibinfo{author}{M.~Zhu}, \bibinfo{author}{G.~Bao}, \bibinfo{author}{H.~Zhang}, \bibinfo{author}{J.~Wang}, \bibinfo{author}{Y.~Zhang}, \bibinfo{author}{L.~Yang},
\newblock \bibinfo{title}{Cycleresearcher: Improving automated research via automated review},
\newblock in: \bibinfo{booktitle}{International Conference on Learning Representations}, volume \bibinfo{volume}{2025}, \bibinfo{year}{2025}, pp. \bibinfo{pages}{3669--3709}.
%Type = Inproceedings
\bibitem[{Guo et~al.(2024)Guo, Chen, Wang, Chang, Pei, Chawla, Wiest, and Zhang}]{10.24963/ijcai.2024/890}
\bibinfo{author}{T.~Guo}, \bibinfo{author}{X.~Chen}, \bibinfo{author}{Y.~Wang}, \bibinfo{author}{R.~Chang}, \bibinfo{author}{S.~Pei}, \bibinfo{author}{N.~V. Chawla}, \bibinfo{author}{O.~Wiest}, \bibinfo{author}{X.~Zhang},
\newblock \bibinfo{title}{Large language model based multi-agents: a survey of progress and challenges},
\newblock in: \bibinfo{booktitle}{Proceedings of the Thirty-Third International Joint Conference on Artificial Intelligence}, IJCAI '24, \bibinfo{year}{2024}, pp. \bibinfo{pages}{8048 -- 8057}. \DOIprefix\doi{10.24963/ijcai.2024/890}.
%Type = Inbook
\bibitem[{Zhang et~al.(2021)Zhang, Yang, and Ba{\c{s}}ar}]{Zhang2021}
\bibinfo{author}{K.~Zhang}, \bibinfo{author}{Z.~Yang}, \bibinfo{author}{T.~Ba{\c{s}}ar}, \bibinfo{title}{Multi-Agent Reinforcement Learning: A Selective Overview of Theories and Algorithms}, \bibinfo{address}{Cham}, \bibinfo{year}{2021}, pp. \bibinfo{pages}{321--384}. \DOIprefix\doi{10.1007/978-3-030-60990-0_12}.
%Type = Article
\bibitem[{Zhao et~al.(2024)Zhao, Huang, Xu, Lin, Liu, and Huang}]{Zhao_Huang_Xu_Lin_Liu_Huang_2024}
\bibinfo{author}{A.~Zhao}, \bibinfo{author}{D.~Huang}, \bibinfo{author}{Q.~Xu}, \bibinfo{author}{M.~Lin}, \bibinfo{author}{Y.-J. Liu}, \bibinfo{author}{G.~Huang},
\newblock \bibinfo{title}{Expel: Llm agents are experiential learners},
\newblock \bibinfo{journal}{Proceedings of the AAAI Conference on Artificial Intelligence} \bibinfo{volume}{38} (\bibinfo{year}{2024}) \bibinfo{pages}{19632--19642}. \DOIprefix\doi{10.1609/aaai.v38i17.29936}.
%Type = Article
\bibitem[{Saadaoui and Alonso(2025)}]{saadaoui2025coordinated}
\bibinfo{author}{S.~Saadaoui}, \bibinfo{author}{E.~Alonso},
\newblock \bibinfo{title}{Coordinated llm multi-agent systems for collaborative question-answer generation},
\newblock \bibinfo{journal}{Knowledge-Based Systems} \bibinfo{volume}{330} (\bibinfo{year}{2025}) \bibinfo{pages}{114627}. \DOIprefix\doi{https://doi.org/10.1016/j.knosys.2025.114627}.
%Type = Misc
\bibitem[{Li et~al.(2025)Li, Lai, Li, Ren, Zhang, Kang, Wang, Li, Zhang, Ma, and Liu}]{li2025agenthospitalsimulacrumhospital}
\bibinfo{author}{J.~Li}, \bibinfo{author}{Y.~Lai}, \bibinfo{author}{W.~Li}, \bibinfo{author}{J.~Ren}, \bibinfo{author}{M.~Zhang}, \bibinfo{author}{X.~Kang}, \bibinfo{author}{S.~Wang}, \bibinfo{author}{P.~Li}, \bibinfo{author}{Y.-Q. Zhang}, \bibinfo{author}{W.~Ma}, \bibinfo{author}{Y.~Liu}, \bibinfo{title}{Agent hospital: A simulacrum of hospital with evolvable medical agents}, \bibinfo{year}{2025}. \URLprefix \url{https://arxiv.org/abs/2405.02957}. \href{http://arxiv.org/abs/2405.02957}{{\tt arXiv:2405.02957}}.
%Type = Misc
\bibitem[{Wang et~al.(2023)Wang, Xie, Jiang, Mandlekar, Xiao, Zhu, Fan, and Anandkumar}]{wang2023voyageropenendedembodiedagent}
\bibinfo{author}{G.~Wang}, \bibinfo{author}{Y.~Xie}, \bibinfo{author}{Y.~Jiang}, \bibinfo{author}{A.~Mandlekar}, \bibinfo{author}{C.~Xiao}, \bibinfo{author}{Y.~Zhu}, \bibinfo{author}{L.~Fan}, \bibinfo{author}{A.~Anandkumar}, \bibinfo{title}{Voyager: An open-ended embodied agent with large language models}, \bibinfo{year}{2023}. \URLprefix \url{https://arxiv.org/abs/2305.16291}. \href{http://arxiv.org/abs/2305.16291}{{\tt arXiv:2305.16291}}.
%Type = Article
\bibitem[{Han et~al.(2025)Han, Yang, Lin, and Qin}]{han2025plug}
\bibinfo{author}{Q.~Han}, \bibinfo{author}{Z.~Yang}, \bibinfo{author}{H.~Lin}, \bibinfo{author}{T.~Qin},
\newblock \bibinfo{title}{A plug-and-play knowledge-enhanced module for medical reports generation},
\newblock \bibinfo{journal}{Knowledge-Based Systems} \bibinfo{volume}{309} (\bibinfo{year}{2025}) \bibinfo{pages}{112805}. \DOIprefix\doi{https://doi.org/10.1016/j.knosys.2024.112805}.
%Type = Inproceedings
\bibitem[{Islam et~al.(2024)Islam, Ali, and Parvez}]{islam-etal-2024-mapcoder}
\bibinfo{author}{M.~A. Islam}, \bibinfo{author}{M.~E. Ali}, \bibinfo{author}{M.~R. Parvez},
\newblock \bibinfo{title}{{M}ap{C}oder: Multi-agent code generation for competitive problem solving},
\newblock in: \bibinfo{booktitle}{Proceedings of the 62nd Annual Meeting of the Association for Computational Linguistics (Volume 1: Long Papers)}, \bibinfo{address}{Bangkok, Thailand}, \bibinfo{year}{2024}, pp. \bibinfo{pages}{4912--4944}. \DOIprefix\doi{10.18653/v1/2024.acl-long.269}.
%Type = Article
\bibitem[{Hong et~al.(2023)Hong, Zhuge, Chen, Zheng, Cheng, Zhang, Wang, Wang, Yau, Lin et~al.}]{hong2024metagpt}
\bibinfo{author}{S.~Hong}, \bibinfo{author}{M.~Zhuge}, \bibinfo{author}{J.~Chen}, \bibinfo{author}{X.~Zheng}, \bibinfo{author}{Y.~Cheng}, \bibinfo{author}{C.~Zhang}, \bibinfo{author}{J.~Wang}, \bibinfo{author}{Z.~Wang}, \bibinfo{author}{S.~K.~S. Yau}, \bibinfo{author}{Z.~Lin}, et~al.,
\newblock \bibinfo{title}{Metagpt: Meta programming for a multi-agent collaborative framework},
\newblock \bibinfo{journal}{arXiv preprint arXiv:2308.00352}  (\bibinfo{year}{2023}).
%Type = Misc
\bibitem[{Gao et~al.(2025)Gao, Lan, Lu, Mao, Piao, Wang, Jin, and Li}]{gao2025s3socialnetworksimulationlarge}
\bibinfo{author}{C.~Gao}, \bibinfo{author}{X.~Lan}, \bibinfo{author}{Z.~Lu}, \bibinfo{author}{J.~Mao}, \bibinfo{author}{J.~Piao}, \bibinfo{author}{H.~Wang}, \bibinfo{author}{D.~Jin}, \bibinfo{author}{Y.~Li}, \bibinfo{title}{S$^3$: Social-network simulation system with large language model-empowered agents}, \bibinfo{year}{2025}. \URLprefix \url{https://arxiv.org/abs/2307.14984}. \href{http://arxiv.org/abs/2307.14984}{{\tt arXiv:2307.14984}}.
%Type = Misc
\bibitem[{Huot et~al.(2025)Huot, Amplayo, Palomaki, Jakobovits, Clark, and Lapata}]{huot2025agentsroomnarrativegeneration}
\bibinfo{author}{F.~Huot}, \bibinfo{author}{R.~K. Amplayo}, \bibinfo{author}{J.~Palomaki}, \bibinfo{author}{A.~S. Jakobovits}, \bibinfo{author}{E.~Clark}, \bibinfo{author}{M.~Lapata}, \bibinfo{title}{Agents' room: Narrative generation through multi-step collaboration}, \bibinfo{year}{2025}. \URLprefix \url{https://arxiv.org/abs/2410.02603}. \href{http://arxiv.org/abs/2410.02603}{{\tt arXiv:2410.02603}}.
%Type = Misc
\bibitem[{Wang et~al.(2024)Wang, Ni, Liu, Lu, Chen, Feng, Wei, Qu, Alinejad-Rokny, Lin, and Yang}]{wang2024autopatentmultiagentframeworkautomatic}
\bibinfo{author}{Q.~Wang}, \bibinfo{author}{S.~Ni}, \bibinfo{author}{H.~Liu}, \bibinfo{author}{S.~Lu}, \bibinfo{author}{G.~Chen}, \bibinfo{author}{X.~Feng}, \bibinfo{author}{C.~Wei}, \bibinfo{author}{Q.~Qu}, \bibinfo{author}{H.~Alinejad-Rokny}, \bibinfo{author}{Y.~Lin}, \bibinfo{author}{M.~Yang}, \bibinfo{title}{Autopatent: A multi-agent framework for automatic patent generation}, \bibinfo{year}{2024}. \URLprefix \url{https://arxiv.org/abs/2412.09796}. \href{http://arxiv.org/abs/2412.09796}{{\tt arXiv:2412.09796}}.
%Type = Inproceedings
\bibitem[{Shao et~al.(2024)Shao, Jiang, Kanell, Xu, Khattab, and Lam}]{shao-etal-2024-assisting}
\bibinfo{author}{Y.~Shao}, \bibinfo{author}{Y.~Jiang}, \bibinfo{author}{T.~Kanell}, \bibinfo{author}{P.~Xu}, \bibinfo{author}{O.~Khattab}, \bibinfo{author}{M.~Lam},
\newblock \bibinfo{title}{Assisting in writing {W}ikipedia-like articles from scratch with large language models},
\newblock in: \bibinfo{booktitle}{Proceedings of the 2024 Conference of the North American Chapter of the Association for Computational Linguistics: Human Language Technologies (Volume 1: Long Papers)}, \bibinfo{address}{Mexico City, Mexico}, \bibinfo{year}{2024}, pp. \bibinfo{pages}{6252--6278}. \DOIprefix\doi{10.18653/v1/2024.naacl-long.347}.
%Type = Misc
\bibitem[{D'Arcy et~al.(2024)D'Arcy, Hope, Birnbaum, and Downey}]{darcy2024margmultiagentreviewgeneration}
\bibinfo{author}{M.~D'Arcy}, \bibinfo{author}{T.~Hope}, \bibinfo{author}{L.~Birnbaum}, \bibinfo{author}{D.~Downey}, \bibinfo{title}{Marg: Multi-agent review generation for scientific papers}, \bibinfo{year}{2024}. \URLprefix \url{https://arxiv.org/abs/2401.04259}. \href{http://arxiv.org/abs/2401.04259}{{\tt arXiv:2401.04259}}.
%Type = Inproceedings
\bibitem[{Xia et~al.(2025)Xia, Peng, Qi, Xu, Li, Lei, and Wang}]{xia2025storywriter}
\bibinfo{author}{H.~Xia}, \bibinfo{author}{H.~Peng}, \bibinfo{author}{Y.~Qi}, \bibinfo{author}{B.~Xu}, \bibinfo{author}{J.~Li}, \bibinfo{author}{H.~Lei}, \bibinfo{author}{X.~Wang},
\newblock \bibinfo{title}{Storywriter: A multi-agent framework for long story generation},
\newblock in: \bibinfo{booktitle}{Proceedings of the 34th ACM International Conference on Information and Knowledge Management}, \bibinfo{address}{New York, NY, USA}, \bibinfo{year}{2025}, p. \bibinfo{pages}{6559–6563}. \DOIprefix\doi{10.1145/3746252.3761616}.
%Type = Inproceedings
\bibitem[{Yang et~al.(2025)Yang, Zeng, Rao, and Zhang}]{yang2025knowing}
\bibinfo{author}{D.~Yang}, \bibinfo{author}{L.~Zeng}, \bibinfo{author}{J.~Rao}, \bibinfo{author}{Y.~Zhang},
\newblock \bibinfo{title}{Knowing you don't know: Learning when to continue search in multi-round rag through self-practicing},
\newblock in: \bibinfo{booktitle}{Proceedings of the 48th International ACM SIGIR Conference on Research and Development in Information Retrieval}, \bibinfo{year}{2025}, pp. \bibinfo{pages}{1305--1315}.
%Type = Inproceedings
\bibitem[{Dong et~al.(2025)Dong, Jin, Li, Zhu, Dou, and Wen}]{dong2025rag}
\bibinfo{author}{G.~Dong}, \bibinfo{author}{J.~Jin}, \bibinfo{author}{X.~Li}, \bibinfo{author}{Y.~Zhu}, \bibinfo{author}{Z.~Dou}, \bibinfo{author}{J.-R. Wen},
\newblock \bibinfo{title}{Rag-critic: Leveraging automated critic-guided agentic workflow for retrieval augmented generation},
\newblock in: \bibinfo{booktitle}{Proceedings of the 63rd Annual Meeting of the Association for Computational Linguistics (Volume 1: Long Papers)}, \bibinfo{year}{2025}, pp. \bibinfo{pages}{3551--3578}.
%Type = Article
\bibitem[{Vaswani et~al.(2017)Vaswani, Shazeer, Parmar, Uszkoreit, Jones, Gomez, Kaiser, and Polosukhin}]{vaswani2017attention}
\bibinfo{author}{A.~Vaswani}, \bibinfo{author}{N.~Shazeer}, \bibinfo{author}{N.~Parmar}, \bibinfo{author}{J.~Uszkoreit}, \bibinfo{author}{L.~Jones}, \bibinfo{author}{A.~N. Gomez}, \bibinfo{author}{{\L}.~Kaiser}, \bibinfo{author}{I.~Polosukhin},
\newblock \bibinfo{title}{Attention is all you need},
\newblock \bibinfo{journal}{Advances in neural information processing systems} \bibinfo{volume}{30} (\bibinfo{year}{2017}).
%Type = Misc
\bibitem[{Shi et~al.(2025)Shi, Kou, Li, Tang, Xie, Yu, Wang, and Zhou}]{shi2025scisage}
\bibinfo{author}{X.~Shi}, \bibinfo{author}{Q.~Kou}, \bibinfo{author}{Y.~Li}, \bibinfo{author}{N.~Tang}, \bibinfo{author}{J.~Xie}, \bibinfo{author}{L.~Yu}, \bibinfo{author}{S.~Wang}, \bibinfo{author}{H.~Zhou}, \bibinfo{title}{Scisage: A multi-agent framework for high-quality scientific survey generation}, \bibinfo{year}{2025}. \URLprefix \url{https://arxiv.org/abs/2506.12689}. \href{http://arxiv.org/abs/2506.12689}{{\tt arXiv:2506.12689}}.
%Type = Inproceedings
\bibitem[{McQueen(1967)}]{mcqueen1967some}
\bibinfo{author}{J.~B. McQueen},
\newblock \bibinfo{title}{Some methods of classification and analysis of multivariate observations},
\newblock in: \bibinfo{booktitle}{Proc. of 5th Berkeley Symposium on Math. Stat. and Prob.}, \bibinfo{year}{1967}, pp. \bibinfo{pages}{281--297}.
%Type = Article
\bibitem[{Donthu et~al.(2021)Donthu, Kumar, Mukherjee, Pandey, and Lim}]{DONTHU2021285}
\bibinfo{author}{N.~Donthu}, \bibinfo{author}{S.~Kumar}, \bibinfo{author}{D.~Mukherjee}, \bibinfo{author}{N.~Pandey}, \bibinfo{author}{W.~M. Lim},
\newblock \bibinfo{title}{How to conduct a bibliometric analysis: An overview and guidelines},
\newblock \bibinfo{journal}{Journal of Business Research} \bibinfo{volume}{133} (\bibinfo{year}{2021}) \bibinfo{pages}{285--296}. \URLprefix \url{https://www.sciencedirect.com/science/article/pii/S0148296321003155}. \DOIprefix\doi{https://doi.org/10.1016/j.jbusres.2021.04.070}.
%Type = Article
\bibitem[{Hicks et~al.(2015)Hicks, Wouters, Waltman, De~Rijcke, and Rafols}]{hicks2015bibliometrics}
\bibinfo{author}{D.~Hicks}, \bibinfo{author}{P.~Wouters}, \bibinfo{author}{L.~Waltman}, \bibinfo{author}{S.~De~Rijcke}, \bibinfo{author}{I.~Rafols},
\newblock \bibinfo{title}{Bibliometrics: the leiden manifesto for research metrics},
\newblock \bibinfo{journal}{Nature} \bibinfo{volume}{520} (\bibinfo{year}{2015}) \bibinfo{pages}{429--431}. \DOIprefix\doi{https://doi.org/10.1038/520429a}.
%Type = Misc
\bibitem[{Niu et~al.(2025)Niu, Liu, Gu, Wang, Ouyang, Zhao, Chu, He, Wu, Zhang, Jin, Liang, Zhang, Zhang, Qu, Ren, Sun, Zheng, Ma, Tang, Niu, Miao, Dong, Qian, Zhang, Chen, Wang, Zhao, Wei, Li, Wang, Xu, Cao, Chen, Wu, Gu, Lu, Wang, Lin, Shen, Zhou, Zhang, Zang, Dong, Wang, Zhang, Bai, Chu, Li, Wu, Wu, Li, Wang, Tu, Xu, Chen, Qiao, Zhou, Lin, Zhang, and He}]{niu2025mineru25decoupledvisionlanguagemodel}
\bibinfo{author}{J.~Niu}, \bibinfo{author}{Z.~Liu}, \bibinfo{author}{Z.~Gu}, \bibinfo{author}{B.~Wang}, \bibinfo{author}{L.~Ouyang}, \bibinfo{author}{Z.~Zhao}, \bibinfo{author}{T.~Chu}, \bibinfo{author}{T.~He}, \bibinfo{author}{F.~Wu}, \bibinfo{author}{Q.~Zhang}, \bibinfo{author}{Z.~Jin}, \bibinfo{author}{G.~Liang}, \bibinfo{author}{R.~Zhang}, \bibinfo{author}{W.~Zhang}, \bibinfo{author}{Y.~Qu}, \bibinfo{author}{Z.~Ren}, \bibinfo{author}{Y.~Sun}, \bibinfo{author}{Y.~Zheng}, \bibinfo{author}{D.~Ma}, \bibinfo{author}{Z.~Tang}, \bibinfo{author}{B.~Niu}, \bibinfo{author}{Z.~Miao}, \bibinfo{author}{H.~Dong}, \bibinfo{author}{S.~Qian}, \bibinfo{author}{J.~Zhang}, \bibinfo{author}{J.~Chen}, \bibinfo{author}{F.~Wang}, \bibinfo{author}{X.~Zhao}, \bibinfo{author}{L.~Wei}, \bibinfo{author}{W.~Li}, \bibinfo{author}{S.~Wang}, \bibinfo{author}{R.~Xu}, \bibinfo{author}{Y.~Cao}, \bibinfo{author}{L.~Chen}, \bibinfo{author}{Q.~Wu}, \bibinfo{author}{H.~Gu}, \bibinfo{author}{L.~Lu}, \bibinfo{author}{K.~Wang},
  \bibinfo{author}{D.~Lin}, \bibinfo{author}{G.~Shen}, \bibinfo{author}{X.~Zhou}, \bibinfo{author}{L.~Zhang}, \bibinfo{author}{Y.~Zang}, \bibinfo{author}{X.~Dong}, \bibinfo{author}{J.~Wang}, \bibinfo{author}{B.~Zhang}, \bibinfo{author}{L.~Bai}, \bibinfo{author}{P.~Chu}, \bibinfo{author}{W.~Li}, \bibinfo{author}{J.~Wu}, \bibinfo{author}{L.~Wu}, \bibinfo{author}{Z.~Li}, \bibinfo{author}{G.~Wang}, \bibinfo{author}{Z.~Tu}, \bibinfo{author}{C.~Xu}, \bibinfo{author}{K.~Chen}, \bibinfo{author}{Y.~Qiao}, \bibinfo{author}{B.~Zhou}, \bibinfo{author}{D.~Lin}, \bibinfo{author}{W.~Zhang}, \bibinfo{author}{C.~He}, \bibinfo{title}{Mineru2.5: A decoupled vision-language model for efficient high-resolution document parsing}, \bibinfo{year}{2025}. \URLprefix \url{https://arxiv.org/abs/2509.22186}. \href{http://arxiv.org/abs/2509.22186}{{\tt arXiv:2509.22186}}.
%Type = Inproceedings
\bibitem[{Zhu et~al.(2025)Zhu, Weng, Yang, and Zhang}]{zhu-etal-2025-deepreview}
\bibinfo{author}{M.~Zhu}, \bibinfo{author}{Y.~Weng}, \bibinfo{author}{L.~Yang}, \bibinfo{author}{Y.~Zhang},
\newblock \bibinfo{title}{{D}eep{R}eview: Improving {LLM}-based paper review with human-like deep thinking process},
\newblock in: \bibinfo{booktitle}{Proceedings of the 63rd Annual Meeting of the Association for Computational Linguistics (Volume 1: Long Papers)}, \bibinfo{year}{2025}, pp. \bibinfo{pages}{29330--29355}. \DOIprefix\doi{10.18653/v1/2025.acl-long.1420}.
%Type = Inproceedings
\bibitem[{Lin(2004)}]{lin2004rouge}
\bibinfo{author}{C.-Y. Lin},
\newblock \bibinfo{title}{Rouge: A package for automatic evaluation of summaries},
\newblock in: \bibinfo{booktitle}{Text summarization branches out}, \bibinfo{year}{2004}, pp. \bibinfo{pages}{74--81}.
%Type = Article
\bibitem[{Nguyen and Luong(2025)}]{NguyenLuongKBS2025}
\bibinfo{author}{T.~H. Nguyen}, \bibinfo{author}{N.~H. Luong},
\newblock \bibinfo{title}{Diverse and high-quality text generation assisted by large language models},
\newblock \bibinfo{journal}{Knowledge-Based Systems}  (\bibinfo{year}{2025}) \bibinfo{pages}{114954}. \DOIprefix\doi{https://doi.org/10.1016/j.knosys.2025.114954}.
%Type = Inproceedings
\bibitem[{Qi et~al.(2024)Qi, Xu, Guo, Wang, Zhang, and Xu}]{qi2024long2rag}
\bibinfo{author}{Z.~Qi}, \bibinfo{author}{R.~Xu}, \bibinfo{author}{Z.~Guo}, \bibinfo{author}{C.~Wang}, \bibinfo{author}{H.~Zhang}, \bibinfo{author}{W.~Xu},
\newblock \bibinfo{title}{$long^{2}rag$: Evaluating long-context {\&} long-form retrieval-augmented generation with key point recall},
\newblock in: \bibinfo{booktitle}{Findings of the Association for Computational Linguistics: EMNLP 2024}, \bibinfo{address}{Miami, Florida, USA}, \bibinfo{year}{2024}, pp. \bibinfo{pages}{4852--4872}. \DOIprefix\doi{10.18653/v1/2024.findings-emnlp.279}.
%Type = Article
\bibitem[{Donthu et~al.(2021)Donthu, Kumar, Mukherjee, Pandey, and Lim}]{donthu2021conduct}
\bibinfo{author}{N.~Donthu}, \bibinfo{author}{S.~Kumar}, \bibinfo{author}{D.~Mukherjee}, \bibinfo{author}{N.~Pandey}, \bibinfo{author}{W.~M. Lim},
\newblock \bibinfo{title}{How to conduct a bibliometric analysis: An overview and guidelines},
\newblock \bibinfo{journal}{Journal of Business Research} \bibinfo{volume}{133} (\bibinfo{year}{2021}) \bibinfo{pages}{285--296}. \DOIprefix\doi{https://doi.org/10.1016/j.jbusres.2021.04.070}.
%Type = Article
\bibitem[{{Corrêa Jr.} et~al.(2017){Corrêa Jr.}, Silva, {da F. Costa}, and Amancio}]{CORREAJR2017498}
\bibinfo{author}{E.~A. {Corrêa Jr.}}, \bibinfo{author}{F.~N. Silva}, \bibinfo{author}{L.~{da F. Costa}}, \bibinfo{author}{D.~R. Amancio},
\newblock \bibinfo{title}{Patterns of authors contribution in scientific manuscripts},
\newblock \bibinfo{journal}{Journal of Informetrics} \bibinfo{volume}{11} (\bibinfo{year}{2017}) \bibinfo{pages}{498--510}. \DOIprefix\doi{https://doi.org/10.1016/j.joi.2017.03.003}.
%Type = Article
\bibitem[{Gasparyan et~al.(2011)Gasparyan, Ayvazyan, Blackmore, and Kitas}]{gasparyan2011writing}
\bibinfo{author}{A.~Y. Gasparyan}, \bibinfo{author}{L.~Ayvazyan}, \bibinfo{author}{H.~Blackmore}, \bibinfo{author}{G.~D. Kitas},
\newblock \bibinfo{title}{Writing a narrative biomedical review: considerations for authors, peer reviewers, and editors},
\newblock \bibinfo{journal}{Rheumatology international} \bibinfo{volume}{31} (\bibinfo{year}{2011}) \bibinfo{pages}{1409--1417}.
%Type = Article
\bibitem[{Sollaci and Pereira(2004)}]{sollaci2004introduction}
\bibinfo{author}{L.~B. Sollaci}, \bibinfo{author}{M.~G. Pereira},
\newblock \bibinfo{title}{The introduction, methods, results, and discussion (imrad) structure: a fifty-year survey},
\newblock \bibinfo{journal}{Journal of the medical library association} \bibinfo{volume}{92} (\bibinfo{year}{2004}) \bibinfo{pages}{364}. \DOIprefix\doi{10.1007/s00296-011-1999-3}.
%Type = Article
\bibitem[{Kumar(2023)}]{kumar2023improvingScientific}
\bibinfo{author}{P.~Kumar},
\newblock \bibinfo{title}{Improving imrad for writing research articles in social, and health sciences},
\newblock \bibinfo{journal}{International Research Journal of Economics and Management Studies IRJEMS} \bibinfo{volume}{2} (\bibinfo{year}{2023}). \DOIprefix\doi{10.56472/25835238/IRJEMS-V2I1P107}.
%Type = Article
\bibitem[{Peppas et~al.(2000)Peppas, Huang, Torres-Lugo, Ward, and Zhang}]{peppas2000physicochemical}
\bibinfo{author}{N.~A. Peppas}, \bibinfo{author}{Y.~Huang}, \bibinfo{author}{M.~Torres-Lugo}, \bibinfo{author}{J.~Ward}, \bibinfo{author}{J.~Zhang},
\newblock \bibinfo{title}{Physicochemical foundations and structural design of hydrogels in medicine and biology},
\newblock \bibinfo{journal}{Annual review of biomedical engineering} \bibinfo{volume}{2} (\bibinfo{year}{2000}) \bibinfo{pages}{9--29}.
%Type = Article
\bibitem[{Liu et~al.(2025)Liu, Cao, Liu, Ding, and Jin}]{liu2025datasets}
\bibinfo{author}{Y.~Liu}, \bibinfo{author}{J.~Cao}, \bibinfo{author}{C.~Liu}, \bibinfo{author}{K.~Ding}, \bibinfo{author}{L.~Jin},
\newblock \bibinfo{title}{Datasets for large language models: A comprehensive survey},
\newblock \bibinfo{journal}{Artificial Intelligence Review} \bibinfo{volume}{58} (\bibinfo{year}{2025}) \bibinfo{pages}{403}. \DOIprefix\doi{10.1007/s10462-025-11403-7}.
%Type = Article
\bibitem[{Wu et~al.(2025)Wu, Yang, Zhan, Yuan, Chao, and Wong}]{wu2023survey}
\bibinfo{author}{J.~Wu}, \bibinfo{author}{S.~Yang}, \bibinfo{author}{R.~Zhan}, \bibinfo{author}{Y.~Yuan}, \bibinfo{author}{L.~S. Chao}, \bibinfo{author}{D.~F. Wong},
\newblock \bibinfo{title}{A survey on llm-generated text detection: Necessity, methods, and future directions},
\newblock \bibinfo{journal}{Computational Linguistics} \bibinfo{volume}{51} (\bibinfo{year}{2025}) \bibinfo{pages}{275--338}. \DOIprefix\doi{10.1162/coli_a_00549}.
%Type = Article
\bibitem[{Zhao et~al.(2024)Zhao, Fan, Li, Liu, Mei, Wang, Wen, Wang, Zhao, Tang, and Li}]{zhao2024recommender}
\bibinfo{author}{Z.~Zhao}, \bibinfo{author}{W.~Fan}, \bibinfo{author}{J.~Li}, \bibinfo{author}{Y.~Liu}, \bibinfo{author}{X.~Mei}, \bibinfo{author}{Y.~Wang}, \bibinfo{author}{Z.~Wen}, \bibinfo{author}{F.~Wang}, \bibinfo{author}{X.~Zhao}, \bibinfo{author}{J.~Tang}, \bibinfo{author}{Q.~Li},
\newblock \bibinfo{title}{Recommender systems in the era of large language models (llms)},
\newblock \bibinfo{journal}{IEEE Transactions on Knowledge and Data Engineering} \bibinfo{volume}{36} (\bibinfo{year}{2024}) \bibinfo{pages}{6889--6907}. \DOIprefix\doi{10.1109/TKDE.2024.3392335}.
%Type = Inproceedings
\bibitem[{Xu and McAuley(2023)}]{xu2023survey}
\bibinfo{author}{C.~Xu}, \bibinfo{author}{J.~McAuley},
\newblock \bibinfo{title}{A survey on model compression and acceleration for pretrained language models},
\newblock in: \bibinfo{booktitle}{Proceedings of the AAAI Conference on Artificial Intelligence}, volume~\bibinfo{volume}{37}, \bibinfo{year}{2023}, pp. \bibinfo{pages}{10566--10575}.
%Type = Article
\bibitem[{Zhu et~al.(2025)Zhu, Yuan, Wang, Liu, Liu, Deng, Chen, Liu, Dou, and Wen}]{zhu2025large}
\bibinfo{author}{Y.~Zhu}, \bibinfo{author}{H.~Yuan}, \bibinfo{author}{S.~Wang}, \bibinfo{author}{J.~Liu}, \bibinfo{author}{W.~Liu}, \bibinfo{author}{C.~Deng}, \bibinfo{author}{H.~Chen}, \bibinfo{author}{Z.~Liu}, \bibinfo{author}{Z.~Dou}, \bibinfo{author}{J.-R. Wen},
\newblock \bibinfo{title}{Large language models for information retrieval: A survey},
\newblock \bibinfo{journal}{ACM Trans. Inf. Syst.} \bibinfo{volume}{44} (\bibinfo{year}{2025}). \DOIprefix\doi{10.1145/3748304}.
%Type = Article
\bibitem[{Hou et~al.(2024)Hou, Zhao, Liu, Yang, Wang, Li, Luo, Lo, Grundy, and Wang}]{hou2024large}
\bibinfo{author}{X.~Hou}, \bibinfo{author}{Y.~Zhao}, \bibinfo{author}{Y.~Liu}, \bibinfo{author}{Z.~Yang}, \bibinfo{author}{K.~Wang}, \bibinfo{author}{L.~Li}, \bibinfo{author}{X.~Luo}, \bibinfo{author}{D.~Lo}, \bibinfo{author}{J.~Grundy}, \bibinfo{author}{H.~Wang},
\newblock \bibinfo{title}{Large language models for software engineering: A systematic literature review},
\newblock \bibinfo{journal}{ACM Transactions on Software Engineering and Methodology} \bibinfo{volume}{33} (\bibinfo{year}{2024}). \DOIprefix\doi{10.1145/3695988}.
%Type = Article
\bibitem[{Zhu et~al.(2024)Zhu, Wang, Chen, Qiao, Ou, Yao, Deng, Chen, and Zhang}]{zhu2024llms}
\bibinfo{author}{Y.~Zhu}, \bibinfo{author}{X.~Wang}, \bibinfo{author}{J.~Chen}, \bibinfo{author}{S.~Qiao}, \bibinfo{author}{Y.~Ou}, \bibinfo{author}{Y.~Yao}, \bibinfo{author}{S.~Deng}, \bibinfo{author}{H.~Chen}, \bibinfo{author}{N.~Zhang},
\newblock \bibinfo{title}{Llms for knowledge graph construction and reasoning: Recent capabilities and future opportunities},
\newblock \bibinfo{journal}{World Wide Web} \bibinfo{volume}{27} (\bibinfo{year}{2024}) \bibinfo{pages}{58}. \DOIprefix\doi{10.1007/s11280-024-01297-w}.
%Type = Article
\bibitem[{Brito et~al.(2023)Brito, Oliveira, Oliveira~Jr, Silva, and Amancio}]{brito2023network}
\bibinfo{author}{A.~C.~M. Brito}, \bibinfo{author}{M.~C.~F. Oliveira}, \bibinfo{author}{O.~N. Oliveira~Jr}, \bibinfo{author}{F.~N. Silva}, \bibinfo{author}{D.~R. Amancio},
\newblock \bibinfo{title}{Network analysis and natural language processing to obtain a landscape of the scientific literature on materials applications},
\newblock \bibinfo{journal}{ACS Applied Materials \& Interfaces} \bibinfo{volume}{15} (\bibinfo{year}{2023}) \bibinfo{pages}{27437--27446}.
%Type = Article
\bibitem[{Silva et~al.(2025)Silva, Gouveia, Zielinski, Oliveira, Amancio, Bruno, and Oliveira~Jr}]{silva2025ai}
\bibinfo{author}{J.~C. Silva}, \bibinfo{author}{R.~P. Gouveia}, \bibinfo{author}{K.~M. Zielinski}, \bibinfo{author}{M.~C.~F. Oliveira}, \bibinfo{author}{D.~R. Amancio}, \bibinfo{author}{O.~M. Bruno}, \bibinfo{author}{O.~N. Oliveira~Jr},
\newblock \bibinfo{title}{Ai-assisted tools for scientific review writing: Opportunities and cautions},
\newblock \bibinfo{journal}{ACS Applied Materials \& Interfaces} \bibinfo{volume}{17} (\bibinfo{year}{2025}) \bibinfo{pages}{47795--47805}.
%Type = Inproceedings
\bibitem[{Zhang et~al.(2025)Zhang, Cao, You, and Ding}]{zhang-etal-2025-prompt-design}
\bibinfo{author}{X.~Zhang}, \bibinfo{author}{J.~Cao}, \bibinfo{author}{C.~You}, \bibinfo{author}{D.~Ding},
\newblock \bibinfo{title}{Why prompt design matters and works: A complexity analysis of prompt search space in {LLM}s},
\newblock in: \bibinfo{booktitle}{Proceedings of the 63rd Annual Meeting of the Association for Computational Linguistics (Volume 1: Long Papers)}, \bibinfo{address}{Vienna, Austria}, \bibinfo{year}{2025}, pp. \bibinfo{pages}{32525--32555}. \DOIprefix\doi{10.18653/v1/2025.acl-long.1562}.
%Type = Article
\bibitem[{Atreja et~al.(2025)Atreja, Ashkinaze, Li, Mendelsohn, and Hemphill}]{atreja2024prompt}
\bibinfo{author}{S.~Atreja}, \bibinfo{author}{J.~Ashkinaze}, \bibinfo{author}{L.~Li}, \bibinfo{author}{J.~Mendelsohn}, \bibinfo{author}{L.~Hemphill},
\newblock \bibinfo{title}{What’s in a prompt?: A large-scale experiment to assess the impact of prompt design on the compliance and accuracy of llm-generated text annotations},
\newblock \bibinfo{journal}{Proceedings of the International AAAI Conference on Web and Social Media} \bibinfo{volume}{19} (\bibinfo{year}{2025}) \bibinfo{pages}{122–145}. \URLprefix \url{http://dx.doi.org/10.1609/icwsm.v19i1.35807}. \DOIprefix\doi{10.1609/icwsm.v19i1.35807}.

\end{thebibliography}

% %\vskip3pt
% \end{document}

% Loading bibliography style file
\clearpage
% Ensure the reference list starts on a clean page
\newpage
\thispagestyle{empty}  % Remove the header/footer

\end{document}